\documentclass[10pt,journal,compsoc]{IEEEtran}

\usepackage{amsmath,amsfonts,bm}
\usepackage{amssymb}
\usepackage{array}
\usepackage{multirow}
\usepackage{booktabs}
\usepackage[caption=false,font=normalsize,labelfont=sf,textfont=sf]{subfig}
\usepackage{textcomp}
\usepackage{stfloats}
\usepackage{url}
\usepackage{verbatim}
\usepackage[table]{xcolor}
\usepackage{tabularx}
\usepackage{graphicx}
\usepackage{tikz}
\usepackage{xcolor,tikz}
\usepackage[most]{tcolorbox}
\usepackage{enumitem}
\usepackage{capt-of}
\usepackage{makecell}
\usepackage{pifont}
\usepackage{amsmath,amsfonts,amssymb}
\usepackage{bm}
\usepackage[ruled,vlined]{algorithm2e}
\newcommand{\righticon}{\raisebox{-0.15ex}{\includegraphics[height=1.05ex,keepaspectratio]{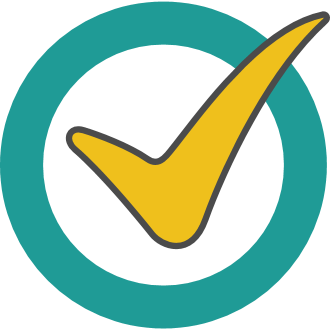}}}
\newcommand{\wrongicon}{\raisebox{-0.15ex}{\includegraphics[height=1.05ex,keepaspectratio]{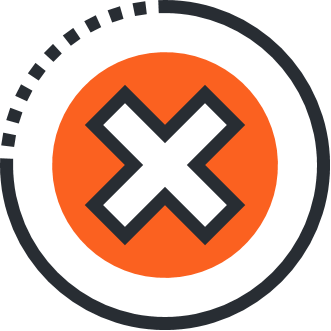}}}

\newcommand{\mapo}{\raisebox{-0.15ex}{\includegraphics[height=1.15ex,keepaspectratio]{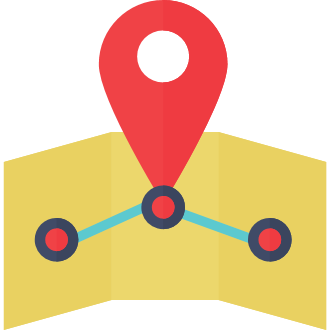}}}

\newcommand{\wifio}{\raisebox{-0.15ex}{\includegraphics[height=1.15ex,keepaspectratio]{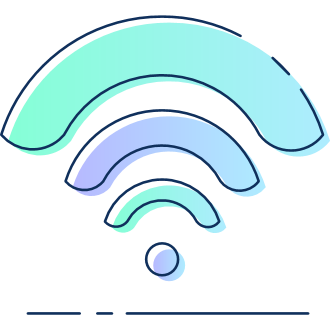}}}

\newcommand{\da}{\raisebox{-0.15ex}{\includegraphics[height=1.15ex,keepaspectratio]{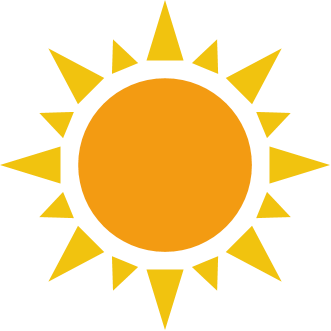}}}
\newcommand{\sr}{\raisebox{-0.15ex}{\includegraphics[height=1.15ex,keepaspectratio]{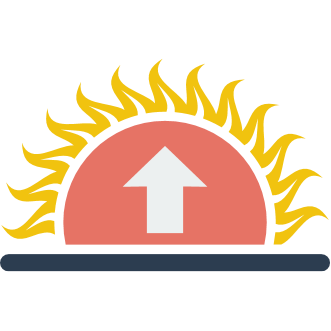}}}
\newcommand{\sd}{\raisebox{-0.15ex}{\includegraphics[height=1.15ex,keepaspectratio]{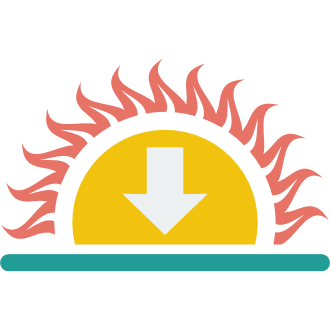}}}
\newcommand{\ev}{\raisebox{-0.15ex}{\includegraphics[height=1.15ex,keepaspectratio]{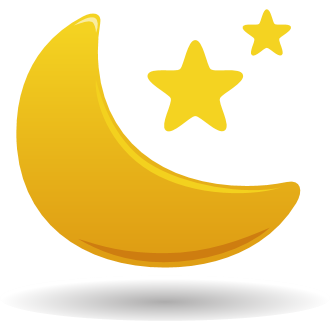}}}

\newcommand{\lidar}{\raisebox{-0.12ex}{\includegraphics[height=1.0ex,keepaspectratio]{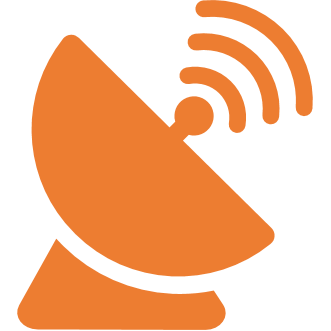}}}

\definecolor{skyblue}{rgb}{0.82,0.89,0.94}
\definecolor{mintgreen}{rgb}{0.83,0.93,0.86}
\definecolor{lavenderlight}{rgb}{0.85,0.84,0.93}
\definecolor{warmyellow}{rgb}{1.00,0.95,0.79}
\definecolor{peach}{rgb}{0.99,0.90,0.84}

\newcommand{\Nbase}[4][0.45]{%
  \tikz[baseline=(char.base)]{
    \shade[
      inner color=#2,
      outer color=#2!30!#3  
    ] (0,0) circle (#1/2);  
    \node[anchor=center] (char) at (0,0) {\scriptsize\textbf{#4}};  
  }%
}

\newcommand{\NO}[1][0.30]{%
  \Nbase[#1]{skyblue}{white}{1}%
}

\newcommand{\NTw}[1][0.30]{%
  \Nbase[#1]{skyblue}{white}{2}}

\newcommand{\NTh}[1][0.30]{%
  \Nbase[#1]{mintgreen}{white}{3}}
\newcommand{\NFo}[1][0.30]{%
  \Nbase[#1]{mintgreen}{white}{4}}

\newcommand{\NFiv}[1][0.30]{%
  \Nbase[#1]{lavenderlight}{white}{5}}
\newcommand{\NSi}[1][0.30]{%
  \Nbase[#1]{lavenderlight}{white}{6}}

\newcommand{\NSe}[1][0.30]{%
  \Nbase[#1]{warmyellow}{white}{7}}
\newcommand{\NEi}[1][0.30]{%
  \Nbase[#1]{warmyellow}{white}{8}}
\newcommand{\NNi}[1][0.30]{%
  \Nbase[#1]{warmyellow}{white}{9}}

\newcommand{\NTe}[1][0.30]{%
  \Nbase[#1]{peach}{white}{10}}
\newcommand{\NEl}[1][0.30]{%
  \Nbase[#1]{peach}{white}{11}}
\newcommand{\NTww}[1][0.30]{%
  \Nbase[#1]{peach}{white}{12}}
\newcommand{\NThh}[1][0.30]{%
  \Nbase[#1]{peach}{white}{13}}

\usepackage{ragged2e} % 用于abstract justifying

\ifCLASSOPTIONcompsoc

  \usepackage[nocompress]{cite}
\else
  \usepackage{cite}
\fi

\ifCLASSINFOpdf
\else
\fi

\newcommand\MYhyperrefoptions{%
  bookmarks=true,
  bookmarksnumbered=true,
  pdfpagemode={UseOutlines},
  plainpages=false,
  pdfpagelabels=true,
  colorlinks=true,
  linkcolor={blue},
  citecolor={blue},
  urlcolor={blue},
  pdftitle={Spheriverse: 3D Scene Understanding from Spherical Observations in the Wild},
  pdfsubject={Spherical 3D Perception},
  pdfauthor={Fei Teng},
  pdfkeywords={panoramic vision, spherical imaging, 3D perception, benchmark}
}

\usepackage[\MYhyperrefoptions]{hyperref}

\usepackage{icomma}
\usepackage{marvosym}

\begin{document}

\title{Spheriverse: 3D Scene Understanding from Spherical Observations in the Wild}

\author{Fei Teng$^*$ \quad Sheng Wu$^*$ \quad Mengfei Duan$^*$ \quad Guoqiang Zhao \quad Junhui Ma \quad Kai Luo\\Siyu Li \quad Hao Shi \quad Zhiyong Li \quad Kailun Yang$\textsuperscript{\Letter}$
\IEEEcompsocitemizethanks{
\IEEEcompsocthanksitem F. Teng, S. Wu, M. Duan, G. Zhao, J. Ma, K. Luo, S. Li, Z. Li, and K. Yang are with the School of Artificial Intelligence and Robotics and the National Engineering Research Center of Robot Visual Perception and Control Technology, Hunan University, China.
\IEEEcompsocthanksitem S. Li is also with the School of Automation and Electrical Engineering, Zhejiang University of Science and Technology, China.
\IEEEcompsocthanksitem H. Shi is with Ant Group, China.
\IEEEcompsocthanksitem H. Shi is also with the State Key Laboratory of Extreme Photonics and Instrumentation, Zhejiang University, China.
\IEEEcompsocthanksitem $^*$Equal contribution.
\IEEEcompsocthanksitem $\textsuperscript{\Letter}$Corresponding author (E-mail: \href{kailun.yang@hnu.edu.cn}{kailun.yang@hnu.edu.cn}).
}
}

\IEEEtitleabstractindextext{%
\begin{center}
    \centering
    \includegraphics[width=\linewidth]{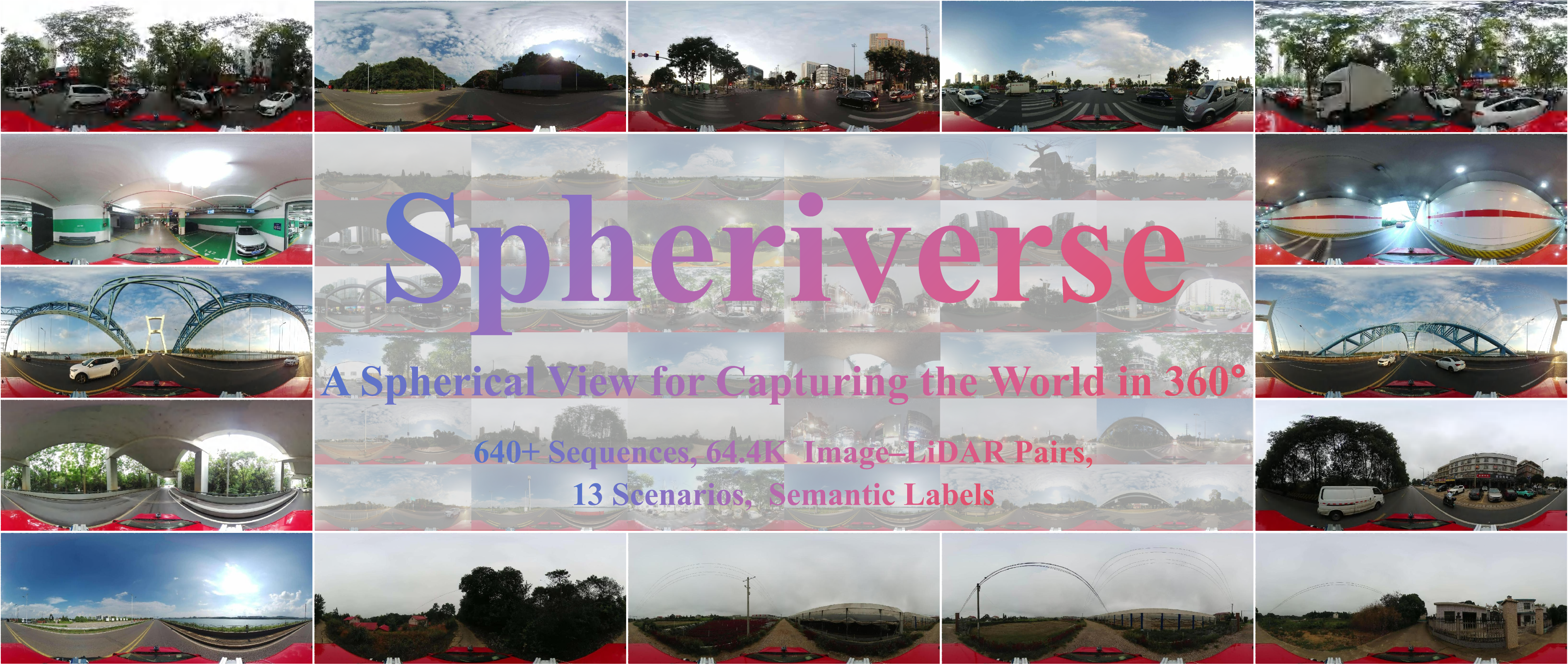}
    \vskip -1.5ex
    \setcounter{figure}{0}
    \captionof{figure}{{Spheriverse: Seeing the world in $360^{\circ}$.}
    From busy urban streets and multi-level transport structures to rural roads and agricultural landscapes, Spheriverse captures the diversity of real-world environments through spherical observations. 
    }
    \label{fig:overview}
    %\vskip -1ex
\end{center}

\begin{abstract} \justifying
Spherical observations provide global visual context for 3D scene understanding. However, visual information is encoded in an angular domain, whereas the physical world is represented in Cartesian coordinates. This cross-space representation gap complicates geometric correspondence and semantic evidence aggregation. To delve into this challenge, we introduce Spheriverse, comprising $64,400$ temporally aligned spherical image-LiDAR pairs organized into $644$ sequences. The dataset spans diverse scenes, illumination, and weather conditions, with fine-grained semantic classes. We further establish benchmarks for semantic occupancy prediction, semantic mapping, and 3D object detection, evaluating $30+$ methods through overall and scene-wise comparisons. For dense prediction, we propose SphereOcc, an occupancy framework that couples spherical geometry modeling with semantic evidence retrieval. Cartesian-Spherical Representation Remodeling (CSRR) incorporates spherical range-azimuth geometry into Cartesian voxel features through region-wise modulation. Spherical Evidence Re-querying (SER) then conditions queries on voxel content and range-height-azimuth geometry to adaptively retrieve relevant semantic evidence from source spherical image features. SphereOcc achieves $13.91\%$ mIoU and $24.65\%$ GeoIoU, yielding relative
improvements of $13.9\%$ and $9.3\%$ over the respective best-performing
methods, TPVFormer and SurroundOcc. It also ranks first in both metrics across all five scene categories, with consistent advantages across the evaluated spatial partitions and reduced fields of view. The established benchmark and source code will be available at \href{https://feit-feiteng.github.io/Spheriverse}{[Homepage]}.

\end{abstract}

}

\maketitle

\IEEEdisplaynontitleabstractindextext

\IEEEpeerreviewmaketitle

\ifCLASSOPTIONcompsoc
\IEEEraisesectionheading{\section{Introduction}\label{sec:intro}}
\else
\section{Introduction}
\label{sec:introduction}
\fi

\IEEEPARstart{S}{pherical} observations provide continuous
$360^\circ$ horizontal coverage and broad vertical context, enabling global and coherent visual representations for spatial modeling and scene understanding~\cite{gao2022review,One,Panogen,NaPano2,NaPano1,NaPano3}. 
This property is highly aligned with the growing demand for embodied intelligence for holistic 3D spatial perception~\cite{Survey,air1,air2}. 
However, existing studies have largely focused on perspective cameras or 2D spherical scene understanding~\cite{coors2018spherenet,zhang2024goodsam,zhong2025omnisam,zhou2024ultrawide}, leaving the potential of spherical vision for 3D spatial perception largely unexplored. 
Yet spherical observations provide full-surround visual coverage. 
For dense 3D scene modeling tasks such as occupancy prediction~\cite{occ3d,openoccupancy,chen2025alocc,huang2024selfocc}, however, visual information is encoded in an angular domain, whereas the target representation resides in a dense Cartesian voxel space.
This divergence creates a cross-space representation gap between angular observations and metric voxels, giving rise to a range of projection-induced distortions in object appearance, geometry, and spatial relationships.
To address these challenges, this work contributes from three perspectives.

To enable the study of spherical 3D perception under the aforementioned challenges, we collect and organize a real-world spherical dataset with rich semantic annotations, termed \textit{Spheriverse}. 
As shown in Tab.~\ref{tab:1}, Spheriverse exhibits the following key properties: $644$ sequences collected across $13$ geographically diverse regions, covering different times of day and a variety of real-world environments;
%intentionally curated within a unified data collection framework to mitigate cross-device domain shifts and enable consistent evaluation of spherical 3D perception;
24-hour real-world data under diverse weather conditions, covering sunrise, sunset, daytime, and late-night scenarios, as well as clear and rainy weather; Fine-grained manual 3D annotations for outdoor object categories, including diverse artifacts such as temporary buildings, advertising objects, art objects, and pavilions; A high-quality sensing setup comprising cameras with a 360{\textdegree} horizontal and 136.7{\textdegree} vertical field of view, equipped with industrial-grade image sensors, together with a 128-beam LiDAR, providing dense and reliable observations for spherical 3D perception; Comprehensive dataset statistics, such as inter-scene appearance variations and object–FoV relationships.

% \begin{enumerate}[label=\textit{\arabic*)}]
%     \item $644$ sequences collected across $13$ geographically and visually diverse regions spanning distinct real-world regions and environments. 
%     %, intentionally curated within a unified data collection framework to mitigate cross-device domain shifts and enable consistent evaluation of spherical 3D perception.
%     \item 24-hour real-world data under diverse weather conditions, covering sunrise, sunset, daytime, and late-night scenarios, as well as clear and rainy weather. 
%     %, enabling evaluation under realistic illumination variations.
%     \item Fine-grained manual 3D annotations for $33$ outdoor object categories, including diverse artifacts such as noise barriers, food vendors, electric scooters, greenhouses, and other artificial objects. 
%     %, supporting comprehensive evaluation in complex modern scenes.
%     \item A high-quality sensing setup comprising cameras with a 360{\textdegree} horizontal and 145{\textdegree} vertical field of view, equipped with industrial-grade image sensors, together with a 128-beam LiDAR, providing dense and reliable observations for spherical 3D perception.
%     \item Comprehensive dataset statistics and analysis, such as inter-scene appearance variations and object–FoV relationships, offering quantitative insights into the intrinsic characteristics of spherical perception.
% \end{enumerate}

Occupancy prediction provides a comprehensive representation for 3D spatial understanding and serves as a challenging testbed for studying holistic scene perception. 
Therefore, we adapt $11$ representative methods to Spheriverse and further develop a spherical perception framework to investigate effective 3D spatial representations under spherical observations.
To broaden the applicability of Spheriverse, we further establish two benchmarks for semantic mapping and 3D object detection. 
Beyond the overall performance comparison, we conduct a fine-grained analysis across the five major scene categories in Spheriverse, revealing the influence of scene geometry, semantic composition, structural complexity, and observation conditions on model performance. 
Overall, we benchmark over $30$ methods on Spheriverse and perform evaluations across three benchmarks, providing an evaluation platform for future research on spherical 3D understanding.

\begin{figure}[ht]
    \centering
    \includegraphics[width=0.48\textwidth]{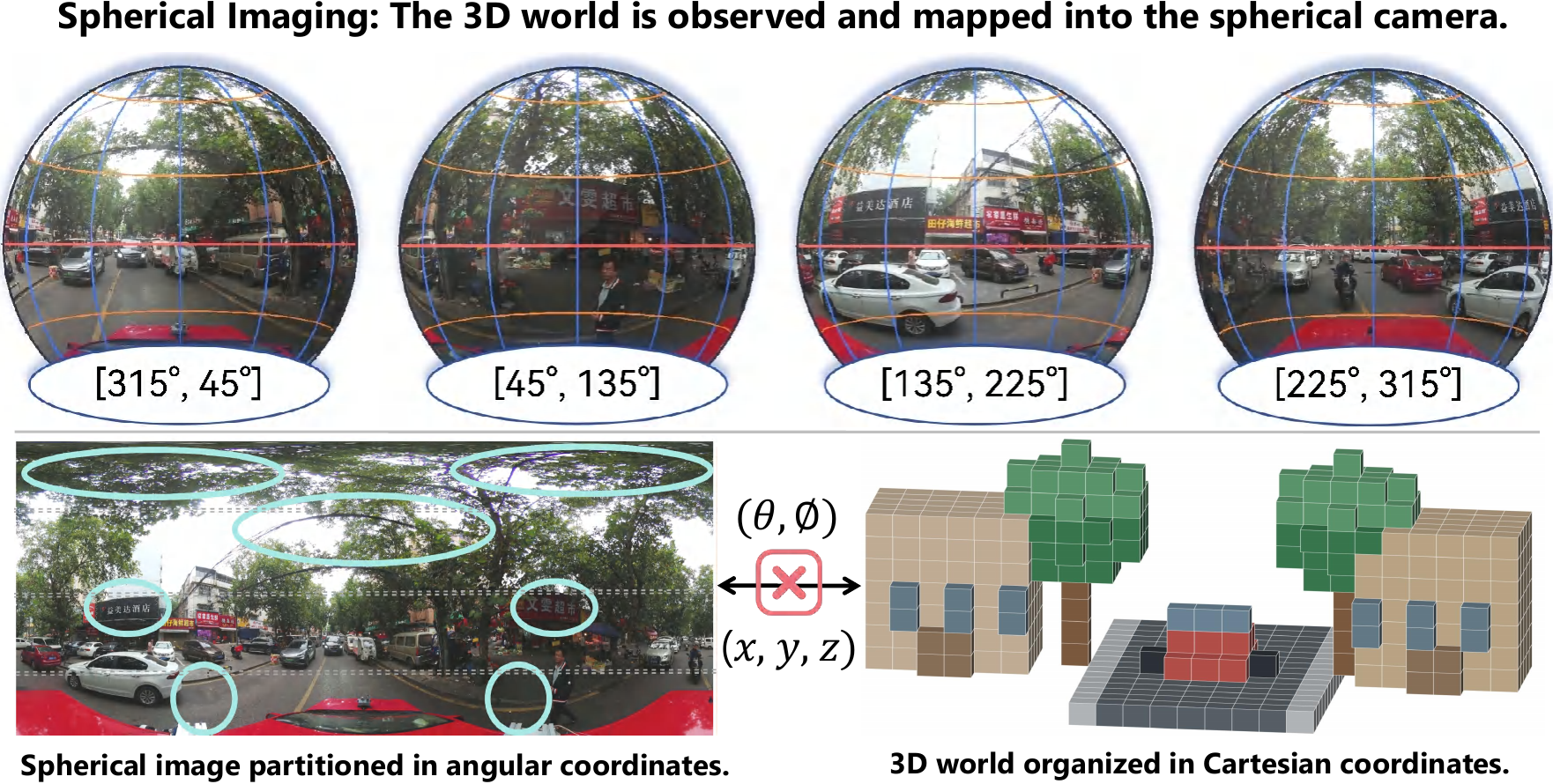} 
    \caption{Spherical--Cartesian representation gap. Spherical imaging captures a continuous $360^{\circ}$ horizontal field of view, with observations parameterized by azimuth $\theta$ and elevation $\phi$. In contrast, the 3D environment and its voxel representation are organized in Cartesian coordinates $(x,y,z)$. The mapping between these angular and metric coordinate systems is spatially non-uniform, causing projection-dependent distortions and preventing image-space proximity from directly preserving metric neighborhood relationships in 3D. The cyan outlines highlight representative regions where spherical projection alters object shape, scale, or local spatial arrangement.
    }
    \label{fig:motivation}
\end{figure}

%{\textbf{Method: SphereOcc.}} 
The divergence between spherical imagery and Cartesian voxel space creates a cross-space representation gap,  giving rise to a
range of projection-induced distortions in object appearance, geometry, and spatial relationships. 
To bridge this gap, we propose SphereOcc, an occupancy framework that integrates panoramic geometry and semantic evidence into metric voxel representations through a {geometry–semantics} coupled {cross-space} mechanism. 
Specifically, Cartesian–Spherical Representation Remodeling (CSRR) encodes range–azimuth relations to form Cartesian–spherical voxel features, thereby aligning Cartesian voxel representations with spherical range–azimuth geometry. Furthermore, Spherical Evidence Re-querying (SER) uses their range–height–azimuth (RTZ) geometry to mimic the 3D-to-2D coordinate mapping induced by spherical imaging, thereby complementing each voxel feature with relevant semantic evidence.

Our unified evaluation across three benchmarks reveals pronounced cross-scene performance variation among existing 3D perception methods under spherical observations. For dense occupancy prediction, SphereOcc achieves $13.91\%$ mIoU and $24.65\%$ GeoIoU. The strongest prior results are $12.21\%$ mIoU from TPVFormer~\cite{TPVFormer} and $22.55\%$ GeoIoU from SurroundOcc~\cite{surroundocc}. Compared with these results, SphereOcc improves mIoU and GeoIoU by $1.70$ and $2.10$ percentage points, corresponding to relative gains of $13.9\%$ and $9.3\%$, respectively. It also ranks first in both metrics across all five scene categories, demonstrating consistent improvements across diverse environments. We further conduct region-wise evaluations across horizontal azimuth intervals and vertical voxel layers, as well as experiments with reduced spherical fields of view. 
SphereOcc maintains effective performance across different settings.

Overall, this work delivers the following contributions:
\begin{itemize}
    \item We introduce \textit{Spheriverse}, a large-scale real-world spherical perception dataset featuring diverse geographic environments, 24-hour observations, fine-grained 3D semantic annotations, and high-quality multimodal sensing, enabling comprehensive studies of panoramic 3D perception under realistic conditions.
    \item We establish benchmarks on Spheriverse, covering three fundamental 3D tasks, \textit{i.e.}, semantic occupancy prediction, semantic mapping, and 3D object detection. 
    Over $30$ methods are evaluated with both overall and scene-wise analyses, reporting the performance of existing approaches under spherical observations.
    \item We propose \textit{SphereOcc}, an occupancy framework that integrates spherical geometry and semantic evidence into voxel representations through a {geometry–semantics} coupled {cross-space} mechanism.
    \item SphereOcc achieves state-of-the-art semantic occupancy results on Spheriverse, reaching $13.91\%$ mIoU and $24.65\%$ GeoIoU and surpassing the strongest prior results by $1.70$ and $2.10$ percentage points, respectively. It also ranks first in both metrics across all five scene categories. Additional region-wise and reduced-FoV evaluations further demonstrate its consistent performance across spatial variations and observation coverage.
\end{itemize}

\section{Related Work}
\label{sec:relat}
In this section, we first review prior studies on spherical scene understanding in Sec.~\ref{sec2.1}. 
Owing to its completeness in scene representation, we focus on reviewing voxel-based scene understanding in Sec.~\ref{sec2.2}. We then discuss semantic mapping and 3D object detection in Sec.~\ref{sec2.3}.

\begin{table*}[t]
\centering

\renewcommand{\righticon}{\raisebox{-0.15em}{\includegraphics[height=1em,keepaspectratio]{figs/icons/right2.png}}}
\renewcommand{\wrongicon}{\raisebox{-0.15em}{\includegraphics[height=1em,keepaspectratio]{figs/icons/wrong5.png}}}
\renewcommand{\mapo}{\raisebox{-0.15em}{\includegraphics[height=1em,keepaspectratio]{figs/icons/mapo.png}}}
\renewcommand{\wifio}{\raisebox{-0.15em}{\includegraphics[height=1em,keepaspectratio]{figs/icons/wifi1.png}}}
\renewcommand{\da}{\raisebox{-0.15em}{\includegraphics[height=1em,keepaspectratio]{figs/icons/day.png}}}
\renewcommand{\sr}{\raisebox{-0.15em}{\includegraphics[height=1em,keepaspectratio]{figs/icons/sr.png}}}
\renewcommand{\sd}{\raisebox{-0.15em}{\includegraphics[height=1em,keepaspectratio]{figs/icons/sd.png}}}
\renewcommand{\ev}{\raisebox{-0.15em}{\includegraphics[height=1em,keepaspectratio]{figs/icons/ev.png}}}
\renewcommand{\lidar}{\raisebox{-0.15em}{\includegraphics[height=1em,keepaspectratio]{figs/icons/lidar.png}}}

\caption{
Comparison of spherical-view datasets.
\mapo\ and \wifio\ denote direct real-world acquisition and
Internet-sourced imagery, respectively; Syn. denotes synthetic data.
\righticon\ and \wrongicon\ indicate the presence and absence of
the corresponding attribute, respectively.
Env. denotes the scene environment: In., Out., and In./Out. indicate
indoor, outdoor, and both indoor and outdoor environments, respectively.
\sr, \da, \sd, and \ev\ denote sunrise, daytime, sunset, and
evening lighting conditions, respectively.
``\#T.Scenes'' denotes the reported scene count.
``Semantic'' and ``3D Boxes'' indicate the availability of
voxel-wise semantic labels and object-level 3D bounding-box
annotations, respectively. Dep. denotes depth sensing, and B. denotes the number of LiDAR beams.
Scale reports image counts in thousands (K).
}
\label{tab:1}

\renewcommand{\arraystretch}{1.35}
\setlength{\tabcolsep}{2.0pt}
\normalsize

\begin{tabularx}{0.99\textwidth}{@{}
>{\raggedright\arraybackslash}X|
>{\centering\arraybackslash}p{0.045\textwidth}
>{\centering\arraybackslash}p{0.045\textwidth}
>{\centering\arraybackslash}p{0.075\textwidth}|
>{\centering\arraybackslash}p{0.080\textwidth}
>{\centering\arraybackslash}p{0.080\textwidth}
>{\centering\arraybackslash}p{0.070\textwidth}
>{\centering\arraybackslash}p{0.100\textwidth}|
>{\centering\arraybackslash}p{0.080\textwidth}
>{\centering\arraybackslash}p{0.080\textwidth}|
>{\centering\arraybackslash}p{0.050\textwidth}@{}}

\toprule
\multicolumn{1}{c|}{\multirow{2}{*}{Datasets}}
& \multicolumn{3}{c|}{Acquisition}
& \multicolumn{4}{c|}{Diversity}
& \multicolumn{2}{c|}{3D Annotations}
& \multirow{2}{*}{Scale} \\

\cmidrule(lr){2-4}
\cmidrule(lr){5-8}
\cmidrule(lr){9-10}

& \textit{Year}
& \textit{Type}
& \textit{Geo. Sen.}
& \textit{Env.}
& \textit{\#T.Scenes}
& \textit{Weather}
& \textit{Lighting}
& \shortstack{\textit{Semantic}}
& \textit{3D Boxes}
& \\

\midrule

Matterport3D~\cite{Matterport3D}
& 2017 & \mapo & Dep.
& In. & \wrongicon & \wrongicon & \wrongicon
& \righticon & \righticon & 9.6K \\

WildPASS~\cite{Wildpass}
& 2021 & \wifio & \wrongicon
& Out. & 9 & \righticon & \da\ \ev
& \wrongicon & \wrongicon & 2.0K \\

DensePASS~\cite{Denspass}
& 2021 & \wifio & \wrongicon
& Out. & 6 & \wrongicon & \da\ \sd\ \ev
& \wrongicon & \wrongicon & 2.1K \\

HoliCity~\cite{Holicity}
& 2021 & \wifio & \wrongicon
& Out. & 4 & \wrongicon & \da
& \wrongicon & \wrongicon & 6.3K \\

SynPASS~\cite{Synpass}
& 2023 & Syn. & \wrongicon
& Out. & 6 & \righticon & \da\ \ev
& \wrongicon & \wrongicon & 9.0K \\

Pano2Geo~\cite{Pano2geo}
& 2024 & \wifio & \wrongicon
& Out. & 3 & \wrongicon & \da
& \wrongicon & \wrongicon & 1.0K \\

360Loc~\cite{360Loc}
& 2024 & \mapo & 16 B.
& In./Out. & 4 & \righticon & \da\ \ev
& \wrongicon & \wrongicon & 9.3K \\

PAIR360~\cite{Pair360}
& 2024 & \mapo & 32 B.
& Out. & 2 & \wrongicon & \sr\ \da
& \wrongicon & \wrongicon & 88.2K \\

Dur360BEV~\cite{Dur360BEV}
& 2025 & \mapo & 128 B.
& Out. & 4 & \wrongicon & \wrongicon
& \wrongicon & \righticon & 16.4K \\

PanoCycle360~\cite{PanoCycle360}
& 2026 & \mapo & \wrongicon
& Out. & 5 & \righticon & \da\ \ev
& \wrongicon & \wrongicon & 10.1K \\

\specialrule{0.08em}{0ex}{0ex}

\rowcolor{yellow!8}
\rule[-0.5ex]{0pt}{3.4ex}\raisebox{0.25ex}{\textbf{Ours}}
& 2026 & \mapo & \textbf{128 B.}\,\lidar
& Out. & \textbf{13} & \righticon & \sr\ \da\ \sd\ \ev
& \righticon & \righticon & \textbf{64.4K} \\

\bottomrule
\end{tabularx}
\end{table*}

\subsection{Spherical Scene Understanding} \label{sec2.1}
As embodied agents advance, the demand for a full field of view emerges as a key prerequisite for effective environmental perception and understanding~\cite{humanoidpano,One}. 

Existing datasets~\cite{Nuscenes, Omnihd,robosense,Semantickitti} typically acquire surround-view observations using multi-camera systems. 
However, inherent limitations in camera placement and imaging pipelines impose constraints on the field of view and introduce spatially discontinuous and overlapping observations~\cite{onebev,hallucinating}. 
%Even when these observations are projected onto a spherical surface, the missing viewpoints cannot be recovered and preventing the formation of a coherent and seamless omnidirectional representation and consequently hindering holistic scene understanding. 
Although spherical imagery offers a large field of view and more complete observations of the surrounding environment, the resulting research has primarily focused on geometric scene understanding and depth estimation~\cite{Pair360,Pano2geo,onebev}. 
As summarized in Tab.~\ref{tab:1}, our systematic review of representative spherical datasets reveals that semantic 3D spatial understanding directly from spherical observations remains largely underexplored~\cite{Pair360,Pano2geo,onebev}.
Meanwhile, panoramic annular cameras provide only horizontal observations; their inherent structural design often confines the vertical field of view to approximately 45{\textdegree}, which substantially restricts the range of scenes that can be effectively captured~\cite{Pass,yang2020ds,Minimalist,luo2025omnitrack,Panoflow,Panoflow2,oneocc}. Matterport3D~\cite{Matterport3D} provides spherical imagery, yet it mainly covers static indoor environments with regular layouts, leaving outdoor and dynamic scenarios unexplored. 

For spherical scene understanding, researchers formulate spherical semantic segmentation from an unsupervised domain adaptation perspective~\cite{Denspass,Trans4Pass,Synpass,360sfuda++}. 
Meanwhile, beyond closed-set semantic segmentation, recent efforts~\cite{open,pods,Occlusion} extend panoramic understanding toward open-vocabulary and zero-shot settings, enabling learning under annotation-scarce and semantically open conditions. 
Furthermore, studies~\cite{omniflownet,revisiting} adapt optical flow estimation to panoramic imagery through distortion-aware modeling and spherical continuity constraints, alleviating geometric distortions and periodic discontinuities induced by spherical projection. 
Recent approaches~\cite{CC3DT,luo2025omnitrack,360VOTS, omnitrack++} extend object detection and tracking to panoramic settings in 2D space by incorporating distortion-aware representations and trajectory-based temporal modeling. 
% to mitigate geometric distortions and identity ambiguities.
% Spherical vision has also been explored in depth estimation~\cite{helvipad, revisiting, elite360d, spherical,spherical2,bifuse++}. 
Spherical vision has also been explored in depth estimation~\cite{zioulis2018omnidepth,tateno2018distortion,wang2020bifuse,
albanis2021pano3d,reyarea2022monodepth,li2022omnifusion,shen2022panoformer,
ai2023hrdfuse}.
Recent studies further explore semantic-conditioned diffusion and
geometry-consistent zero-shot priors~\cite{mohadikar2025omnidiffusion,yuan2026vggt360}, complementing
contemporary efficient and real-world spherical depth estimation~\cite{helvipad,Revisiting_360,elite360d,bifuse++}. 
Nevertheless, lifting semantic representations from the non-Euclidean spherical manifold into a 3D Cartesian space is essential for embodied agents to construct spatially grounded representations of their surroundings, yet this problem remains largely unexplored. In this work, we introduce Spheriverse, an open-world and diverse spherical dataset with rich spatial annotations.

\subsection{Semantic Occupancy Prediction}~\label{sec2.2} Semantic occupancy prediction aims to infer a dense 3D representation from one or multiple 2D image observations by assigning each voxel an occupancy state and a semantic label, explicitly describing free space, object geometry, and scene semantics, thereby providing essential spatial information for embodied intelligence. 

Existing camera-based approaches can be broadly categorized into three paradigms. 
Explicit projection-based methods, such as MonoScene~\cite{monoscene}, OccDepth~\cite{occdepth}, and COTR~\cite{cotr}, project 2D image features into a predefined 3D voxel space, preserving direct image-to-voxel correspondences and offering strong geometric interpretability. 
Query-based methods~\cite{bevformer,TPVFormer,surroundocc,protoocc,li2023voxformer,occformer} employ learnable BEV or voxel queries to adaptively aggregate image evidence, enabling global contextual reasoning while reducing redundant dense feature lifting. 
Gaussian-based methods~\cite{gaussianformer,gaussianformer2,quadricformer} represent scenes using sparse continuous primitives, allowing computational resources to be concentrated on occupied regions while providing compact and geometrically flexible 3D representations. 
Those methods reconstruct 3D Cartesian representations under conventional perspective-imaging assumptions. 
Consequently, they do not explicitly address the cross-space representation gap between spherical image and  Cartesian voxel space. 
While OneOcc~\cite{oneocc} and PanoMMOcc~\cite{PanoRingCalib} have studied semantic occupancy prediction from panoramic annular cameras~\cite{gao2022review}, their vertical field of view is limited to approximately $45^\circ$, leaving wide-FoV spherical occupancy unexplored. 
Furthermore, several LiDAR-assisted approaches, including SPHERE~\cite{sphere} and EFFOcc~\cite{EFFOcc}, employ point clouds to provide explicit geometric cues and alleviate the ambiguity of image-to-voxel reconstruction. However, introducing LiDAR entails additional sensor payload, power consumption, and onboard computational overhead, imposing considerable constraints on compact embodied platforms. 
In this work, we propose SphereOcc for camera-based spherical 3D perception to address a cross-space representation gap between the angular domain and Cartesian voxel space.

\begin{figure*}[!t]
    \centering
    \includegraphics[width=0.99\textwidth]{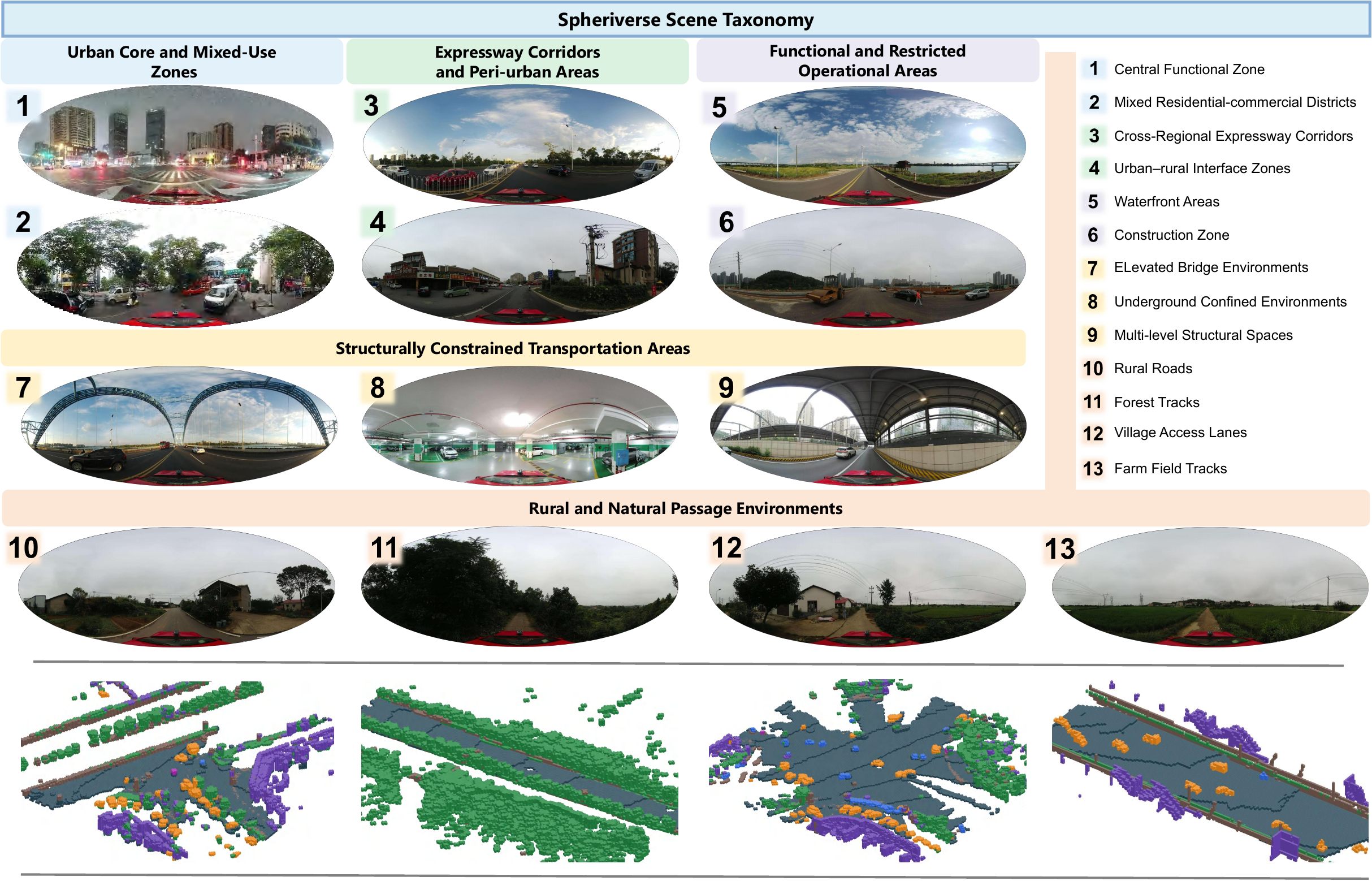} 
    \caption{{Scene categories in Spheriverse.}
    The figure presents five major scene categories and 13 fine-grained categories, with representative spherical images shown for each fine-grained category. Examples 1--13 correspond to the categories listed on the right. Spheriverse covers heterogeneous driving environments with substantial variations in traffic function, environmental context, spatial layout, and visual conditions.
The lower row shows semantic occupancy visualizations of four representative scenes, corresponding from left to right to a construction-zone residential complex in Functional and Restricted Operational Areas, an open expressway corridor in Expressway Corridors and Peri-urban Areas, a dense mixed-use intersection in Urban Core and Mixed-Use Zones, and an elevated bridge in Structurally Constrained Transportation Areas.}
    \label{fig:category}
\end{figure*}

\subsection{Semantic Mapping and 3D Object Detection}~\label{sec2.3}

Semantic mapping focuses on recovering the semantic composition and structural layout of surrounding environments in the Bird's-Eye View (BEV) space. Existing methods can be broadly categorized into rasterized and vectorized representations. 
Rasterized BEV methods, including OneBEV~\cite{onebev}, HDMapNet~\cite{hdmapnet}, and SeqBEV~\cite{bevseq}, encode environments as dense BEV grids and perform pixel-level semantic prediction through geometry-based transformation or learnable feature aggregation. 
In contrast, vectorized BEV methods, such as VectorMapNet~\cite{vectormapnet}, MapTR~\cite{maptr}, and PivotNet~\cite{pivotnet}, represent semantic elements as structured points or polylines, providing compact and topology-aware descriptions of scene structures. Beyond architectural design, recent work also investigates noise-resilient BEV semantic learning with synthetic data generated by driving world models~\cite{li2026nrseg}. 
3D object detection aims to localize and recognize individual objects in 3D space. 
For 3D object detection, geometry-based BEV projection methods, such as PolarBEV~\cite{PolarBEV}, SOLOFusion~\cite{Solofusion}, and MV2DFusion~\cite{MV2DFusion}, exploit camera geometry, depth estimation, and temporal fusion to lift image features into BEV space. 
Query-based BEV methods~\cite{detr3d,bevformer,sparsebev} introduce learnable queries or adaptive BEV representations to aggregate image evidence and predict 3D bounding boxes. 
Although these approaches achieve promising performance in conventional perspective settings, they do not account for the distinctive geometric and visual characteristics of spherical imaging. To characterize the capability of existing methods under spherical observations, we establish benchmarks for semantic mapping and 3D object detection under spherical inputs.

\begin{table}[t!]
\centering
\caption{Data distribution across four daily time periods for 644 sequences:
night (20:00--05:00), dawn (05:00--07:00), daytime (07:00--18:00), and
evening (18:00--20:00). Density denotes the number of samples per hour within
each period, and proportion denotes the percentage of all samples.}
\label{tab:time_density}
\setlength{\tabcolsep}{3pt}
\renewcommand{\arraystretch}{1.18}
\footnotesize
\begin{tabular*}{0.48\textwidth}{@{\extracolsep{\fill}}lcccc@{}}
\toprule
\multirow[c]{2}{*}[-0.65ex]{\textbf{Time Period}}
& \multirow[c]{2}{*}[-0.65ex]{\makecell[c]{\textbf{Duration}\\\textbf{(h)}}}
& \multicolumn{3}{c}{\textbf{Sample Distribution}} \\
\cmidrule(lr){3-5}
& & \textbf{Count}
& \textbf{Density}
& \textbf{Proportion (\%)} \\
\midrule
Night   & 9  & 7{,}300  & 811.11     & 11.34 \\
Dawn    & 2  & 6{,}700  & 3{,}350.00 & 10.40 \\
Daytime & 11 & 31{,}100 & 2{,}827.27 & 48.29 \\
Evening & 2  & 19{,}300 & 9{,}650.00 & 29.97 \\
\bottomrule
\end{tabular*}
\end{table}

\section{Spheriverse Dataset}
\label{sec:meth}

Spherical observations provide complete vertical and horizontal field-of-view coverage, enabling global and coherent visual representations for spatial modeling and scene understanding. 
However, research on spatial understanding under spherical images remains limited. 
To bridge this gap, we introduce Spheriverse, a real-world spherical dataset with rich semantic annotations, comprising $644$ sequences from $13$ geographically and visually diverse regions. 
Equipped with a high-quality spherical camera and a 128-beam LiDAR, Spheriverse provides dense observations to support research on spherical 3D perception.

\subsection{Data Acquisition and Processing}
\textbf{Data Collection:} 
{We integrated a DuxCam M$4$ spherical camera~\cite{panodux_m4} with an image resolution of $5188\times1979$ and a Hesai OT$128$ LiDAR}~\cite{hesai_ot128} as the primary data acquisition sensors. 
The platform is equipped with two positioning sources: a GNSS receiver integrated into the camera and an external GNSS/INS unit. %The camera-integrated GNSS receiver provides an auxiliary global-position
%reference, while the external GNSS/INS unit provides the vehicle's 6-DoF pose. 
The camera intrinsic parameters are calibrated using the Zhang calibration method~\cite{zhang1999flexible}. 
Using the LiDAR coordinate system as the reference, we further calibrate the camera extrinsic parameters~\cite{PanoRingCalib}. 
The entire platform is centrally managed by a domain controller, which utilizes the Precision Time Protocol to ensure accurate clock alignment.

\textbf{Data Annotation and Postprocessing:}
To advance research on semantic spatial perception from spherical observations in real-world environments, Spheriverse encompasses a diverse range of scene types.
Specifically, the collected sequences are organized into five major categories and thirteen fine-grained subcategories according to traffic characteristics, environmental context, and geometric structure, as shown in Fig.~\ref{fig:category}. 
Point cloud annotations are produced through an outsourced annotation pipeline (Stardust AI~\cite{stardustai}), where manual labeling is performed for semantic categories. 
Furthermore, to enhance the structural consistency and standardization of the dataset, we adopt a database-oriented storage architecture, following previous datasets~\cite{Nuscenes,Omnihd}. 

\subsection{Dataset Statistics}
The raw collection contains $89,674$ spherical images.
After temporal alignment and data filtering, the benchmark subset
contains $64,400$ synchronized image--LiDAR pairs organized into
$644$ sequences, each spanning $20$ seconds at $5$\,Hz, following~\cite{Nuscenes,Omnihd,kitti}. 
To facilitate comprehensive and diverse research on spherical perception, we analyze the dataset from the following perspectives.
%{Scene Statistics},  which present the distribution of scene categories during data acquisition;  {Temporal Statistics}, 
% which characterize the distribution of acquisition time; and  {Semantic Statistics}.% which demonstrate the categorical richness of the annotations. 

\textbf{Time Statistics:} 
As shown in Table~\ref{tab:time_density}, the established Spheriverse dataset is collected across four representative time periods, including \emph{night}, \emph{dawn}, \emph{daytime}, and \emph{evening}.
The data cover the entire daily illumination cycle, with daytime images forming the largest proportion, accounting for $48.29\%$ of the dataset. 
Evening scenes contribute $29.97\%$ of the data and exhibit the highest acquisition density. 
Meanwhile, night and dawn periods account for $11.34\%$ and $10.40\%$, respectively, providing valuable samples under low-light and transitional lighting environments.
This temporal distribution indicates that Spheriverse is not limited to standard daytime scenarios but instead captures diverse real-world illumination conditions, establishing a valuable foundation for future research on robust perception across varying visibility and environmental settings. 

\begin{figure*}[!t]
    \centering
    \includegraphics[width=0.96\textwidth]{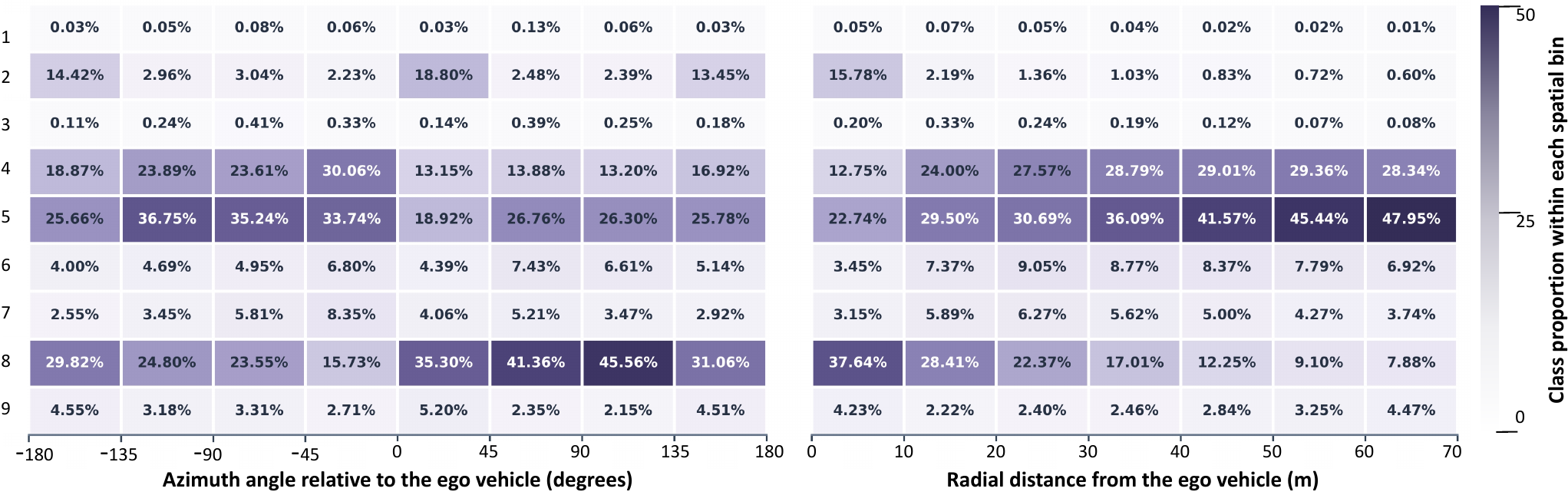} 
    \caption{{Angular and radial class composition of the nine-class taxonomy.} 
    Each column corresponds to an angular or radial bin and is normalized over the nine semantic classes, such that the values in each column sum to $100\%$.
    Each cell represents the proportion of observations within that spatial bin assigned to the corresponding class. 
    Darker colors indicate larger within-bin class proportions. 
    Angular intervals are defined relative to the ego-vehicle heading, which is aligned with the horizontal center of the image, whereas radial intervals are measured from the ego-vehicle origin.
    Rows 1--9 are \emph{1. Pedestrian}, \emph{2. Vehicle}, \emph{3. Cyclist}, \emph{4. Building}, \emph{5. Vegetation}, \emph{6. Pole \& Barrier}, \emph{7. Surface}, \emph{8. Road}, and \emph{9. Others}, respectively.}
    \label{fig:distance}
\end{figure*}

\textbf{Semantic Statistics:} 
Spheriverse provides a three-level semantic annotation hierarchy that organizes scene elements at different levels of granularity.
The annotations cover traffic participants, built structures, and natural environments, including pedestrians, vehicles, buildings, vegetation, and roads.
More detailed categories include temporary buildings, construction vehicles, curbs, and gravel.
Detailed category definitions, hierarchy mappings, and class statistics are provided in the Appendix. Following methods~\cite{onebev,oneocc,PanoRingCalib}, Fig.~\ref{fig:distance} summarizes the representative composition of the nine semantic classes across azimuthal and radial bins centered on the ego vehicle. The left panel partitions the azimuth range from $-180^\circ$ to $180^\circ$ into eight $45^\circ$ intervals, with $0^\circ$ aligned with the ego-vehicle heading. The right panel partitions the radial range from $0$ to $70$\,m into seven $10$\,m intervals. 
Each column is normalized across the nine classes to $100\%$, such that each cell represents the class proportion within one spatial bin. 
Comparing cells vertically within each column, vegetation and surface dominate most bins, with buildings also contributing substantially, whereas pedestrians and cyclists remain consistently sparse. Across azimuthal bins, the proportions of surface, building, and vehicle vary substantially, indicating a direction-dependent scene composition. 
Across radial bins, foreground traffic participants become progressively less prevalent, consistent with stronger occlusion and reduced observability at longer ranges. 
Specifically, the vehicle proportion decreases from $15.78\%$ to $0.60\%$, while pedestrian and cyclist proportions show overall decreases from $0.05\%$ to $0.01\%$ and from $0.20\%$ to $0.08\%$, respectively. Conversely, vegetation increases from $22.74\%$ to $47.95\%$, while buildings rise from $12.75\%$ to approximately $29\%$ and then remain stable at longer ranges. Together, these azimuth- and range-dependent variations reveal a pronounced spatial class prior inherent to driving scenes.

\section{{SphereOcc: Proposed Method}}
Under spherical observations, the geometric correspondence between the angular observation and Cartesian voxel space is non-uniform, creating a cross-space representation gap, giving rise to a range of projection-induced distortions in object appearance, geometry, and spatial relationships. 
{To bridge this gap, we propose SphereOcc, an occupancy framework that couples spherical geometry with semantic evidence to address the cross-space representation gap between angular observations and metric voxels.}

\begin{figure*}[ht]
    \centering
    \includegraphics[width=0.98\textwidth]{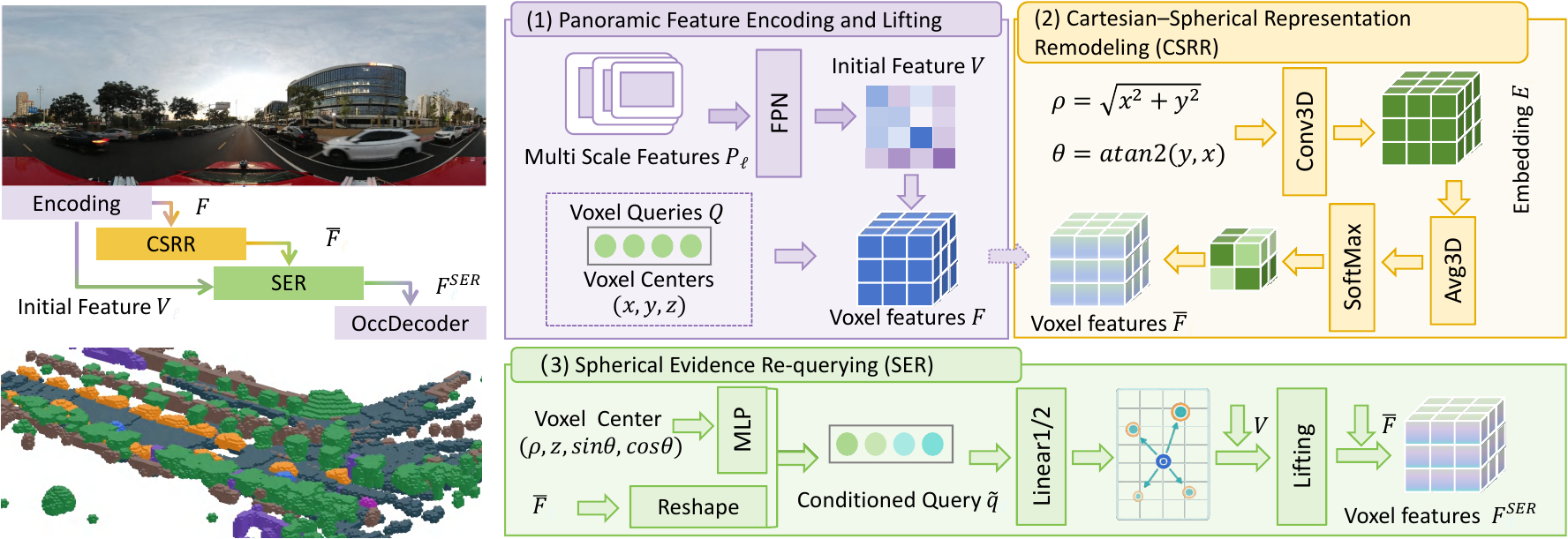} 
    \caption{Overview of SphereOcc for bridging spherical observations and Cartesian voxel representations.
    (1) The spherical image encoder and feature pyramid extract multi-scale features, which are lifted into an initial Cartesian voxel representation using learnable voxel queries.
    (2) CSRR supplements the lifted voxel features with spherical geometric information derived from range-azimuth relations and region-wise importance estimation.
    (3) SER conditions voxel queries on spherical range-height-azimuth geometry and re-queries the source spherical image features using learned sampling offsets and attention weights, thereby supplementing the voxel features with relevant semantic information.
    }
    \label{fig:method}
    \vspace{-1em}
\end{figure*}

\subsection{Problem Formulation}
Given a spatial point 
\(\mathbf{p}_i = (x_i, y_i, z_i)^\top \in \mathbb{R}^3\) 
in the real-world Cartesian coordinate system, spherical imaging maps it onto the spherical coordinate domain as \((\theta_i, \phi_i)\), where
\begin{equation}
    \theta_i = \operatorname{arctan2}(y_i, x_i), \quad
    \phi_i = \arcsin \left( \frac{z_i}{\sqrt{x_i^2 + y_i^2 + z_i^2}} \right).
    \label{eq:spherical_projection}
\end{equation}
Here, \(\mathbf{p}_i\) denotes the \(i\)-th spatial point. 
The variables \(x_i\), \(y_i\), and \(z_i\) represent the Cartesian coordinates of \(\mathbf{p}_i\) along the \(X\)-, \(Y\)-, and \(Z\)-axes, respectively. The angle \(\theta_i \in [-\pi,\pi)\) denotes the horizontal azimuth angle of \(\mathbf{p}_i\), which corresponds to the horizontal position. 
% The angle \(\phi_i \in [-\pi/4,\pi/2]\), \textit{i.e.}, \([-45^\circ,90^\circ]\), denotes the vertical elevation angle of \(\mathbf{p}_i\), which corresponds to the vertical position within the observable vertical field of view of the camera.
The angle $\phi_i \in [-0.815,\pi/2]$, \textit{i.e.}, $[-46.70^\circ,90^\circ]$, denotes the vertical elevation angle of $\mathbf{p}_i$, which corresponds to the vertical position within the observable vertical field of view of the camera. 
Through this spherical projection, the original 3D point \(\mathbf{p}_i\) in Cartesian space is transformed into a 2D angular representation \((\theta_i,\phi_i)\) on the spherical imaging domain.

Furthermore, given the spherical angular coordinate \((\theta_i, \phi_i)\), its corresponding pixel coordinate \((u_i, v_i)\) in the equirectangular projection (ERP) image can be expressed as
\begin{equation}
    u_i = \frac{\theta_i + \pi}{2\pi} W, \quad 
    v_i = \frac{\phi_{\max} - \phi_i}{\phi_{\max} - \phi_{\min}} H,
    \label{eq:erp_projection}
\end{equation}
where \(u_i\) and \(v_i\) denote the horizontal and vertical pixel coordinates in the ERP image, respectively; \(W\) and \(H\) are the ERP image width and height; and \(\phi_{\min}\) and \(\phi_{\max}\) denote the lower and upper bounds of the camera's observation.

Dense occupancy prediction aims to recover a voxelized 3D semantic representation from the ERP image:
\begin{equation}
    \mathcal{S} \in \{0,1,\ldots,C\}^{X \times Y \times Z},
    \label{eq:semantic_occupancy}
\end{equation}
where \(\mathcal{S}\) denotes the 3D semantic occupancy grid, \(X\), \(Y\), and \(Z\) indicate the voxel grid dimensions along the three spatial axes, and \(C\) indicates the number of semantic classes. Each discrete voxel corresponds to a local 3D space unit in the real-world Cartesian system.

\subsection{{Overview of SphereOcc}}
The encoder produces three hierarchical image feature maps
following~\cite{surroundocc}. The feature at level $\ell$ is denoted by:
\begin{equation}
    \bm{X}_{\ell}
    \in
    \mathbb{R}^{C_{\ell}\times H_{\ell}\times W_{\ell}},
    \qquad \ell\in\{0,1,2\},
\end{equation}
where $C_{\ell}$, $H_{\ell}$, and $W_{\ell}$ denote the channel
dimension, height, and width of the feature at level $\ell$,
respectively.

The neck aggregates $\{\bm{X}_{j}\}_{j=0}^{2}$ into multi-scale feature
maps $\{\bm{P}_{\ell}\}_{\ell=0}^{2}$. Each feature map is then projected to
match the channel dimension of its corresponding voxel level:
\begin{equation}
    \bm{V}_{\ell}
    =
    \tau_{\ell}
    \left(
        \bm{P}_{\ell}
    \right)
    \in
    \mathbb{R}^{D_{\ell}\times H_{\ell}\times W_{\ell}},
\end{equation}
where $\tau_{\ell}$ denotes the scale-specific channel projection and
$D_{\ell}$ is the voxel-feature dimension at level $\ell$. The features $\bm{V}_{\ell}$ are first lifted into 3D space
through scale-specific perception transformers:
\begin{equation}
    \bm{F}_{\ell}
    =
    \mathcal{L}_{\ell}
    \left(
        \bm{V}_{\ell},
        \bm{Q}_{\ell}
    \right)
    \in
    \mathbb{R}^{D_{\ell}\times
    H_{\ell}^{\mathrm{v}}\times
    W_{\ell}^{\mathrm{v}}\times
    Z_{\ell}^{\mathrm{v}}},
    \qquad \ell\in\{0,1,2\},
    \label{eq:initial_2d_to_3d_lifting}
\end{equation}
where $\mathcal{L}_{\ell}$ denotes the scale-specific 2D-to-3D
lifting operation, $\bm{Q}_{\ell}$ denotes the learnable voxel
queries, and $D_{\ell}$ denotes the channel dimension of the lifted
voxel feature. Moreover, $H_{\ell}^{\mathrm{v}}$,
$W_{\ell}^{\mathrm{v}}$, and $Z_{\ell}^{\mathrm{v}}$ denote the
spatial dimensions of the voxel grid at level $\ell$. The resulting voxel features are processed by Cartesian--Spherical
Representation Remodeling (CSRR), which encodes {sphere-aligned} {range--azimuth}
relations to form {Cartesian--spherical} voxel features:
\begin{equation}
    \left\{
    \overline{\bm{F}}_{\ell}
    \right\}_{\ell=0}^{2}
    =
    \mathcal{A}
    \left(
    \left\{\bm{F}_{\ell}\right\}_{\ell=0}^{2}
    \right), \overline{\bm{F}}_{\ell}
    \in
    \mathbb{R}^{D_{\ell}\times
    H_{\ell}^{\mathrm{v}}\times
    W_{\ell}^{\mathrm{v}}\times
    Z_{\ell}^{\mathrm{v}}}.
\end{equation}
$\mathcal{A}$ denotes CSRR in Sec.~\ref{sec:CSRR}.

Building on these features, Spherical Evidence Re-querying (SER) complements
each voxel feature with relevant semantic evidence from the source features:
\begin{equation}
    \begin{aligned}
        \left\{\bm{F}_{\ell}^{\mathrm{SER}}\right\}_{\ell=0}^{2}
        =
        \mathcal{R}\Big(
            \left\{\overline{\bm{F}}_{\ell}\right\}_{\ell=0}^{2},
            {
            \left\{\bm{V}_{\ell},\bm{p}_{\ell}\right\}_{\ell=0}^{2}}
        \Big).
    \end{aligned}
\end{equation}
$\mathcal{R}$ denotes SER in Sec.~\ref{sec:SER}, and $\bm{p}_{\ell}$ denotes the voxel-to-ERP reference points cached during the initial lifting.
SER preserves the dimensions of all three voxel levels.
Finally, the refined multi-scale voxel features are fused by the occupancy decoder, following~\cite{surroundocc}, producing four hierarchical occupancy predictions:
\begin{equation}
    \left\{
    \bm{O}^{(s)}
    \right\}_{s=0}^{3}
    =
    \mathcal{D}
    \left(
        \left\{
        \bm{F}_{\ell}^{\mathrm{SER}}
        \right\}_{\ell=0}^{2}
    \right),
\end{equation}
where $\bm{O}^{(s)}$
% \begin{equation}
%     \bm{O}^{(s)}
%     \in
%     \mathbb{R}^{{(C+1)}\times
%     H_{s}^{\mathrm{o}}\times
%     W_{s}^{\mathrm{o}}\times
%     Z_{s}^{\mathrm{o}}}
% \end{equation}
denotes the occupancy prediction at decoder stage $s$. All four
predictions are supervised during training, while only the high-resolution prediction $\bm{O}^{(3)}$ is used during inference.

All four decoder outputs are supervised; the loss is:
\begin{equation}
    \mathcal{L}_{\mathrm{occ}}
    =
    \sum_{s=0}^{3}
    \lambda_s
    \left(
        \mathcal{L}_{\mathrm{ce}}^{(s)}
        +
        \mathcal{L}_{\mathrm{sem}}^{(s)}
        +
        \mathcal{L}_{\mathrm{geo}}^{(s)}
    \right),
\end{equation}
where $s$ denotes the decoder stage and $\lambda_s$ is its supervision
weight. $\mathcal{L}_{\mathrm{ce}}$ denotes the cross-entropy loss,
$\mathcal{L}_{\mathrm{sem}}$ denotes the semantic loss, and
$\mathcal{L}_{\mathrm{geo}}$ denotes the geometric loss.

\subsection{{Cartesian-Spherical Representation Remodeling}} \label{sec:CSRR}
Initial lifting transfers image features to Cartesian voxels, but these features lack an explicit mechanism for connecting voxels along the spherical observation geometry. 

Given the multi-scale voxel features
$\{\bm{F}_{\ell}\}_{\ell=0}^{2}$, CSRR embeds sphere-aligned range–azimuth relations into the Cartesian representation, yielding {Cartesian–spherical} voxel features.

Specifically, given a voxel center $(x,y,z)$, we combine its Cartesian coordinates with the
horizontal range $\rho$ and the spherical azimuth
$\theta$. The radial distance and spherical azimuth are computed as:
\begin{equation}
    \rho
    =
    \sqrt{x^2+y^2},
    \qquad
    \theta
    =
    \operatorname{atan2}(y,x),
\end{equation}
where $z$ denotes the vertical height. We omit spherical elevation because it is already determined by the retained
$z$ and $\rho$, making it a redundant rather than independent geometric cue.

The resulting coordinate descriptor is defined as:
\begin{equation}
    \bm{c}
    =
    \left[
        x,
        y,
        \sin\theta,
        \cos\theta,
        z,
        \rho
    \right].
    \label{eq:cartesian_spherical_descriptor}
\end{equation}  
At scale $\ell$, the descriptors of all voxels are stacked into
$\bm{C}_{\ell}\in\mathbb{R}^{6\times H_{\ell}^{\mathrm{v}}\times
W_{\ell}^{\mathrm{v}}\times Z_{\ell}^{\mathrm{v}}}$ and embedded as:
\begin{equation}
    \bm{E}_{\ell}
    =
    \phi_{\ell}
    \left(
        \bm{C}_{\ell}
    \right)
    \in
    \mathbb{R}^{D_{\mathrm{c}}\times
    H_{\ell}^{\mathrm{v}}\times
    W_{\ell}^{\mathrm{v}}\times
    Z_{\ell}^{\mathrm{v}}},
\end{equation}
where $\phi_{\ell}$ denotes a $1\times1\times1$ convolution followed by ReLU,
and $D_{\mathrm{c}}$ denotes the coordinate-embedding dimension.

To estimate geometry-guided voxel importance, the lifted voxel features
$\bm{F}_{\ell}$ are concatenated with the coordinate embedding
$\bm{E}_{\ell}$. After that, a lightweight saliency estimator produces a voxel-wise score map $\bm{S}_{\ell}$:

\begin{equation}
    \bm{S}_{\ell}
    =
    \psi_{\ell}
    \left(
        \left[
            \bm{F}_{\ell};
            \bm{E}_{\ell}
        \right]
    \right)
    \in
    \mathbb{R}^{1\times
    H_{\ell}^{\mathrm{v}}\times
    W_{\ell}^{\mathrm{v}}\times
    Z_{\ell}^{\mathrm{v}}},
\end{equation}
where $\psi_{\ell}$ is a 3D convolution that maps the fused feature to a scalar importance score for each voxel. 
The voxel-wise scores are aggregated into non-overlapping regions using three-dimensional average pooling to avoid fragmented voxel-level prioritization and promote coherent remodeling:
\begin{equation}
    \bm{R}_{\ell}
    =
    \operatorname{AvgPool}_{\bm{r}}
    \left(
        \bm{S}_{\ell}
    \right)
    \in
    \mathbb{R}^{1\times
    \widehat{H}_{\ell}\times
    \widehat{W}_{\ell}\times
    \widehat{Z}_{\ell}},
\end{equation}
where
\begin{equation}
    \bm{r}=(r_H,r_W,r_Z)
\end{equation}
denotes the region size. The total number of pooled regions at scale $\ell$ is:
\begin{equation}
    N_{\ell}
    =
    \widehat{H}_{\ell}
    \widehat{W}_{\ell}
    \widehat{Z}_{\ell}.
\end{equation}
After that, we apply a Softmax over the pooled regions and restore the resulting weights
to the original voxel resolution:
\begin{equation}
    \begin{aligned}
        \bm{A}_{\ell}
        &=
        \operatorname{Softmax}_{\mathrm{reg}}
        \left(
            \bm{R}_{\ell}
        \right),\\
        \bm{G}_{\ell}
        &=
        N_{\ell}
        \operatorname{Up}_{\mathrm{nearest}}
        \left(
            \bm{A}_{\ell}
        \right).
    \end{aligned}
\end{equation}
Here, $\operatorname{Softmax}_{\mathrm{reg}}$ assigns relative weights across
the $N_{\ell}$ regions, and the factor
$N_{\ell}$ compensates for the scale reduction introduced by
this normalization.

\begin{table*}[ht]
\centering
\caption{
Overall and scene-wise semantic occupancy results on Spheriverse.
mIoU, GeoIoU, and class-wise IoU are reported in \%.
Light orange and light blue indicate the best and second-best results, respectively; ties are highlighted equally, and higher values indicate better performance.
TPVFo., Surro., OccDe., MonoS., OccDA., Quadr., ProDA., Proto., Gauss., and BEVFo. denote TPVFormer~\cite{TPVFormer}, SurroundOcc~\cite{surroundocc}, OccDepth~\cite{occdepth}, MonoScene~\cite{monoscene}, OccDepth~\cite{occdepth} with Depth Anything~\cite{depth}, QuadricFormer~\cite{quadricformer}, ProtoOcc~\cite{protoocc} with Depth Anything~\cite{depth}, ProtoOcc~\cite{protoocc}, GaussianFormer~\cite{gaussianformer}, and BEVFormer~\cite{bevformer}, respectively.
}
\label{tab:occ_complete_results}

\definecolor{nperson}{RGB}{156,39,176}
\definecolor{nvehicle}{RGB}{255,167,38}
\definecolor{nbike}{RGB}{66,133,244}
\definecolor{nbuilding}{RGB}{126,87,194}
\definecolor{nvegetation}{RGB}{67,160,71}
\definecolor{npillar}{RGB}{141,110,99}
\definecolor{nroad}{RGB}{84,110,122}
\definecolor{nsurface}{RGB}{176,190,197}
\definecolor{nothers}{RGB}{236,64,122}
\definecolor{occsectioncolor}{RGB}{247,248,250}
\newcommand{\occMethodHead}[1]{\rotatebox{55}{\strut #1}}
\newcommand{\occOursMethodHead}[1]{\rotatebox{55}{\strut #1}}
\newcommand{\occClassMetric}[2]{%
  \raisebox{-0.2ex}{\textcolor{#1}{{\Large$\bullet$}}}\,#2%
}
\definecolor{occbestcolor}{RGB}{255,232,204}
\definecolor{occsecondcolor}{RGB}{225,240,255}
\newcommand{\occbest}[1]{\cellcolor{occbestcolor}\textbf{#1}}
\newcommand{\occsecond}[1]{\cellcolor{occsecondcolor}\underline{#1}}

\setlength{\tabcolsep}{1.5pt}
\renewcommand{\arraystretch}{1.08}
\footnotesize

\begin{tabular*}{0.98\textwidth}{@{\extracolsep{\fill}}ll|*{12}{c}@{}}
\toprule
\multicolumn{1}{c}{\multirow[c]{2}{*}[-8ex]{\textbf{Group}}}
&
\multicolumn{1}{c|}{\multirow[c]{2}{*}[-8ex]{\textbf{Metric}}}
&
\multicolumn{12}{c}{\textbf{Methods}}
\\
\cmidrule(lr){3-14}
&
&
\occMethodHead{TPVFo.}
&
\occMethodHead{Surro.}
&
\occMethodHead{OccDe.}
&
\occMethodHead{MonoS.}
&
\occMethodHead{OccDA.}
&
\occMethodHead{Quadr.}
&
\occMethodHead{ProDA.}
&
\occMethodHead{Proto.}
&
\occMethodHead{CoTR}
&
\occMethodHead{Gauss.}
&
\occMethodHead{BEVFo.}
&
\occOursMethodHead{\textbf{Ours}}
\\
\specialrule{0.10em}{0.7ex}{0.3ex}

\rowcolor{occsectioncolor}
\multicolumn{14}{c}{%
  \textbf{Overall Semantic Occupancy Results}
}
\\
\midrule

% =========================================================
% Overall semantic occupancy
% =========================================================
\multicolumn{2}{l|}{mIoU $\uparrow$}
& \occsecond{12.21} & 12.05 & 11.63 & 11.44 & 11.42 & 11.09
& 11.07 & 10.93 & 9.72 & 9.16 & 8.22 & \occbest{13.91} \\

\multicolumn{2}{l|}{GeoIoU $\uparrow$}
& 22.35 & \occsecond{22.55} & 21.59 & 21.68 & 21.97 & 20.76
& 20.77 & 21.02 & 19.23 & 18.04 & 18.37 & \occbest{24.65} \\

\cmidrule(lr){1-14}

\multicolumn{2}{l|}{\occClassMetric{nperson}{Person} $\uparrow$}
& \occbest{4.07} & 1.65 & 3.21 & 2.49 & 2.41 & 0.00
& 2.96 & \occsecond{3.30} & 1.97 & 0.00 & 0.20 & 3.02 \\

\multicolumn{2}{l|}{\occClassMetric{nvehicle}{Vehicle} $\uparrow$}
& 12.88 & \occsecond{13.97} & 11.27 & 10.90 & 12.04 & 11.09
& 12.36 & 12.12 & 11.47 & 9.05 & 5.36 & \occbest{15.71} \\

\multicolumn{2}{l|}{\occClassMetric{nbike}{Bike} $\uparrow$}
& \occbest{8.54} & 5.16 & 7.25 & 6.95 & 7.22 & 4.51
& \occsecond{7.84} & 6.52 & 6.11 & 4.62 & 1.42 & 7.50 \\

\multicolumn{2}{l|}{\occClassMetric{nbuilding}{Building} $\uparrow$}
& 8.93 & \occsecond{9.40} & 8.49 & 8.18 & 7.85 & 8.09
& 7.29 & 7.18 & 6.71 & 5.82 & 6.61 & \occbest{10.25} \\

\multicolumn{2}{l|}{\occClassMetric{nvegetation}{Vegetation} $\uparrow$}
& 14.00 & \occsecond{14.08} & 13.26 & 13.55 & 13.44 & 12.45
& 13.18 & 13.08 & 12.18 & 9.26 & 10.09 & \occbest{15.57} \\

\multicolumn{2}{l|}{\occClassMetric{npillar}{Pillar} $\uparrow$}
& 12.34 & 12.57 & 10.67 & 10.93 & 10.68 & \occsecond{12.59}
& 8.94 & 8.45 & 7.33 & 9.21 & 10.12 & \occbest{16.60} \\

\multicolumn{2}{l|}{\occClassMetric{nroad}{Road} $\uparrow$}
& 34.92 & 36.58 & 35.74 & 36.62 & 35.06 & \occsecond{37.62}
& 35.04 & 34.47 & 29.84 & 34.96 & 28.80 & \occbest{39.99} \\

\multicolumn{2}{l|}{\occClassMetric{nsurface}{Surface} $\uparrow$}
& 12.01 & 11.99 & \occsecond{12.62} & 11.61 & 12.26 & 11.91
& 10.06 & 11.11 & 9.85 & 9.22 & 10.05 & \occbest{13.01} \\

\multicolumn{2}{l|}{\occClassMetric{nothers}{Others} $\uparrow$}
& 2.21 & \occsecond{3.07} & 2.16 & 1.75 & 1.83 & 1.54
& 1.95 & 2.13 & 1.97 & 0.32 & 1.33 & \occbest{3.53} \\

% =========================================================
% Separator
% =========================================================
\specialrule{0.10em}{0.7ex}{0.3ex}
\rowcolor{occsectioncolor}
\multicolumn{14}{c}{%
  \textbf{Scene-wise Semantic Occupancy Results}
}
\\
\midrule

% =========================================================
% Expressway
% =========================================================
\multirow[c]{2}{*}{\textbf{Expre.}}
& mIoU $\uparrow$
& 10.51 & \occsecond{11.29} & 10.43 & 10.35 & 10.23 & 10.05
& 9.84 & 9.78 & 8.41 & 7.58 & 7.34 & \occbest{12.82} \\

& GeoIoU $\uparrow$
& 25.00 & \occsecond{25.36} & 24.41 & 24.46 & 24.77 & 23.66
& 22.98 & 23.45 & 21.37 & 19.48 & 20.72 & \occbest{27.02} \\

\midrule

% =========================================================
% Functional
% =========================================================
\multirow[c]{2}{*}{\textbf{Funct.}}
& mIoU $\uparrow$
& 8.33 & \occsecond{8.38} & 7.92 & 8.12 & 7.47 & 8.02
& 7.43 & 7.82 & 6.71 & 7.16 & 6.13 & \occbest{10.17} \\

& GeoIoU $\uparrow$
& \occsecond{23.72} & 23.39 & 23.14 & 23.02 & 23.38 & 22.80
& 22.37 & 22.95 & 20.29 & 20.62 & 21.11 & \occbest{25.90} \\

\midrule

% =========================================================
% Rural
% =========================================================
\multirow[c]{2}{*}{\textbf{Rural}}
& mIoU $\uparrow$
& 10.13 & \occsecond{10.22} & 9.62 & 9.79 & 9.78 & 8.94
& 9.93 & 9.71 & 9.10 & 7.57 & 7.34 & \occbest{11.32} \\

& GeoIoU $\uparrow$
& 19.78 & \occsecond{20.23} & 18.82 & 18.85 & 19.23 & 17.56
& 18.53 & 18.84 & 17.61 & 15.23 & 16.18 & \occbest{22.43} \\

\midrule

% =========================================================
% Structural
% =========================================================
\multirow[c]{2}{*}{\textbf{Struc.}}
& mIoU $\uparrow$
& \occsecond{13.66} & \occsecond{13.66} & 12.39 & 12.34 & 12.42 & 12.70
& 12.67 & 11.86 & 9.47 & 10.50 & 8.42 & \occbest{16.60} \\

& GeoIoU $\uparrow$
& 32.23 & \occsecond{34.79} & 32.38 & 32.18 & 32.51 & 33.07
& 28.72 & 28.39 & 24.60 & 29.67 & 26.00 & \occbest{39.04} \\

\midrule

% =========================================================
% Urban
% =========================================================
\multirow[c]{2}{*}{\textbf{Urban}}
& mIoU $\uparrow$
& 8.05 & \occsecond{8.31} & 7.52 & 7.49 & 7.37 & 6.73
& 7.82 & 7.55 & 6.69 & 5.87 & 4.45 & \occbest{9.64} \\

& GeoIoU $\uparrow$
& \occsecond{17.49} & \occsecond{17.49} & 16.49 & 16.74 & 16.91 & 15.31
& 17.01 & 17.10 & 15.99 & 13.95 & 13.48 & \occbest{19.32} \\

\bottomrule
\end{tabular*}

\end{table*}

Finally,  the Cartesian–spherical voxel features are constructed through a spatially gated channel residual.
At scale $\ell$, a learnable residual
$\bm{b}_{\ell}\in\mathbb{R}^{D_{\ell}\times1\times1\times1}$ is broadcast
over the spatial dimensions, while the scene-conditioned gate
$\bm{G}_{\ell}$ determines its spatial strength:
\begin{equation}
    \overline{\bm{F}}_{\ell}
    =
    \bm{F}_{\ell}
    +
    \bm{G}_{\ell}
    \odot
    \bm{b}_{\ell},
\end{equation}
where $\odot$ denotes element-wise multiplication with broadcasting along the
channel dimension. This factorization separates spatial prioritization from
channel modulation: $\bm{G}_{\ell}$ controls where and how strongly the update
is applied, whereas $\bm{b}_{\ell}$ controls the channel-wise adjustment. The
channel residual is initialized to zero and learned jointly with the regional
importance estimator, preserving the original voxel features at initialization.
The resulting features
$\{\overline{\bm{F}}_{\ell}\}_{\ell=0}^{2}$ are subsequently passed to
spherical
evidence re-querying.

\subsection{{Spherical Evidence Re-querying}} \label{sec:SER}
Although CSRR enriches the lifted voxel features with explicit spherical range-azimuth geometry, the resulting Cartesian-spherical voxel features do not explicitly model their adaptive semantic correspondence with the source spherical image features. SER addresses this limitation by re-querying the source features to retrieve relevant semantic evidence for each voxel feature.

At scale $\ell$, the Cartesian–spherical voxel feature volume $\overline{\bm{F}}_{\ell}$ is flattened
into content queries $\{\bm{q}_{\ell,n}\}_{n=1}^{N_{\ell}^{\mathrm{v}}}$, where
$N_{\ell}^{\mathrm{v}}=H_{\ell}^{\mathrm{v}}W_{\ell}^{\mathrm{v}}Z_{\ell}^{\mathrm{v}}$.
For each query, SER derives an RTZ positional descriptor from its voxel-grid
location:
\begin{equation}
    \bm{t}_{\ell,n}
    =
    \left[
        {\rho_{\ell,n}},
        {z_{\ell,n}},
        {\sin\theta_{\ell,n}},
        {\cos\theta_{\ell,n}}
    \right].
\end{equation}
Here, $\bm{t}_{\ell,n}\in\mathbb{R}^{4}$ denotes the positional descriptor of
query $n$ at scale $\ell$. Furthermore, SER uses a
two-layer MLP $\eta_{\ell}$ with a SiLU activation to embed the RTZ geometry
and condition the evidence query. The conditioned query $\widetilde{\bm{q}}_{\ell,n}$ is used to predict
deformable sampling offsets and attention weights, as shown in the following:
\begin{equation}
    \widetilde{\bm{q}}_{\ell,n}
    =
    \bm{q}_{\ell,n}
    +
    \eta_{\ell}
    \left(
        \bm{t}_{\ell,n}
    \right).
\end{equation}
 The Cartesian–spherical voxel feature corresponding to query $n$ 
($\overline{\bm{F}}_{\ell,n}$) remains unchanged until the retrieved image evidence
is incorporated through the final residual update.

After that, let $\bm{p}_{\ell,n}\in[0,1]^2$ be the reference point of voxel $n$ cached
during the initial 2D-to-3D lifting. For attention head $h$ and sampling point $k$, the RTZ-conditioned query
predicts a two-dimensional offset and its attention weight:
\begin{equation}
    \begin{aligned}
        \Delta\bm{p}_{\ell,n,h,k}
        &=
        \bm{W}_{\ell}^{\Delta}
        \widetilde{\bm{q}}_{\ell,n},\\
        a_{\ell,n,h,k}
        &=
        \operatorname{Softmax}_{k}
        \left(
            \bm{W}_{\ell}^{a}
            \widetilde{\bm{q}}_{\ell,n}
        \right).
    \end{aligned}
\end{equation}
The sampling location is:
\begin{equation}
    \bm{s}_{\ell,n,h,k}
    =
    \bm{p}_{\ell,n}
    +
    \Delta\bm{p}_{\ell,n,h,k}
    \oslash
    \left[
        {W_{\ell}},
        {H_{\ell}}
    \right],
\end{equation}
where $H_{\ell}$ and $W_{\ell}$ are the height and
width of the source feature map $\bm{V}_{\ell}$, and $\oslash$ denotes element-wise division. Thus,
the cached projection provides a geometrically valid anchor, while the learned
offsets adapt the re-query to the current voxel feature and its RTZ geometry.

\begin{figure*}[!t]
    \centering
    \includegraphics[width=0.98\textwidth]{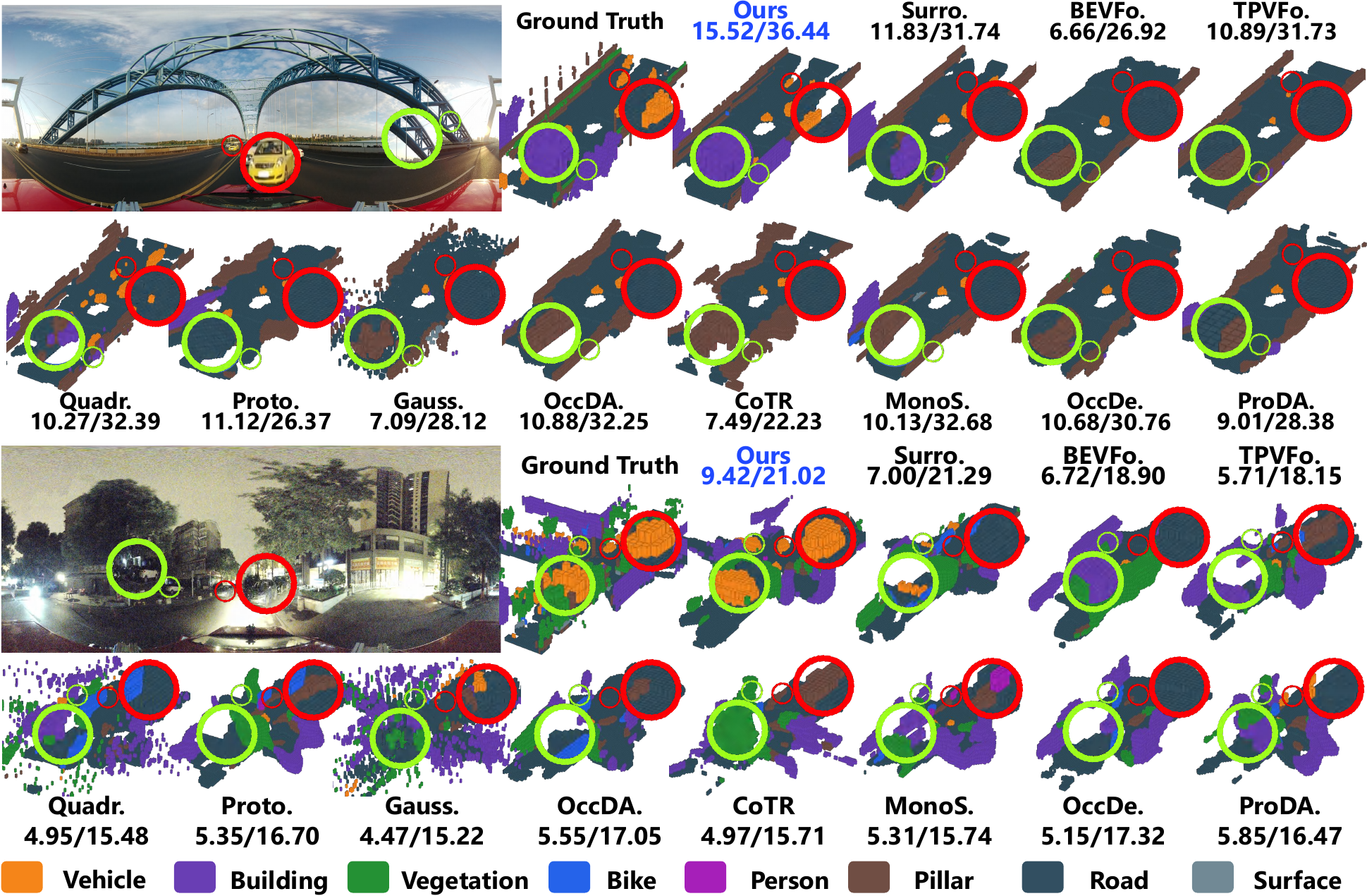} 
\caption{
Qualitative comparison of semantic occupancy prediction on Spheriverse. The mIoU and GeoIoU scores (in \%) are reported above each prediction.
Surro., BEVFo., TPVFo., Quadr., Proto., Gauss., CoTR, MonoS., and OccDe. denote SurroundOcc~\cite{surroundocc}, BEVFormer~\cite{bevformer}, TPVFormer~\cite{TPVFormer}, QuadricFormer~\cite{quadricformer}, ProtoOcc~\cite{protoocc}, GaussianFormer~\cite{gaussianformer}, Compact Occupancy Transformer~\cite{cotr}, MonoScene~\cite{monoscene}, and OccDepth~\cite{occdepth}, respectively.
OccDA. and ProDA. denote OccDepth and ProtoOcc augmented with Depth Anything~\cite{depth}, respectively.
Colors represent the semantic classes shown in the bottom legend: Vehicle, Building, Vegetation, Bike, Person, Pillar, Road, and Surface.
}
    \label{fig:Occ}
    \vspace{-1em}
\end{figure*}

The sampled evidence is aggregated across sampling points and attention
heads:
\begin{equation}
    \bm{e}_{\ell,n}
    =
    \bm{W}_{\ell}^{o}
    \left(
        \mathop{\Vert}_{h=1}^{N_{\mathrm{h}}}
        \sum_{k=1}^{K_{\ell}}
        a_{\ell,n,h,k}
        \mathcal{B}
        \left(
            \bm{W}_{\ell}^{v}\bm{V}_{\ell},
            \bm{s}_{\ell,n,h,k}
        \right)
    \right),
\end{equation}
where $\mathcal{B}$ denotes bilinear sampling and $\Vert$ denotes head-wise
concatenation. We use $N_{\mathrm{h}}=8$ attention heads and
$K_{\ell}\in\{2,4,8\}$ sampling points from fine to coarse scales, following SurroundOcc~\cite{surroundocc}. Finally,
the retrieved evidence is written back through a residual update:
\begin{equation}
    \bm{F}_{\ell,n}^{\mathrm{SER}}
    =
    \overline{\bm{F}}_{\ell,n}
    +
    \bm{e}_{\ell,n}.
\end{equation}
By reusing the lifting reference points and learning only local deformable
offsets around them, SER re-queries relevant semantic evidence for Cartesian–spherical voxel features. The resulting
multi-scale voxel representations are subsequently forwarded to the
occupancy decoder.

\section{Benchmark Experiments}
\label{sec:bench}
All methods are evaluated under a unified experimental protocol. The input spherical images are resized to $2480 \times 512$. 
All models are trained on four NVIDIA RTX 3090 GPUs for $28$ epochs. 
For each benchmark, we report the overall performance across all scenes, as well as the performance in each scene type, enabling a comprehensive evaluation of the model in heterogeneous environments. Unless otherwise noted, all methods are reproduced from their official configurations with only minimal dataset-specific adaptations. Specifically, we standardize the training schedule, input image size, and voxelization, while retaining each method's core representation, task-specific loss, and remaining hyperparameters. The baseline method of SphereOcc is SurroundOcc~\cite{surroundocc}.

\begin{table*}[ht!]
\centering
\caption{
Overall and scene-wise BEV semantic mapping results on Spheriverse.
Class-wise IoU and mIoU are reported in \%, and higher values indicate better performance.
Scene-wise results are obtained using the globally selected checkpoint and class-specific thresholds.
OneBE., HDMap., Pivot., Spars., Vecto., BEVFo., PETRv., SeqBE., MapTR, and TPVFo. denote OneBEV~\cite{onebev}, HDMapNet~\cite{hdmapnet}, PivotNet~\cite{pivotnet}, SparseBEV~\cite{sparsebev}, VectorMapNet~\cite{vectormapnet}, BEVFormer~\cite{bevformer}, PETRv2~\cite{petr}, SeqBEV~\cite{bevseq}, MapTR~\cite{maptr}, and TPVFormer~\cite{TPVFormer}, respectively.
}
\label{tab:bev_global_scene}

\begingroup

\definecolor{bevdynamic}{RGB}{255,167,38}
\definecolor{bevbuilding}{RGB}{126,87,194}
\definecolor{bevvegetation}{RGB}{67,160,71}
\definecolor{bevpillar}{RGB}{141,110,99}
\definecolor{bevroad}{RGB}{84,110,122}
\definecolor{bevsurface}{RGB}{176,190,197}
\definecolor{bevother}{RGB}{236,64,122}
\definecolor{bevsectioncolor}{RGB}{247,248,250}

\newcommand{\bevMethodHead}[1]{%
  \rotatebox{55}{\strut #1}%
}

\newcommand{\bevClassMetric}[2]{%
  \raisebox{-0.2ex}{%
    \textcolor{#1}{{\Large$\bullet$}}%
  }\thinspace #2%
}

\setlength{\tabcolsep}{2.2pt}
\renewcommand{\arraystretch}{1.08}
\normalsize

\begin{tabular*}{0.96\textwidth}
{@{\extracolsep{\fill}}l|*{10}{c}@{}}
\toprule

\multicolumn{1}{c|}{
  \multirow[c]{2}{*}[-7ex]{\textbf{Metric}}
}
&
\multicolumn{10}{c}{\textbf{Methods}}
\\

\cmidrule(lr){2-11}

&
\bevMethodHead{OneBE.}
&
\bevMethodHead{HDMap.}
&
\bevMethodHead{Pivot.}
&
\bevMethodHead{Spars.}
&
\bevMethodHead{Vecto.}
&
\bevMethodHead{BEVFo.}
&
\bevMethodHead{PETRv.}
&
\bevMethodHead{SeqBE.}
&
\bevMethodHead{MapTR}
&
\bevMethodHead{TPVFo.}
\\

\specialrule{0.10em}{0.7ex}{0.3ex}

\rowcolor{bevsectioncolor}
\multicolumn{11}{c}{
  \textbf{Overall BEV Semantic Mapping Results}
}
\\

\midrule

\bevClassMetric{bevdynamic}{Participant} $\uparrow$
& 18.84
& 10.50
& 11.42
& 11.27
& 10.27
& 10.65
& 8.80
& 7.44
& 8.73
& 10.03
\\

\bevClassMetric{bevbuilding}{Building} $\uparrow$
& 19.23
& 19.80
& 17.01
& 16.73
& 16.04
& 14.26
& 15.87
& 12.43
& 11.80
& 11.85
\\

\bevClassMetric{bevvegetation}{Vegetation} $\uparrow$
& 24.76
& 23.75
& 23.47
& 22.68
& 22.35
& 21.20
& 21.65
& 20.63
& 20.70
& 18.35
\\

\bevClassMetric{bevpillar}{Pillar} $\uparrow$
& 18.51
& 17.03
& 15.75
& 16.36
& 15.15
& 13.78
& 12.76
& 11.69
& 9.98
& 8.44
\\

\bevClassMetric{bevroad}{Road} $\uparrow$
& 52.53
& 48.98
& 47.23
& 46.94
& 47.53
& 45.51
& 44.49
& 40.66
& 40.76
& 39.70
\\

\bevClassMetric{bevsurface}{Surface} $\uparrow$
& 19.67
& 19.00
& 19.21
& 18.33
& 19.61
& 17.73
& 18.88
& 18.30
& 16.12
& 17.51
\\

\bevClassMetric{bevother}{Others} $\uparrow$
& 5.05
& 5.89
& 5.43
& 5.01
& 4.74
& 4.08
& 4.68
& 3.48
& 2.73
& 3.61
\\

\cmidrule(lr){1-11}

mIoU $\uparrow$
& 22.66
& 20.71
& 19.93
& 19.62
& 19.38
& 18.17
& 18.16
& 16.37
& 15.83
& 15.64
\\

\specialrule{0.10em}{0.7ex}{0.3ex}

\rowcolor{bevsectioncolor}
\multicolumn{11}{c}{
  \textbf{Scene-wise BEV Semantic Mapping Results}
}
\\

\midrule

Expressway
& 20.67
& 19.57
& 18.58
& 17.81
& 17.93
& 16.81
& 16.99
& 14.55
& 13.70
& 11.24
\\

Functional
& 16.69
& 15.25
& 13.95
& 14.48
& 14.56
& 13.47
& 13.68
& 10.11
& 11.97
& 10.83
\\

Rural
& 21.57
& 19.49
& 19.26
& 18.60
& 18.47
& 16.86
& 16.65
& 15.61
& 16.27
& 16.79
\\

Structural
& 23.15
& 20.28
& 18.58
& 18.55
& 18.76
& 17.58
& 17.48
& 15.28
& 13.15
& 12.55
\\

Urban
& 14.60
& 13.45
& 12.75
& 12.73
& 12.46
& 12.52
& 12.14
& 9.53
& 10.06
& 10.99
\\

\bottomrule
\end{tabular*}

\endgroup
\end{table*}

\subsection{Dense Occupancy Prediction}
\subsubsection{Evaluation Metrics and Implementation Details}
\textbf{Evaluation Metrics:} 
3D occupancy prediction serves as a fundamental task for holistic scene understanding. Therefore, we establish an evaluation that jointly assesses semantic segmentation performance and scene reconstruction quality. 
Following existing works~\cite{surroundocc,gaussianformer,Quadreamer}, we report three evaluation metrics: class-wise IoU, mIoU, and GeoIoU. 
Class-wise IoU evaluates the semantic prediction accuracy for each category, while mIoU averages IoU across all semantic categories. GeoIoU evaluates class-agnostic geometric reconstruction by measuring the spatial overlap between the predicted and ground-truth occupied regions. 

\textbf{Implementation Details:} The occupancy volume is represented as a voxel grid of size $200 \times 200 \times 16$ at a voxel resolution of $0.5$\,m. 
All evaluations use the nine primary semantic classes. 
We reproduce representative camera-based occupancy methods from three paradigms.
This includes MonoScene~\cite{monoscene}, OccDepth~\cite{occdepth}, and COTR~\cite{cotr}. The query-based group includes BEVFormer~\cite{bevformer}, TPVFormer~\cite{TPVFormer}, SurroundOcc~\cite{surroundocc}, and ProtoOcc~\cite{protoocc}. 
The Gaussian-based methods include GaussianFormer~\cite{gaussianformer} and QuadricFormer~\cite{quadricformer}. We additionally evaluate OccDepth and ProtoOcc with a frozen Depth Anything prior~\cite{depth}.

\textbf{Results and Analyses:} As shown in Table~\ref{tab:occ_complete_results}, SphereOcc achieves the best overall performance, reaching \(13.91\%\) mIoU and \(24.65\%\) GeoIoU.
For mIoU, SphereOcc surpasses TPVFormer~\cite{TPVFormer}, the strongest prior method for this metric (\(13.91\%\) vs. \(12.21\%\)).
For GeoIoU, it outperforms SurroundOcc~\cite{surroundocc}, the strongest prior method for this metric (\(24.65\%\) vs. \(22.55\%\)).
These results correspond to absolute gains of \(1.70\) and \(2.10\) percentage points, or relative improvements of \(13.9\%\) and \(9.3\%\), respectively.
SphereOcc also ranks first in both metrics across all five scene subsets and achieves the best class-wise IoU in seven of nine semantic categories.
Although it does not rank first for Person and Bike, it improves their IoUs over SurroundOcc by \(83.0\%\) (\(3.02\%\) vs. \(1.65\%\)) and \(45.3\%\) (\(7.50\%\) vs. \(5.16\%\)), respectively.
Among the seven leading categories, particularly clear gains over SurroundOcc are observed for Pillar (\(16.60\%\) vs. \(12.57\%\)) and Road (\(39.99\%\) vs. \(36.58\%\)). These gains span thin structures, continuous surfaces, and small traffic participants, consistent with the complementary roles of CSRR and SER in enriching voxel features with spherical geometry and relevant semantic evidence. The qualitative results in Fig.~\ref{fig:Occ} further corroborate these quantitative improvements. In the challenging example, vehicles and vegetation are spatially intertwined, resulting in ambiguous boundaries and complex local geometry. By linking Cartesian voxel features with spherical observations at both geometric and semantic levels, SphereOcc bridges the cross-space representation gap and enables more complete and accurate reconstruction of regions where vehicles and vegetation are spatially intertwined, whereas the compared methods exhibit missing structures or semantic confusion in the same areas.

\begin{table*}[ht]
\centering
\caption{
Overall and scene-wise 3D object detection results on Spheriverse.
mAP and NDS are reported following the nuScenes evaluation protocol~\cite{Nuscenes}, and higher values indicate better performance.
Spar., Dens., BEVF., Pola., Solo., DETR., CoiN., PD-B., and GeoB. denote SparseBEV~\cite{sparsebev}, DenseBEV~\cite{densebev}, BEVFormer~\cite{bevformer}, PolarBEVDet~\cite{polarbevdet}, SOLOFusion~\cite{Solofusion}, DETR3D~\cite{detr3d}, CoIn3D~\cite{coin3d}, PD-BEV~\cite{lu2025towards}, and GeoBEV~\cite{zhang2025geobev}, respectively.
Expre., Funct., and Struc. denote Expressway, Functional, and Structural scenes, respectively.
}
\label{tab:car_scene_detection_results}

\begingroup

\definecolor{detsectioncolor}{RGB}{247,248,250}
\newcommand{\detMethodHead}[1]{\rotatebox{55}{\strut #1}}

\setlength{\tabcolsep}{2.2pt}
\renewcommand{\arraystretch}{0.96}
\normalsize

\begin{tabular*}{0.96\textwidth}
{@{\extracolsep{\fill}}ll|*{9}{c}@{}}
\toprule

\multicolumn{1}{c}{
\multirow[c]{2}{*}[-7ex]{\textbf{Group}}
}
&
\multicolumn{1}{c|}{
\multirow[c]{2}{*}[-7ex]{\textbf{Metric}}
}
&
\multicolumn{9}{c}{\textbf{Methods}} \\

\cmidrule(lr){3-11}

&
& \detMethodHead{Spar.}
& \detMethodHead{Dens.}
& \detMethodHead{BEVF.}
& \detMethodHead{Pola.}
& \detMethodHead{Solo.}
& \detMethodHead{DETR.}
& \detMethodHead{CoiN.}
& \detMethodHead{PD-B.}
& \detMethodHead{GeoB.} \\

\specialrule{0.10em}{0.7ex}{0.3ex}

\rowcolor{detsectioncolor}
\multicolumn{11}{c}{
\textbf{Overall 3D Object Detection Results}
} \\
\midrule

& mAP $\uparrow$
& 0.1689
& 0.0941
& 0.0941
& 0.1364
& 0.1316
& 0.1296
& 0.1078
& 0.1178
& 0.0654 \\

& NDS $\uparrow$
& 0.1221
& 0.0874
& 0.0806
& 0.1086
& 0.1017
& 0.0996
& 0.0939
& 0.0985
& 0.0715 \\

\specialrule{0.10em}{0.7ex}{0.3ex}

\rowcolor{detsectioncolor}
\multicolumn{11}{c}{
\textbf{Scene-wise 3D Object Detection Results}
} \\
\midrule

\multirow[c]{2}{*}{\textbf{Expre.}}
& mAP $\uparrow$
& 0.1250
& 0.0701
& 0.0813
& 0.1110
& 0.1175
& 0.1009
& 0.0868
& 0.1021
& 0.0579 \\

& NDS $\uparrow$
& 0.0972
& 0.0687
& 0.0691
& 0.0925
& 0.0926
& 0.0822
& 0.0805
& 0.0891
& 0.0653 \\

\midrule

\multirow[c]{2}{*}{\textbf{Funct.}}
& mAP $\uparrow$
& 0.0531
& 0.0266
& 0.0232
& 0.0709
& 0.0674
& 0.0314
& 0.0508
& 0.0722
& 0.0242 \\

& NDS $\uparrow$
& 0.0636
& 0.0411
& 0.0409
& 0.0593
& 0.0668
& 0.0478
& 0.0604
& 0.0733
& 0.0305 \\

\midrule

\multirow[c]{2}{*}{\textbf{Rural}}
& mAP $\uparrow$
& 0.0833
& 0.0398
& 0.0433
& 0.0721
& 0.0648
& 0.0619
& 0.0526
& 0.0647
& 0.0312 \\

& NDS $\uparrow$
& 0.0829
& 0.0519
& 0.0516
& 0.0595
& 0.0504
& 0.0675
& 0.0466
& 0.0514
& 0.0332 \\

\midrule

\multirow[c]{2}{*}{\textbf{Struc.}}
& mAP $\uparrow$
& 0.3175
& 0.1864
& 0.1821
& 0.2565
& 0.2970
& 0.2113
& 0.2743
& 0.2648
& 0.1122 \\

& NDS $\uparrow$
& 0.2124
& 0.1313
& 0.1068
& 0.1900
& 0.1996
& 0.1438
& 0.2075
& 0.2040
& 0.0725 \\

\midrule

\multirow[c]{2}{*}{\textbf{Urban}}
& mAP $\uparrow$
& 0.1976
& 0.1207
& 0.1115
& 0.1467
& 0.1452
& 0.1538
& 0.1167
& 0.1233
& 0.0735 \\

& NDS $\uparrow$
& 0.1367
& 0.0962
& 0.0858
& 0.1179
& 0.1124
& 0.1126
& 0.1010
& 0.1041
& 0.0764 \\

\bottomrule
\end{tabular*}

\endgroup
\end{table*}

\begin{figure*}[t]
    \centering
    \includegraphics[width=0.96\textwidth]{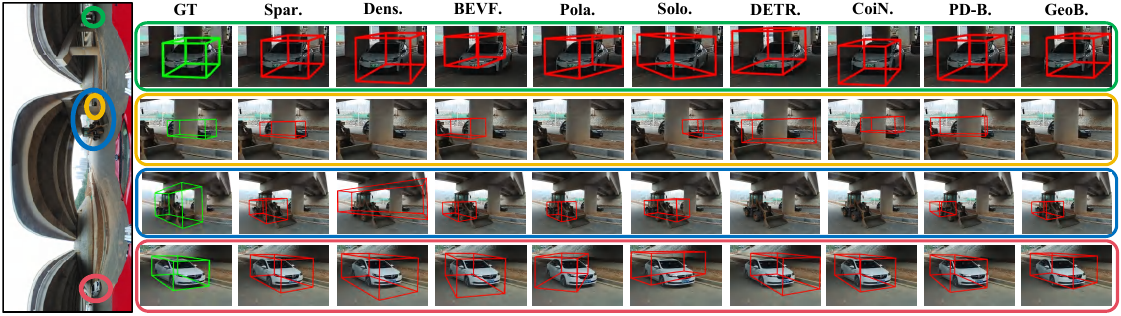} 
\caption{{Qualitative comparison of spherical 3D object detection on Spheriverse.}
The four color-coded regions in the input image are enlarged by row, while the ground truth (GT) and predictions from representative methods are compared by column. Spar., Dens., BEVF., Pola., Solo., DETR., CoiN., PD-B., and GeoB. denote SparseBEV~\cite{sparsebev}, DenseBEV~\cite{densebev}, BEVFormer~\cite{bevformer}, PolarBEVDet~\cite{polarbevdet}, SOLOFusion~\cite{Solofusion}, DETR3D~\cite{detr3d}, CoIn3D~\cite{coin3d}, PD-BEV~\cite{lu2025towards}, and GeoBEV~\cite{zhang2025geobev}, respectively. Green and red boxes denote the projected ground-truth and predicted 3D boxes.}
\label{fig:detection_qualitative}
    \label{fig:Od}
\end{figure*}

\subsection{Semantic Mapping and 3D Object Detection}
\textbf{Evaluation Metrics:}
For \textit{semantic mapping}, we evaluate the semantic consistency between the predicted BEV representation and the ground-truth semantic map. Following established BEV mapping protocols~\cite{onebev,hdmapnet}, we use mean Intersection over Union (mIoU) as the primary metric, which averages the overlap between predicted and ground-truth regions across all semantic categories. For \textit{3D object detection}, we follow the nuScenes evaluation protocol and report mean Average Precision (mAP) and the nuScenes Detection Score (NDS).

\textbf{Implementation Details:}
For \textit{semantic mapping}, the target BEV maps are represented as dense $200 \times 200$ grids containing seven semantic categories. We adapt ten representative methods to Spheriverse: OneBEV~\cite{onebev}, HDMapNet~\cite{hdmapnet}, PivotNet~\cite{pivotnet}, VectorMapNet~\cite{vectormapnet}, MapTR~\cite{maptr}, SparseBEV~\cite{sparsebev}, BEVFormer~\cite{bevformer}, PETRv2~\cite{petr}, SeqBEV~\cite{bevseq}, and TPVFormer~\cite{TPVFormer}. For \textit{3D object detection}, we evaluate nine representative methods: SparseBEV~\cite{sparsebev}, DenseBEV~\cite{densebev}, BEVFormer~\cite{bevformer}, PolarBEVDet~\cite{polarbevdet}, SOLOFusion~\cite{Solofusion}, DETR3D~\cite{detr3d}, CoIn3D~\cite{coin3d}, PD-BEV~\cite{lu2025towards}, and GeoBEV~\cite{zhang2025geobev}. Within each benchmark, all methods are trained for $28$ epochs using the same data splits and a unified task-specific evaluation protocol.

\begin{table*}[!ht]
\centering
\caption{
Ablation study of the proposed components on Spheriverse.
The upper panel reports the overall and class-wise semantic occupancy results, while the lower panel reports the scene-wise mIoU and GeoIoU.
All metrics are reported in \%, and higher values indicate better performance.
Base. denotes the original SurroundOcc~\cite{surroundocc} baseline;
\ding{172}, \ding{173}, and \ding{174} denote the InternImage-T backbone~\cite{internimage}, Cartesian--Spherical Representation Remodeling (CSRR), and Spherical Evidence Re-querying (SER), respectively.
Pers., Veh., Bldg., Veg., Pill., and Surf. denote Person, Vehicle, Building, Vegetation, Pillar, and Surface, respectively.
}
\label{tab:ablation_study}

\begingroup

\definecolor{abperson}{RGB}{156,39,176}
\definecolor{abvehicle}{RGB}{255,167,38}
\definecolor{abbike}{RGB}{66,133,244}
\definecolor{abbuilding}{RGB}{126,87,194}
\definecolor{abvegetation}{RGB}{67,160,71}
\definecolor{abpillar}{RGB}{141,110,99}
\definecolor{abroad}{RGB}{84,110,122}
\definecolor{absurface}{RGB}{176,190,197}
\definecolor{abothers}{RGB}{236,64,122}
\definecolor{absectioncolor}{RGB}{247,248,250}

\newcommand{\abClassHead}[2]{%
  \mbox{%
    \raisebox{-0.2ex}{\textcolor{#1}{{\large$\bullet$}}}\,#2%
  }%
}
\newcommand{\abMIoU}{\mbox{mIoU~$\uparrow$}}
\newcommand{\abGeoIoU}{\mbox{GeoIoU~$\uparrow$}}

\setlength{\tabcolsep}{1.5pt}
\renewcommand{\arraystretch}{1.10}
\tiny

\resizebox{0.99\textwidth}{!}{%
\begin{tabular}{*{4}{c}|cc|*{9}{c}}
\toprule
\rowcolor{absectioncolor}
\multicolumn{15}{c}{
\textbf{Overall and Class-wise Semantic Occupancy Results}
} \\
\midrule

\multicolumn{4}{c|}{\textbf{Configuration}}
& \multicolumn{2}{c|}{\textbf{Overall}}
& \multicolumn{9}{c}{\textbf{Class-wise IoU}} \\

\cmidrule(lr){1-4}
\cmidrule(lr){5-6}
\cmidrule(lr){7-15}

Base.
& \ding{172}
& \ding{173}
& \ding{174}
& \abMIoU
& \abGeoIoU
& \abClassHead{abperson}{Pers.}
& \abClassHead{abvehicle}{Veh.}
& \abClassHead{abbike}{Bike}
& \abClassHead{abbuilding}{Bldg.}
& \abClassHead{abvegetation}{Veg.}
& \abClassHead{abpillar}{Pill.}
& \abClassHead{abroad}{Road}
& \abClassHead{absurface}{Surf.}
& \abClassHead{abothers}{Others} \\
\midrule

\checkmark & -- & -- & --
& 12.05 & 22.55
& 1.65 & 13.97 & 5.16 & 9.40 & 14.08
& 12.57 & 36.58 & 11.99 & 3.07 \\

\checkmark & \checkmark & -- & --
& 12.14 & 23.35
& 1.34 & 14.50 & 4.90 & 9.26 & 14.98
& 13.19 & 36.80 & 11.33 & 2.95 \\

\checkmark & \checkmark & \checkmark & --
& 12.71 & 23.50
& 1.70 & 14.92 & 6.30 & 9.41 & 14.99
& 13.50 & 37.77 & 12.65 & 3.10 \\

\checkmark & -- & \checkmark & \checkmark
& 13.17 & 23.85
& 2.72 & 15.50 & 6.62 & 9.99 & 15.01
& 14.55 & 38.47 & 12.39 & 3.32 \\

\checkmark & \checkmark & \checkmark & \checkmark
& \textbf{13.91} & \textbf{24.65}
& \textbf{3.02} & \textbf{15.71} & \textbf{7.50}
& \textbf{10.25} & \textbf{15.57} & \textbf{16.60}
& \textbf{39.99} & \textbf{13.01} & \textbf{3.53} \\

\bottomrule
\end{tabular}%
}

\vspace{0.55em}

\resizebox{0.99\textwidth}{!}{%
\begin{tabular}{*{4}{c}|cc|cc|cc|cc|cc}
\toprule
\rowcolor{absectioncolor}
\multicolumn{14}{c}{
\textbf{Scene-wise Semantic Occupancy Results}
} \\
\midrule

\multicolumn{4}{c|}{\textbf{Configuration}}
& \multicolumn{2}{c|}{\textbf{Expressway}}
& \multicolumn{2}{c|}{\textbf{Functional}}
& \multicolumn{2}{c|}{\textbf{Rural}}
& \multicolumn{2}{c|}{\textbf{Structural}}
& \multicolumn{2}{c}{\textbf{Urban}} \\

\cmidrule(lr){1-4}
\cmidrule(lr){5-6}
\cmidrule(lr){7-8}
\cmidrule(lr){9-10}
\cmidrule(lr){11-12}
\cmidrule(lr){13-14}

Base.
& \ding{172}
& \ding{173}
& \ding{174}
& \abMIoU & \abGeoIoU
& \abMIoU & \abGeoIoU
& \abMIoU & \abGeoIoU
& \abMIoU & \abGeoIoU
& \abMIoU & \abGeoIoU \\
\midrule

\checkmark & -- & -- & --
& 11.29 & 25.36
& 8.38 & 23.39
& 10.22 & 20.23
& 13.66 & 34.79
& 8.31 & 17.49 \\

\checkmark & \checkmark & -- & --
& 11.37 & 25.79
& 9.12 & 23.88
& 10.69 & 21.22
& 14.39 & 35.64
& 8.68 & 17.95 \\

\checkmark & \checkmark & \checkmark & --
& 12.09 & 25.68
& 9.36 & 25.13
& 10.98 & 21.68
& 14.22 & 36.33
& 8.78 & 18.21 \\

\checkmark & -- & \checkmark & \checkmark
& 12.29 & 26.47
& 9.33 & 25.48
& 11.14 & 21.40
& 15.28 & 37.55
& 9.35 & 18.79 \\

\checkmark & \checkmark & \checkmark & \checkmark
& \textbf{12.82} & \textbf{27.02}
& \textbf{10.17} & \textbf{25.90}
& \textbf{11.32} & \textbf{22.43}
& \textbf{16.60} & \textbf{39.04}
& \textbf{9.64} & \textbf{19.32} \\

\bottomrule
\end{tabular}%
}

\endgroup
\end{table*}

\textbf{Results and Analyses:}
For \textit{semantic mapping}, Table~\ref{tab:bev_global_scene} shows that OneBEV~\cite{onebev}, which is specifically designed for spherical observations, achieves the highest overall mIoU and outperforms the second-ranked HDMapNet~\cite{hdmapnet} by $1.95$ percentage points ($22.66\%$ \textit{vs.} $20.71\%$). OneBEV also ranks first across all five scene subsets, with its largest advantage observed in structurally constrained scenes ($23.15\%$ \textit{vs.} $20.28\%$). At the category level, HDMapNet~\cite{hdmapnet} achieves the highest IoUs for {Building} and {Others}. For \textit{3D object detection}, Table~\ref{tab:car_scene_detection_results} shows that SparseBEV~\cite{sparsebev} achieves the best overall performance, with an mAP of $0.1689$ and an NDS of $0.1221$, followed by PolarBEVDet~\cite{polarbevdet}. SparseBEV also performs best in expressway, rural, structurally constrained, and urban scenes, whereas PD-BEV~\cite{lu2025towards} performs best in functional scenes. Performance varies substantially across the five scene categories, with higher detection accuracy in structurally constrained scenes and lower accuracy in functional and rural scenes. These results provide reproducible baselines for spherical BEV perception and demonstrate that spherical 3D object detection remains sensitive to scene structure and object distribution. As shown in Fig.~\ref{fig:detection_qualitative}, the compared methods frequently produce missed detections or inaccurate localization in the selected regions. These results illustrate the difficulty of recovering object locations in metric Cartesian 3D space from wide-field-of-view spherical observations represented in angular coordinates. The non-uniform mapping between the two coordinate systems, together with projection-induced appearance distortions, complicates image-to-3D correspondence and remains an important challenge for future research.

\section{Ablation Studies}
\label{sec:abla}

\subsection{Main Components}
To evaluate the contribution of each component, we progressively incorporate InternImage-T~\cite{internimage}, CSRR, and SER into the original SurroundOcc baseline~\cite{surroundocc}.
As shown in Table~\ref{tab:ablation_study}, replacing ResNet~\cite{he2016deep} with the pretrained InternImage-T~\cite{internimage} improves mIoU from \(12.05\%\) to \(12.14\%\) and GeoIoU from \(22.55\%\) to \(23.35\%\), confirming the effectiveness of the stronger backbone.
Adding CSRR further increases mIoU to \(12.71\%\) and GeoIoU to \(23.50\%\), with gains of \(0.57\) and \(0.15\) percentage points, respectively.
By incorporating range-azimuth relations into the lifted voxel features, CSRR strengthens their correspondence with spherical geometry before occupancy decoding. Furthermore, SER produces the largest incremental improvement, raising mIoU and GeoIoU to \(13.91\%\) and \(24.65\%\), respectively.
During the initial lifting, sampling offsets and attention weights are predicted from learnable voxel queries around projected voxel reference points. In contrast, SER combines the Cartesian–spherical voxel with explicit RTZ embeddings to construct conditioned queries, which subsequently predict offsets and attention weights for re-querying the spherical image features. Furthermore, we evaluate CSRR and SER while retaining the same ResNet backbone~\cite{he2016deep}.
Adding both modules improves mIoU from $12.05\%$ to $13.17\%$ and GeoIoU
from $22.55\%$ to $23.85\%$.
These gains demonstrate that CSRR and SER improve occupancy prediction
without relying on increased backbone capacity. Overall, CSRR establishes an explicit correspondence between Cartesian voxel coordinates and spherical range-azimuth geometry. Building on this representation, SER conditions each Cartesian--spherical voxel query on its spherical geometry, enabling targeted retrieval of relevant evidence from the source spherical image features.

\begin{table}[t!]
\centering
\caption{
Semantic occupancy performance across horizontal azimuth intervals and vertical voxel layers on Spheriverse.
mIoU and GeoIoU are reported in \%.
Horizontal partitions span the full \(360^{\circ}\) field of view, whereas vertical partitions are defined by voxel-height index \(z\): \(12\)--\(15\), \(8\)--\(11\), \(4\)--\(7\), and \(0\)--\(3\) denote the top, upper-middle, lower-middle, and bottom layers, respectively.
TPVFo., Surro., MonoS., and Quadr. denote TPVFormer~\cite{TPVFormer}, SurroundOcc~\cite{surroundocc}, MonoScene~\cite{monoscene}, and QuadricFormer~\cite{quadricformer}, respectively.
Light orange and light blue indicate the best and second-best results.
}
\label{tab:angular_vertical_results}

\begingroup
\definecolor{regbestcolor}{RGB}{255,232,204}
\definecolor{regsecondcolor}{RGB}{225,240,255}
\newcommand{\regbest}[1]{\cellcolor{regbestcolor}#1}
\newcommand{\regsecond}[1]{\cellcolor{regsecondcolor}#1}

\setlength{\tabcolsep}{2.0pt}
\renewcommand{\arraystretch}{1.04}
\scriptsize

\resizebox{\columnwidth}{!}{%
\begin{tabular}{cc|l|ccccc}
\toprule
\multicolumn{2}{c|}{\textbf{Partition}}
& \textbf{Metric}
& \textbf{TPVFo.}
& \textbf{Surro.}
& \textbf{MonoS.}
& \textbf{Quadr.}
& \textbf{Ours} \\
\midrule

\multirow[c]{16}{*}{\rotatebox{90}{\textbf{Horizontal}}}
& \multirow[c]{2}{*}{\(0\)--\(45^{\circ}\)}
& mIoU
& \regsecond{10.60}
& 10.25
& 9.03
& 9.60
& \regbest{11.96} \\
& & GeoIoU
& \regsecond{21.41}
& 20.92
& 19.93
& 19.22
& \regbest{22.95} \\

\cmidrule(lr){2-8}
& \multirow[c]{2}{*}{\(45\)--\(90^{\circ}\)}
& mIoU
& \regsecond{11.73}
& 10.92
& 10.82
& 10.44
& \regbest{12.85} \\
& & GeoIoU
& \regsecond{19.10}
& 19.00
& 18.03
& 17.17
& \regbest{20.59} \\

\cmidrule(lr){2-8}
& \multirow[c]{2}{*}{\(90\)--\(135^{\circ}\)}
& mIoU
& \regsecond{11.17}
& 10.97
& 10.96
& 10.62
& \regbest{13.46} \\
& & GeoIoU
& \regsecond{20.16}
& 19.89
& 19.60
& 18.76
& \regbest{22.33} \\

\cmidrule(lr){2-8}
& \multirow[c]{2}{*}{\(135\)--\(180^{\circ}\)}
& mIoU
& 11.64
& \regsecond{12.50}
& 11.38
& 11.10
& \regbest{14.31} \\
& & GeoIoU
& 29.38
& \regsecond{29.85}
& 29.03
& 28.27
& \regbest{32.36} \\

\cmidrule(lr){2-8}
& \multirow[c]{2}{*}{\(180\)--\(225^{\circ}\)}
& mIoU
& \regsecond{13.93}
& 13.83
& 13.22
& 12.89
& \regbest{16.26} \\
& & GeoIoU
& 28.00
& \regsecond{28.34}
& 27.98
& 27.14
& \regbest{30.91} \\

\cmidrule(lr){2-8}
& \multirow[c]{2}{*}{\(225\)--\(270^{\circ}\)}
& mIoU
& 11.27
& \regsecond{11.32}
& 10.80
& 10.21
& \regbest{13.36} \\
& & GeoIoU
& 18.29
& \regsecond{20.38}
& 18.16
& 17.00
& \regbest{22.71} \\

\cmidrule(lr){2-8}
& \multirow[c]{2}{*}{\(270\)--\(315^{\circ}\)}
& mIoU
& \regsecond{11.59}
& 11.58
& \regsecond{11.59}
& 10.27
& \regbest{12.80} \\
& & GeoIoU
& 17.90
& \regsecond{18.49}
& 16.88
& 15.58
& \regbest{20.94} \\

\cmidrule(lr){2-8}
& \multirow[c]{2}{*}{\(315\)--\(360^{\circ}\)}
& mIoU
& \regsecond{12.32}
& 11.96
& 11.57
& 10.89
& \regbest{13.70} \\
& & GeoIoU
& \regsecond{21.17}
& 21.14
& 20.67
& 19.35
& \regbest{22.79} \\

\midrule

\multirow[c]{8}{*}{\rotatebox{90}{\textbf{Vertical}}}
& \multirow[c]{2}{*}{\(12\)--\(15\)}
& mIoU
& 2.56
& \regsecond{2.97}
& 2.49
& 2.83
& \regbest{3.20} \\
& & GeoIoU
& 10.41
& \regsecond{11.73}
& 10.25
& 8.93
& \regbest{12.63} \\

\cmidrule(lr){2-8}
& \multirow[c]{2}{*}{\(8\)--\(11\)}
& mIoU
& \regsecond{12.88}
& 12.27
& 11.87
& 11.23
& \regbest{14.18} \\
& & GeoIoU
& 20.56
& \regsecond{20.67}
& 19.74
& 18.89
& \regbest{22.48} \\

\cmidrule(lr){2-8}
& \multirow[c]{2}{*}{\(4\)--\(7\)}
& mIoU
& 11.28
& \regsecond{11.50}
& 10.38
& 10.98
& \regbest{13.57} \\
& & GeoIoU
& 30.00
& \regsecond{30.51}
& 29.68
& 28.87
& \regbest{33.14} \\

\cmidrule(lr){2-8}
& \multirow[c]{2}{*}{\(0\)--\(3\)}
& mIoU
& 1.71
& \regsecond{3.06}
& 1.19
& 2.57
& \regbest{3.85} \\
& & GeoIoU
& 2.28
& \regsecond{5.67}
& 1.69
& 4.47
& \regbest{6.99} \\

\bottomrule
\end{tabular}%
}
\endgroup
\end{table}

\subsection{Angular and Vertical Region Analysis.} To assess the ability of SphereOcc, we analyze its performance across horizontal azimuth intervals and vertical voxel layers.
We compare SphereOcc with TPVFormer~\cite{TPVFormer}, SurroundOcc~\cite{surroundocc}, MonoScene~\cite{monoscene}, and QuadricFormer~\cite{quadricformer}.
As shown in Table~\ref{tab:angular_vertical_results}, SphereOcc achieves the highest mIoU and GeoIoU in every horizontal and vertical partition.
This consistent performance supports the effectiveness of embedding spherical geometry into voxel features and re-querying source spherical image features to construct spatially robust voxel representations. Across the full \(360^{\circ}\) horizontal field of view, SphereOcc outperforms all comparison methods in every azimuth interval.
The improvements are particularly clear in the \(135^{\circ}\)--\(180^{\circ}\) and \(180^{\circ}\)--\(225^{\circ}\) intervals for both semantic recognition and geometric reconstruction.
These consistent gains across viewing directions indicate improved robustness to uneven angular information distributions and projection distortions arising from geometric variations. 
Vertically, SphereOcc achieves the highest performance across all voxel-height ranges, including the sparsely observed top and bottom regions.
The improvements are especially pronounced in the upper-middle and lower-middle layers, where semantic structures are densely interleaved and therefore more difficult to distinguish.
% These gains are consistent with the explicit integration of spherical observation cues into Cartesian voxel features, which helps distinguish spatially entangled features. Overall, t
These results show that SphereOcc produces reliable occupancy representations across different spatial orientations and height ranges, supporting its effectiveness for spherical occupancy prediction.

\begin{table}[!t]
\centering
\caption{
Robustness of semantic occupancy prediction under reduced spherical fields of view on Spheriverse.
mIoU and GeoIoU are reported in \%.
The horizontal FoV is reduced symmetrically from the left and right boundaries, whereas the vertical FoV is reduced from top to bottom while retaining the lower portion of the spherical image.
Masked pixels are set to zero.
TPVFo., Surro., MonoS., and Quadr. denote TPVFormer~\cite{TPVFormer}, SurroundOcc~\cite{surroundocc}, MonoScene~\cite{monoscene}, and QuadricFormer~\cite{quadricformer}, respectively.
Light orange and light blue indicate the best and second-best results.
}
\label{tab:fov_robustness_results}

\begingroup
\definecolor{fovbestcolor}{RGB}{255,232,204}
\definecolor{fovsecondcolor}{RGB}{225,240,255}
\newcommand{\fovbest}[1]{\cellcolor{fovbestcolor}#1}
\newcommand{\fovsecond}[1]{\cellcolor{fovsecondcolor}#1}

\setlength{\tabcolsep}{2.0pt}
\renewcommand{\arraystretch}{1.04}
\scriptsize

\resizebox{\columnwidth}{!}{%
\begin{tabular}{cc|l|ccccc}
\toprule
\multicolumn{2}{c|}{\textbf{Retained FoV}}
& \textbf{Metric}
& \textbf{TPVFo.}
& \textbf{Surro.}
& \textbf{MonoS.}
& \textbf{Quadr.}
& \textbf{Ours} \\
\midrule

\multirow[c]{10}{*}{\rotatebox{90}{\textbf{Horizontal}}}
& \multirow[c]{2}{*}{\(360^{\circ}\)}
& mIoU
& \fovsecond{12.21}
& 12.05
& 11.44
& 11.09
& \fovbest{13.91} \\
& & GeoIoU
& 22.35
& \fovsecond{22.55}
& 21.68
& 20.76
& \fovbest{24.65} \\

\cmidrule(lr){2-8}
& \multirow[c]{2}{*}{\(300^{\circ}\)}
& mIoU
& 10.84
& \fovsecond{11.23}
& 9.70
& 9.76
& \fovbest{13.06} \\
& & GeoIoU
& 20.43
& \fovsecond{21.30}
& 18.62
& 18.07
& \fovbest{23.82} \\

\cmidrule(lr){2-8}
& \multirow[c]{2}{*}{\(240^{\circ}\)}
& mIoU
& 9.68
& \fovsecond{10.24}
& 8.55
& 8.81
& \fovbest{11.55} \\
& & GeoIoU
& 19.06
& \fovsecond{19.51}
& 16.65
& 16.39
& \fovbest{21.35} \\

\cmidrule(lr){2-8}
& \multirow[c]{2}{*}{\(180^{\circ}\)}
& mIoU
& 8.64
& \fovsecond{9.37}
& 7.54
& 7.93
& \fovbest{10.23} \\
& & GeoIoU
& 18.01
& \fovsecond{18.14}
& 14.96
& 14.75
& \fovbest{19.27} \\

\cmidrule(lr){2-8}
& \multirow[c]{2}{*}{\(120^{\circ}\)}
& mIoU
& 7.46
& \fovsecond{8.34}
& 6.33
& 6.82
& \fovbest{8.98} \\
& & GeoIoU
& \fovsecond{17.09}
& 16.37
& 12.68
& 12.78
& \fovbest{17.42} \\

\midrule

\multirow[c]{8}{*}{\rotatebox{90}{\textbf{Vertical}}}
& \multirow[c]{2}{*}{\(136.70^{\circ}\)}
& mIoU
& \fovsecond{12.21}
& 12.05
& 11.44
& 11.09
& \fovbest{13.91} \\
& & GeoIoU
& 22.35
& \fovsecond{22.55}
& 21.68
& 20.76
& \fovbest{24.65} \\

\cmidrule(lr){2-8}
& \multirow[c]{2}{*}{\(106.70^{\circ}\)}
& mIoU
& \fovsecond{12.10}
& 12.06
& 11.26
& 10.88
& \fovbest{13.90} \\
& & GeoIoU
& 21.84
& \fovsecond{22.49}
& 21.48
& 20.40
& \fovbest{24.61} \\

\cmidrule(lr){2-8}
& \multirow[c]{2}{*}{\(76.70^{\circ}\)}
& mIoU
& 10.42
& \fovsecond{11.82}
& 10.15
& 9.73
& \fovbest{13.80} \\
& & GeoIoU
& 20.37
& \fovsecond{22.26}
& 19.61
& 19.01
& \fovbest{24.48} \\

\cmidrule(lr){2-8}
& \multirow[c]{2}{*}{\(46.70^{\circ}\)}
& mIoU
& 7.62
& \fovsecond{9.99}
& 6.92
& 6.51
& \fovbest{11.59} \\
& & GeoIoU
& 16.63
& \fovsecond{18.68}
& 16.19
& 13.05
& \fovbest{21.07} \\

\bottomrule
\end{tabular}%
}
\endgroup
\end{table}

\subsection{Robustness under Reduced Spherical FoVs}

To evaluate SphereOcc under incomplete spherical observations, we progressively reduce the horizontal and vertical fields of view by masking the input image. As shown in Table~\ref{tab:fov_robustness_results}, SphereOcc achieves the highest mIoU and GeoIoU across all evaluated FoVs, indicating that it maintains reliable voxel representations under incomplete observations. Under horizontal FoV reduction, all methods exhibit performance degradation as the available angular coverage decreases.
At the most restrictive \(120^{\circ}\) FoV, SphereOcc still achieves \(8.98\%\) mIoU and \(17.42\%\) GeoIoU, outperforming the corresponding second-best results of \(8.34\%\) and \(17.09\%\).
SphereOcc exhibits greater stability under vertical FoV reduction.
When the vertical FoV is reduced from \(136.70^{\circ}\) to \(76.70^{\circ}\), its mIoU decreases only from \(13.91\%\) to \(13.80\%\), while its GeoIoU decreases from \(24.65\%\) to \(24.48\%\).
Notably, these results remain higher than the best full-FoV results of all competing methods, with \(13.80\%\) versus \(12.21\%\) mIoU and \(24.48\%\) versus \(22.55\%\) GeoIoU. These results are consistent with the design objectives of SphereOcc.
CSRR explicitly embeds spherical range--azimuth relations into Cartesian voxel features, providing a geometry-aware cross-space prior as angular coverage decreases.
Building on these features, SER constructs conditional queries from the current voxel representations and their RTZ positions to selectively retrieve relevant semantic evidence from the available spherical observations.

\begin{figure*}[ht]
    \centering
    \includegraphics[width=0.96\textwidth]{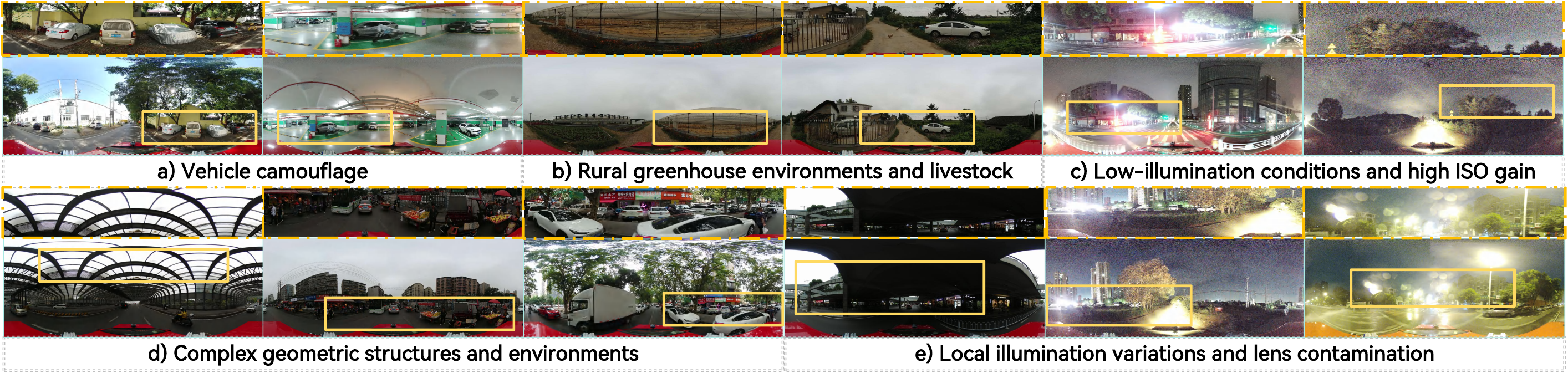} 
    \caption{Representative challenging scenarios in Spheriverse, including vehicle camouflage, uncommon rural environments, low-light imaging, complex geometry, and lens contamination.}
    \label{fig:challenges}
    \vspace{-1em}
\end{figure*}

\subsection{Representative Challenging Scenarios in Dataset}
As shown in Fig.~\ref{fig:challenges}, we present representative challenging scenarios in Spheriverse that are not fully characterized by aggregate statistics. 
First, vehicle camouflage and uncommon rural content, such as greenhouses and livestock, introduce substantial semantic ambiguity and long-tail appearance variations, corresponding to Fig.~\ref{fig:challenges}(a,b). Camouflaged vehicles offer weak object--background contrast, while rural
content challenges recognition beyond the appearance patterns of
conventional urban traffic scenes. 
Second, curved, overhead, and multi-level structures, together with complex
spatial layouts across the full field of view, as illustrated in Fig.~\ref{fig:challenges}(d), pose distinctive challenges
for spherical perception. 
Third, low illumination, high ISO gain, local exposure variations, and lens contamination cause spatially non-uniform degradation across spherical images, as illustrated in Fig.~\ref{fig:challenges}(c,e). 
These factors frequently occur together, creating compound challenges for semantic recognition and geometric reconstruction.
Overall, Spheriverse extends beyond well-illuminated and structurally regular urban environments, providing a realistic testbed for evaluating spherical perception under diverse in-the-wild conditions.

\section{Conclusion}
This work addresses the cross-space representation gap between angular spherical observations and Cartesian voxels, which complicates the construction of geometrically consistent and semantically informative 3D representations. To systematically study this problem, we introduce Spheriverse, including $644$ synchronized image-LiDAR sequences collected across $13$ diverse regions. Spheriverse covers varied scene types, times of day, and weather conditions, with a three-level semantic annotation hierarchy. Based on this dataset, we establish unified benchmarks for semantic occupancy prediction, semantic mapping, and 3D object detection, evaluating more than $30$ representative methods through overall and scene-wise analyses.

We further propose SphereOcc for dense semantic occupancy prediction. CSRR embeds spherical range-azimuth geometry into Cartesian voxel features, while SER retrieves complementary semantic evidence from source spherical image features using RTZ-conditioned voxel states. SphereOcc achieves $13.91\%$ mIoU and $24.65\%$ GeoIoU, improving the best prior results by $1.70$ and $2.10$ percentage points, respectively, and ranks first in both metrics across all five scene categories. Its consistent performance across spatial partitions and reduced fields of view further supports its robustness to directional variations and incomplete observations. Together, Spheriverse, its benchmarks, and SphereOcc provide a unified foundation for studying 3D perception from spherical observations.

Building on Spheriverse and SphereOcc, future work will extend spherical perception toward end-to-end embodied driving by connecting spherical observations and dense 3D representations with safety-critical decision-making, language-grounded spatial reasoning, and trajectory planning. This direction will enable evaluation from 3D scene understanding to physical action and further investigate how global spherical context supports reliable decisions.

\ifCLASSOPTIONcompsoc
  % The Computer Society usually uses the plural form
  \section*{Acknowledgments}
\else
  % regular IEEE prefers the singular form
  \section*{Acknowledgment}
\fi
This work was supported in part by the National Natural Science Foundation of China (Grant No. 62388101 and No. 62473139), in part by the Hunan Provincial Research and Development Project (Grant No. 2025QK3019), in part by the Hunan Provincial Innovation Foundation for Postgraduate (Grant No. CX20250579), and in part by the State Key Laboratory of Autonomous Intelligent Unmanned Systems (the opening project number ZZKF2025-2-10).

\clearpage

\bibliographystyle{IEEEtran}
\bibliography{main}

\clearpage

\appendices

\section{Motivation}
\label{app:motivation}
Spherical images encode visual information in an angular domain, whereas dense representations of the 3D world are organized in Cartesian voxel space.
This divergence creates a cross-space representation gap, giving rise to projection-induced distortions in object appearance, geometry, and spatial relationships.
To address this challenge, we contribute a dataset, unified benchmarks, and a dense occupancy prediction framework.

\textbf{Dataset.} Spheriverse provides synchronized spherical image--LiDAR observations and hierarchical 3D semantic annotations across diverse real-world environments, lighting conditions, and weather. \textbf{Benchmarks.} We establish unified benchmarks for semantic occupancy prediction, BEV semantic mapping, and 3D object detection, evaluating over $30$ representative methods through overall and scene-wise comparisons. \textbf{Method.} For dense semantic occupancy prediction, SphereOcc aligns Cartesian voxel features with spherical observations through two complementary modules.
Cartesian--Spherical Representation Remodeling (CSRR) establishes an explicit correspondence between spherical range--azimuth geometry and Cartesian voxel representations.
Spherical Evidence Re-querying (SER) selectively retrieves relevant semantic evidence from source spherical image features according to each voxel's current representation and spherical position. \textbf{Results.} SphereOcc achieves $13.91\%$ mIoU and $24.65\%$ GeoIoU, exceeding the strongest methods for each metric by $1.70$ and $2.10$ percentage points, respectively.
It ranks first in both metrics across all five scene categories and maintains consistent advantages across the evaluated spatial partitions and reduced fields of view.

\section{Dataset Details}
\label{app:dataset}
\subsection{Acquisition and Scene-Level Annotation}
\label{app:acquisition}
The raw collection contains $89,674$ spherical images.
Temporal alignment and filtering yield $64,400$ synchronized image-LiDAR pairs organized into $644$ sequences.
Each sequence spans $20$ seconds at $5$\,Hz.
The collection covers $13$ geographically and visually diverse regions.

\textbf{Acquisition platform and geometric supervision.} 
Spheriverse records spherical images together with calibrated, time-aligned LiDAR observations.
The acquisition platform integrates a DuxCam M4 spherical camera~\cite{panodux_m4} and a 128-beam Hesai OT128 LiDAR~\cite{hesai_ot128}.
The camera records images at $5188\times1979$ pixels and provides an effective field of view of $360^\circ$ horizontally and $136.70^\circ$ vertically.
It combines Sony 8MP image sensors with fisheye lenses to support image acquisition under low-light conditions.
The platform also includes a camera-integrated GNSS receiver and an external GNSS/INS unit.
Camera intrinsics are calibrated using Zhang's method~\cite{zhang1999flexible}, and camera--LiDAR extrinsics are calibrated in the LiDAR coordinate system~\cite{PanoRingCalib}.
A domain controller uses the Precision Time Protocol for clock alignment, enabling synchronized image--LiDAR acquisition.

Compared with PanoMMOcc~\cite{PanoRingCalib} and related works~\cite{oneocc,omnitrack++}, which use a panoramic annular camera with a $360^\circ \times 45^\circ$ field of view and a $2048\times2048$ resolution paired with a MID-360 LiDAR, Spheriverse extends the vertical visual coverage while collecting synchronized 128-beam point clouds to support dense 3D annotation.

\textbf{Site selection and route planning.}
Site scouting involved six experienced drivers, discussions with local traffic and urban-planning authorities, and more than $1,300$\,km of driving exploration.
Routes were selected according to traffic function, spatial structure, road geometry, and environmental context.
Collection covered night, dawn, daytime, and evening conditions.
These periods provide varied illumination conditions without implying a uniform distribution across times of day.

\begin{figure}[!t]
\centering
\includegraphics[width=\linewidth]{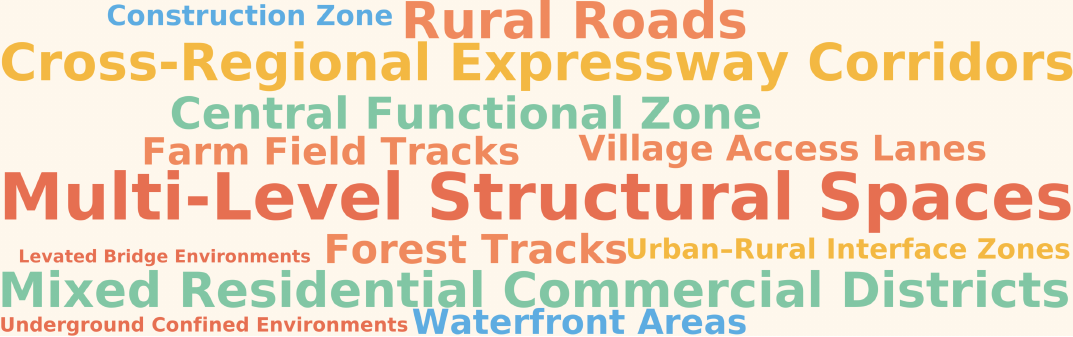}
\caption{Fine-grained scene distribution in Spheriverse. The word cloud summarizes the relative occurrence of the thirteen scene categories; exact split counts are reported in Table~\ref{app:tab:splits}.}
\label{app:fig:scenes}
\end{figure}

\subsection{Scene Taxonomy}
\label{app:scene-taxonomy}
The scene taxonomy groups thirteen fine-grained categories into five major categories according to functional zoning, spatial structure, and traffic characteristics.
The abbreviated category names match the scene-wise benchmark tables, and Fig.~\ref{app:fig:scenes} summarizes the fine-grained scene distribution.

\textbf{Scene annotation.}
Seven annotators established representative examples and independently assigned scene labels to the sequences.
They cross-checked their annotations, discussed disagreements, and resolved ambiguous cases through consensus voting.
This procedure concerns sequence-level scene labels.
Point-cloud semantic labels were produced separately through the outsourced manual annotation pipeline described in the main paper.

\textbf{Urban Core and Mixed-Use Zones.}
These scenes combine dense urban functions with frequent interactions among pedestrians, cyclists, and vehicles.
\textit{1) Central functional zone.} Arterial intersections, business districts, and transportation hubs concentrate traffic flows and complex interactions among traffic participants.
\textit{2) Mixed residential-commercial districts.} Residential and commercial functions coexist along smaller streets, with frequent close-range interactions involving low-speed traffic and static obstacles.

\textbf{Expressway Corridors and Peri-urban Areas.}
These scenes connect urban peripheries with suburban and expressway environments, spanning variations in road openness, building density, and traffic composition.
\textit{1) Cross-regional expressway corridors.} Highways and urban expressways combine structured road layouts with high-speed vehicle motion.
\textit{2) Urban--rural interface zones.} Transitional areas instead feature irregular building distributions, variable road geometry, and heterogeneous traffic participants.

\textbf{Structurally Constrained Transportation Areas.}
These scenes feature complex spatial configurations, structural occlusion, and pronounced illumination variations.
\textit{1) Elevated bridge environments.} Elevated roadways and bridge structures introduce height variations and structural occlusion.
\textit{2) Underground confined environments.} Enclosed spaces, including subterranean parking facilities, combine restricted visibility with illumination changes and localized reflections.
\textit{3) Multi-level structural spaces.} Vertically overlapping lanes and structural elements complicate 3D spatial correspondence across different levels.

\textbf{Rural and Natural Passage Environments.}
These scenes share ambiguous road boundaries, sparse traffic, and strong influences from terrain, vegetation, and natural illumination.
\textit{1) Rural roads.} {Roadways through open countryside emphasize terrain variation and weakly defined road boundaries.}
\textit{2) Forest tracks.} {Passages through wooded areas emphasize vegetation occlusion and restricted driving space.}
\textit{3) Village access lanes.} {Local routes serving villages combine rural road layouts with nearby residential structures.}
\textit{4) Farm field tracks.} {Paths through agricultural land emphasize unstructured surfaces and field boundaries.}

\textbf{Functional and Restricted Operational Areas.}
These scenes are characterized by specialized environmental conditions or operational constraints beyond conventional road layouts.
\textit{1) Waterfront areas.} Roads adjacent to water are exposed to surface reflections that can alter visual appearance.
\textit{2) Construction zones.} Temporary layouts, non-standard signage, irregular obstacles, and interactions between machinery and personnel create changing spatial configurations.

\begin{table}[t]
\centering
\caption{Sequence-level image embedding distances across scene categories.
Diagonal entries report intra-sequence distances averaged within each
scene category, while off-diagonal entries report inter-category distances.
All values are cosine distances scaled by $100$ for easier readability.}
\label{app:tab:distances}
\setlength{\tabcolsep}{0.8pt}
\renewcommand{\arraystretch}{1.15}
\footnotesize
\resizebox{\linewidth}{!}{%
\begin{tabular}{lccccc}
\toprule
\textbf{Scene Category}
& \textbf{Functional}
& \textbf{Rural}
& \textbf{Urban}
& \textbf{Structural}
& \textbf{Expressway} \\
\midrule
Functional & \textbf{{3.8}} & {24.1} & {22.2} & {24.2} & {18.3} \\
Rural      & {24.1} & \textbf{{6.9}} & {27.1} & {28.6} & {24.4} \\
Urban      & {22.2} & {27.1} & \textbf{{4.7}} & {24.9} & {19.8} \\
Structural & {24.2} & {28.6} & {24.9} & \textbf{{5.1}} & {24.0} \\
Expressway & {18.3} & {24.4} & {19.8} & {24.0} & \textbf{{5.2}} \\
\bottomrule
\end{tabular}%
}
\end{table}

\subsection{{Sequence-Level Embedding Analysis}}
\label{app:embeddings}

We analyze image-level representation distances to quantify whether frames from
the same driving sequence remain visually consistent and whether different scene
categories are distinguishable, as shown in Table~\ref{app:tab:distances}.
Let $\mathcal{C}$ denote the five major scene categories, and let
$\mathcal{S}_{c}$ be the set of sequences assigned to category $c\in\mathcal{C}$.
For each sequence $s$, we sample a fixed temporally ordered window of
$T_s=16$ spherical images, denoted as
$\{\bm{I}_{s,t}\}_{t=1}^{T_s}$.

Each image is processed by an ImageNet-pretrained ResNet-50 with the
classification layer removed.
The resulting $2048$-dimensional feature is $\ell_2$-normalized:
\begin{equation}
\bm{f}_{s,t}
=
\frac{
g(\bm{I}_{s,t})
}{
\left\lVert g(\bm{I}_{s,t}) \right\rVert_2
},
\end{equation}
where $g(\cdot)$ denotes the feature extractor.
Thus, $\bm{f}_{s,t}^{\top}\bm{f}_{r,t'}$ gives the cosine similarity between
two spherical images. The cosine distances are multiplied by $100$ for easier readability.

\textbf{Intra-sequence distance.}
For each sequence, we compute the mean pairwise cosine distance among its
sampled frames:
\begin{equation}
d_{s}^{\mathrm{intra}}
=
\frac{2}{T_s(T_s-1)}
\sum_{1\leq t<t'\leq T_s}
\left(
1-\bm{f}_{s,t}^{\top}\bm{f}_{s,t'}
\right).
\end{equation}
The category-level intra-sequence distance is obtained by averaging over all
sequences in the category:
\begin{equation}
D_{c}^{\mathrm{intra}}
=
\frac{100}{|\mathcal{S}_{c}|}
\sum_{s\in\mathcal{S}_{c}}
d_{s}^{\mathrm{intra}} .
\end{equation}
This metric measures visual variation within temporally continuous sequences,
with each sequence contributing equally.

\textbf{Inter-category distance.}
For two different scene categories $c$ and $c'$, we compute the average cosine
distance between their image embeddings:
\begin{equation}
\begin{aligned}
D_{c,c'}^{\mathrm{inter}}
=
\frac{100}{
|\mathcal{S}_{c}|
|\mathcal{S}_{c'}|
}
\sum_{s\in\mathcal{S}_{c}}
\sum_{r\in\mathcal{S}_{c'}}
\frac{1}{T_sT_r}
\sum_{t=1}^{T_s}
\sum_{t'=1}^{T_r}
\left(
1-\bm{f}_{s,t}^{\top}\bm{f}_{r,t'}
\right).
\end{aligned}
\end{equation}

For efficient computation, we first compute the sequence-level and
category-level mean embeddings:
\begin{equation}
\bm{\mu}_{s}
=
\frac{1}{T_s}
\sum_{t=1}^{T_s}
\bm{f}_{s,t},
\qquad
\bm{\mu}_{c}
=
\frac{1}{|\mathcal{S}_{c}|}
\sum_{s\in\mathcal{S}_{c}}
\bm{\mu}_{s}.
\end{equation}
Since these mean embeddings are not re-normalized, the inter-category distance
can equivalently be computed as
\begin{equation}
D_{c,c'}^{\mathrm{inter}}
=
100
\left(
1-\bm{\mu}_{c}^{\top}\bm{\mu}_{c'}
\right),
\qquad c\neq c' .
\end{equation}

Finally, we form a symmetric distance matrix:
\begin{equation}
\bm{M}_{c,c'} =
\begin{cases}
D_{c}^{\mathrm{intra}}, & c=c', \\
D_{c,c'}^{\mathrm{inter}}, & c\neq c' .
\end{cases}
\end{equation}
The diagonal entries therefore describe within-sequence variation aggregated by
scene category, whereas the off-diagonal entries describe distances between
different scene categories.
All distances are multiplied by $100$ for readability. The intra-sequence distances range
from $3.8$ to $6.9$, whereas inter-category distances range from $18.3$
to $28.6$. This contrast indicates that observations within the same
sequence remain relatively consistent in the image embedding space,
while observations from different scene categories exhibit larger
visual differences. Rural and Structural exhibit the largest
inter-category distance ($28.6$), whereas Functional and Expressway
form the closest category pair ($18.3$). Even this smallest
inter-category distance exceeds all reported intra-sequence distances.
Together, these results support the coexistence of within-sequence
visual consistency and cross-category diversity in Spheriverse.

\subsection{Data Splits}
\label{app:splits}
Table~\ref{app:tab:splits} reports the dataset split, which contains
$457$ training sequences, $134$ evaluation sequences, and $53$ test sequences.
Each sequence consists of $100$ temporally continuous frames, corresponding to
$45{,}700$, $13{,}400$, and $5{,}300$ synchronized image--LiDAR pairs,
respectively. The table reports the number of sequences and the proportion of
each fine-grained scene category in each split. Following the nuScenes
evaluation protocol~\cite{Nuscenes}, all benchmark results in the main paper are
reported on the evaluation split.

\begin{table}[ht]
\centering
\footnotesize
\renewcommand{\arraystretch}{1.1}
\caption{Sequence distribution across train, eval (validation), and test. Counts refer to sequences; percentages are normalized within each split.}
\label{app:tab:splits}
% ---------------- Train ----------------
\textbf{Train ($457$ sequences)}
\begin{tabularx}{\columnwidth}{c|>{\raggedright\arraybackslash}X|c|c}
\hline
\textbf{ID} & \textbf{Scenes} & \textbf{Count} & \textbf{Ratio (\%)} \\ \hline
\NO   & central functional zone                 & 39 &  8.53 \\
\NTw  & mixed residential-commercial districts  & 80 & 17.51 \\
\NTh  & cross-regional expressway corridors     & 79 & 17.29 \\
\NFo  & urban--rural interface zones            &  9 &  1.97 \\
\NFiv & waterfront areas                        & 10 &  2.19 \\
\NSi  & construction zones                       &  9 &  1.97 \\
\NSe  & elevated bridge environments            &  2 &  0.44 \\
\NEi  & underground confined environments       &  8 &  1.75 \\
\NNi  & multi-level structural spaces           & 97 & 21.23 \\
\NTe  & rural roads                             & 74 & 16.19 \\
\NEl  & forest tracks                           & 24 &  5.25 \\
\NTww & village access lanes                    & 10 &  2.19 \\
\NThh & farm field tracks                       & 16 &  3.50 \\
\hline
\end{tabularx}

\vspace{1em}

% ---------------- Val ----------------
\textbf{Eval ($134$ sequences)}
\begin{tabularx}{\columnwidth}{c|>{\raggedright\arraybackslash}X|c|c}
\hline
\textbf{ID} & \textbf{Scenes} & \textbf{Count} & \textbf{Ratio (\%)} \\ \hline
\NO   & central functional zone                 & 11 &  8.21 \\
\NTw  & mixed residential-commercial districts  & 22 & 16.42 \\
\NTh  & cross-regional expressway corridors     & 23 & 17.16 \\
\NFo  & urban--rural interface zones            &  3 &  2.24 \\
\NFiv & waterfront areas                        &  3 &  2.24 \\
\NSi   & construction zones                       &  3 &  2.24 \\
\NSe   & elevated bridge environments            &  2 &  1.49 \\
\NEi  & underground confined environments       &  2 &  1.49 \\
\NNi  & multi-level structural spaces           & 27 & 20.15 \\
\NTe  & rural roads                             & 22 & 16.42 \\
\NEl  & forest tracks                           &  6 &  4.48 \\
\NTww & village access lanes                    &  4 &  2.99 \\
\NThh & farm field tracks                       &  6 &  4.48 \\
\hline
\end{tabularx}

\vspace{1em}

% ---------------- Test ----------------
\textbf{Test ($53$ sequences)}
\begin{tabularx}{\columnwidth}{c|>{\raggedright\arraybackslash}X|c|c}
\hline
\textbf{ID} & \textbf{Scenes} & \textbf{Count} & \textbf{Ratio (\%)} \\ \hline
\NO   & central functional zone                 &  4 &  7.55 \\
\NTw  & mixed residential-commercial districts  &  9 & 16.98 \\
\NTh  & cross-regional expressway corridors     & 11 & 20.75 \\
\NNi  & multi-level structural spaces           & 12 & 22.64 \\
\NTe  & rural roads                             & 16 & 30.19 \\
\NEl  & forest tracks                           &  1 &  1.89 \\
\hline
\end{tabularx}

\end{table}

\begin{table*}[ht]
\centering
\caption{Mapping from fine-grained attributes to intermediate and unified semantic classes.}
\label{app:tab:taxonomy}

\renewcommand{\arraystretch}{1.05}
\setlength{\tabcolsep}{3pt}
\footnotesize
\newcommand{\classcline}[1]{\cmidrule(lr){#1}}
\begin{tabular*}{0.98\textwidth}{@{\extracolsep{\fill}}p{5.0cm}|c|c|c|c|c@{}}
\toprule
\textbf{Fine-grained attribute} &
\textbf{Pct.} &
\textbf{Intermediate class} &
\textbf{Pct.} &
\textbf{Unified class} &
\textbf{Pct.} \\
\midrule

1 Pedestrian & {0.0520\%} & 1. Pedestrian & {0.0520\%} & 1. Pedestrian & {0.0520\%} \\
\midrule

2 Passenger vehicle    & {9.6620\%} & \multirow{3}{*}{2. Vehicle} & \multirow{3}{*}{{10.0105\%}} & \multirow{3}{*}{2. Vehicle} & \multirow{3}{*}{{10.0105\%}} \\
\classcline{1-2}
3 Heavy vehicle        & {0.2452\%} &                             &                       &                            &                       \\
\classcline{1-2}
4 Construction vehicle & {0.1033\%} &                             &                       &                            &                       \\
\midrule

5 Moving bicycle        & {0.0015\%} & \multirow{7}{*}{3. Cyclist} & \multirow{7}{*}{{0.2242\%}} & \multirow{7}{*}{3. Cyclist} & \multirow{7}{*}{{0.2242\%}} \\
\classcline{1-2}
6 Stationary bicycle    & {0.0066\%} &                             &                       &                            &                       \\
\classcline{1-2}
7 Bicycle group         & {0.0015\%} &                             &                       &                            &                       \\
\classcline{1-2}
8 Moving motorcycle     & {0.0559\%} &                             &                       &                            &                       \\
\classcline{1-2}
9 Stationary motorcycle & {0.0828\%} &                             &                       &                            &                       \\
\classcline{1-2}
10 Motorcycle group     & {0.0028\%} &                             &                       &                            &                       \\
\classcline{1-2}
11 Other cyclist        & {0.0731\%} &                             &                       &                            &                       \\
\midrule

12 Permanent building & {15.7269\%} & \multirow{2}{*}{4. Building} & \multirow{2}{*}{{15.8154\%}} & \multirow{9}{*}{4. Building} & \multirow{9}{*}{{18.0318\%}} \\
\classcline{1-2}
13 Temporary building & {0.0885\%} &                              &                    &                              &                    \\
\classcline{1-4}
14 Bridge             & {0.2767\%} & 5. Bridge                    & {0.2767\%}              &                              &                    \\
\classcline{1-4}
15 Tunnel             & {1.2082\%} & 6. Tunnel                    & {1.2082\%}              &                              &                    \\
\classcline{1-4}
16 Advertising object & {0.3067\%} & \multirow{4}{*}{12. Man-made} & \multirow{4}{*}{{0.5802\%}} &                              &                    \\
\classcline{1-2}
17 Art object         & {0.0298\%} &                               &                    &                              &                    \\
\classcline{1-2}
18 Functional object  & {0.2073\%} &                               &                    &                              &                    \\
\classcline{1-2}
19 Other man-made object & {0.0364\%} &                            &                    &                              &                    \\
\classcline{1-4}
20 Pavilion           & {0.1526\%} & 13. Pavilion                  & {0.1526\%}              &                              &                    \\
\midrule

21 Vegetation & {26.6991\%} & 7. Vegetation & {26.6991\%} & 5. Vegetation & {26.6991\%} \\
\midrule

22 Bollard       & {2.4379\%} & \multirow{4}{*}{11. Barrier} & \multirow{4}{*}{{6.6195\%}} & \multirow{5}{*}{6. Pole \& Barrier} & \multirow{5}{*}{{7.3890\%}} \\
\classcline{1-2}
23 Fence         & {3.7391\%} &                              &                    &                                      &                    \\
\classcline{1-2}
24 Curb          & {0.4326\%} &                              &                    &                                      &                    \\
\classcline{1-2}
25 Other barrier & {0.0099\%} &                              &                    &                                      &                    \\
\classcline{1-4}
26 Pole          & {0.7695\%} & 8. Pole                      & {0.7695\%}              &                                      &                    \\
\midrule

27 Drivable road     & {28.4774\%} & \multirow{2}{*}{9. Road} & \multirow{2}{*}{{30.9190\%}} & \multirow{2}{*}{7. Road} & \multirow{2}{*}{{30.9190\%}} \\
\classcline{1-2}
28 Non-drivable road & {2.4416\%} &                           &                    &                         &                    \\
\midrule

29 Vegetated surface & {0.9616\%} & \multirow{4}{*}{10. Surface} & \multirow{4}{*}{{2.8002\%}} & \multirow{4}{*}{8. Surface} & \multirow{4}{*}{{2.8002\%}} \\
\classcline{1-2}
30 Farmland          & {1.5752\%} &                              &                    &                             &                    \\
\classcline{1-2}
31 Gravel            & {0.1357\%} &                              &                    &                             &                    \\
\classcline{1-2}
32 Other surface     & {0.1277\%} &                              &                    &                             &                    \\
\midrule

33 Other & {3.8728\%} & 14. Other & {3.8728\%} & 9. Other & {3.8728\%} \\
\bottomrule
\end{tabular*}
\end{table*}

\subsection{Semantic Annotation Hierarchy}
\label{app:taxonomy}
The annotation hierarchy organizes traffic participants, built structures, and natural environments at different levels of detail.
In our experiments, we follow common semantic occupancy protocols~\cite{oneocc,surroundocc,monoscene} and use nine unified classes for evaluation.
Table~\ref{app:tab:taxonomy} indicates the mapping from fine-grained attributes to the intermediate and unified taxonomies.
In the benchmark tables, Person, Bike, and Pillar correspond to Pedestrian, Cyclist, and Pole \& Barrier, respectively.
The hierarchy covers common categories such as vehicles, buildings, vegetation, and roads, while further distinguishing fine-grained classes such as construction vehicles, temporary buildings, advertising objects, art objects, pavilions, curbs, and gravel.
This fine-grained taxonomy also provides a basis for downstream tasks such as abnormal-event detection and rare-object detection. %Figure~\ref{app:fig:taxonomy} intuitively illustrates the hierarchical organization of Spheriverse annotations, with colored links tracing the aggregation of fine-grained categories into intermediate and unified semantic classes.

% \begin{figure*}[ht]
% \centering
% \includegraphics[width=0.96\textwidth]{Fig6.pdf}
% \caption{Three-level semantic annotation hierarchy in Spheriverse. From bottom to top, fine-grained categories are aggregated into intermediate and unified semantic classes. Colored links indicate category mappings across levels, while the bottom-row numbers denote fine-grained class IDs.}
% \label{app:fig:taxonomy}
% \vspace{-1em}
% \end{figure*}

\section{Implementation Details}
\label{app:method}
\subsection{Training and Evaluation Settings}
\label{app:settings}
All baselines follow their official configurations with minimal dataset-specific adaptations.
Images are resized to $2480\times512$ pixels, and models are trained for $28$ epochs on four NVIDIA RTX 3090 GPUs.
Within each benchmark, the training schedule, input resolution, and output discretization are standardized.
Each method retains its core representation, task-specific losses, and remaining configuration settings.

Semantic occupancy uses a $200\times200\times16$ voxel grid with $0.5$\,m resolution and nine semantic classes.
BEV semantic mapping uses a $200\times200$ grid and seven semantic classes.
Both overall and scene-wise results use the eval split described in Sec.~\ref{app:splits}.
The three-dimensional detection benchmark follows the nuScenes evaluation protocol~\cite{Nuscenes} and reports mAP and NDS.

\subsection{Metrics}
\label{app:metrics}
For semantic class $c$, intersection over union is
\begin{equation}
\mathrm{IoU}_c=\frac{\mathrm{TP}_c}
{\mathrm{TP}_c+\mathrm{FP}_c+\mathrm{FN}_c},
\end{equation}
where $\mathrm{TP}_c$, $\mathrm{FP}_c$, and $\mathrm{FN}_c$ are class-specific true positives, false positives, and false negatives.
Mean IoU averages across the evaluated semantic classes:
\begin{equation}
\mathrm{mIoU}=\frac{1}{C}\sum_{c=1}^{C}\mathrm{IoU}_c.
\end{equation}
Here, $C=9$ for occupancy and $C=7$ for BEV semantic mapping.
Empty voxels are not treated as an additional semantic class in the occupancy mean.

GeoIoU measures class-agnostic occupancy overlap:
\begin{equation}
\mathrm{GeoIoU}=
\frac{|V_{\mathrm{pred}}\cap V_{\mathrm{gt}}|}
{|V_{\mathrm{pred}}\cup V_{\mathrm{gt}}|},
\end{equation}
where $V_{\mathrm{pred}}$ and $V_{\mathrm{gt}}$ denote predicted and ground-truth occupied voxel sets.
Occupancy and mapping metrics are reported in percentages.
Detection mAP and NDS are reported following~\cite{Nuscenes}.

\subsection{Additional Ablation Studies} \label{sec:supp_method_details}

\subsubsection{Clarification of Motivation.}
\label{app:initial_lifting}
The initial lifting follows the query-based deformable cross-attention design
of SurroundOcc~\cite{surroundocc}. We clarify why the voxel queries should interact with the Cartesian
and spherical coordinates. \textit{Do the UV coordinates interact with the voxel queries? (Initial Lifting)}
Yes, but only indirectly. The projected UV coordinate serves as a
reference anchor rather than a feature fused into the query. Each voxel
query predicts sampling offsets and attention weights around this anchor
to retrieve the corresponding spherical features. \textit{Motivation for CSRR and SER.}
Consequently, the initial lifting uses image geometry primarily to locate
the reference anchor, without explicitly encoding spherical geometric
relations into the lifted voxel features. We therefore introduce CSRR to
embed sphere-aligned range--azimuth relations into the Cartesian voxel
representation, yielding Cartesian--spherical voxel features. Building on
these features, SER incorporates range--height--azimuth geometry into the queries and reuses the same UV anchors to adaptively retrieve
complementary semantic evidence from the source features. Overall, \textit{The difference between initial lifting and CSRR/SER.} Intuitively, initial lifting follows an image-driven mapping, where features indexed by image coordinates $(u,v)$ are projected into 3D Cartesian space $(x,y,z)$. In contrast, CSRR/SER follows a geometry-driven construction, where panoramic coordinates $(\rho,\theta)$ are used to explicitly construct locations in 3D Cartesian space. Therefore, initial lifting performs $(u,v)\rightarrow(x,y,z)$, whereas CSRR/SER performs $(\rho,\theta)\rightarrow(x,y,z)$. Thanks to CSRR and SER, SphereOcc achieves the joint modeling of angular observations and metric voxels.

\begin{figure}[!h]
\centering
\includegraphics[width=0.84\linewidth]{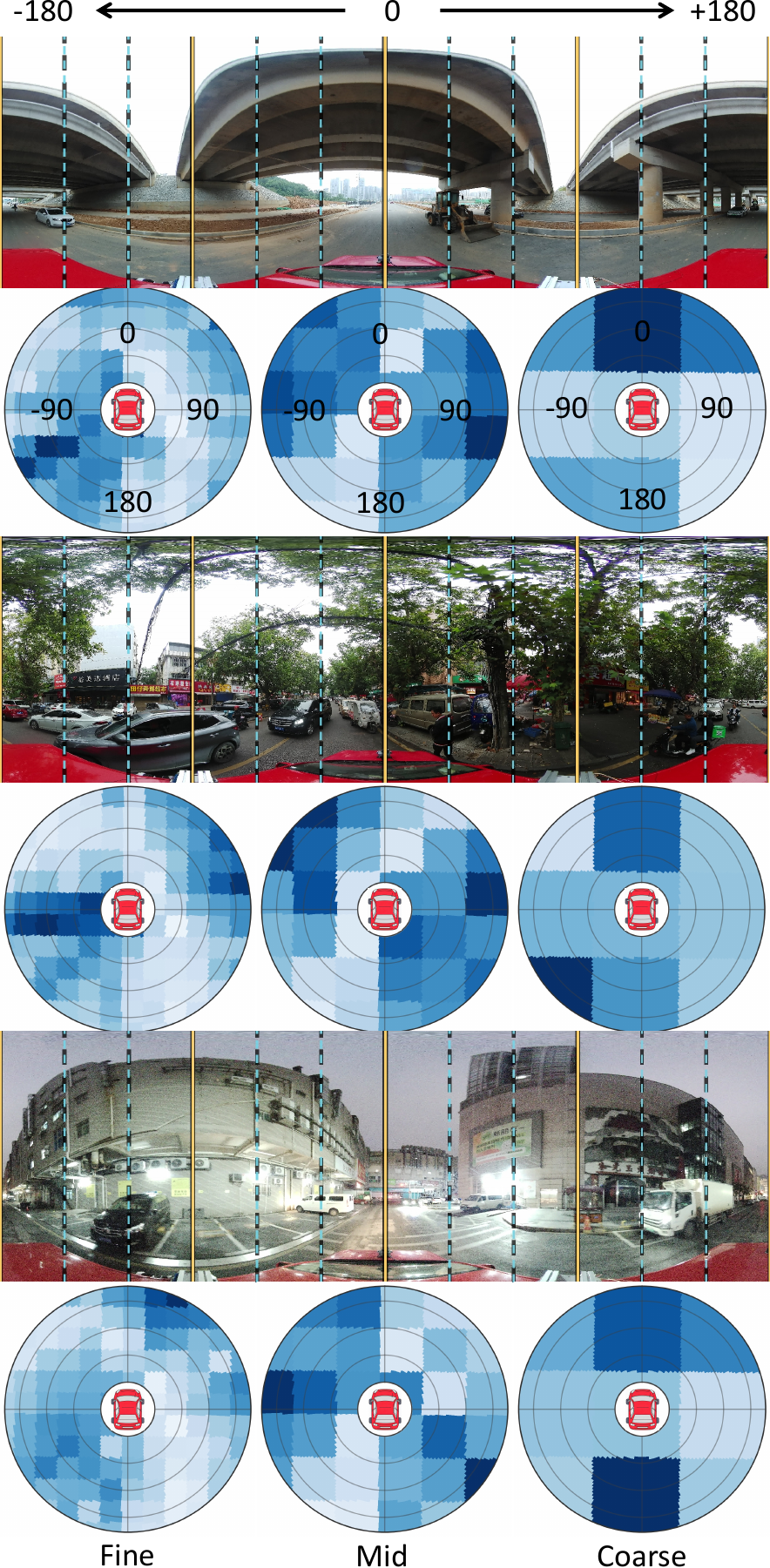}
\caption{Multi-scale CSRR response visualization. For each spherical input, we show the feature-change response maps before and after CSRR at the fine, middle, and coarse scales. The circular direction represents the full $0^\circ$--$360^\circ$ azimuth range around the ego vehicle, and the radial direction represents the horizontal distance $\rho=\sqrt{x^2+y^2}$ from each voxel center to the ego-centric vertical axis. Since the occupancy volume contains 16 vertical voxel layers, responses are averaged along the height dimension for visualization. Blue shading indicates the magnitude of feature update.}
\label{app:fig:csrr}
\end{figure}

\subsubsection{Ablation for CSRR}
\label{app:responses}
To verify the effectiveness of CSRR, we visualize the magnitude of voxel-feature changes before and after the CSRR module.
As shown in Fig.~\ref{app:fig:csrr}, the circular direction around the ego vehicle denotes the full $0^\circ$--$360^\circ$ azimuth $\theta$, while the radial direction denotes the horizontal distance from each voxel center to the vertical ego-centric axis, i.e., $\rho=\sqrt{x^2+y^2}$.
Since the occupancy volume contains 16 vertical voxel layers, we average the response values over these height layers for visualization, resulting in a top-down CSRR response map.
The response is computed as the feature-change magnitude between the input and output voxel features of CSRR.
The visualization shows that voxel features exhibit a range-dependent remodeling pattern, with stronger responses appearing progressively from near to far regions.
This supports the role of CSRR in jointly encoding Cartesian coordinates $(x,y,z)$ with spherical cues $(\sin\theta,\cos\theta,\rho)$, thereby enriching Cartesian voxel features with sphere-aware geometry.
Such geometry-conditioned remodeling guides the voxel representation to better adapt to the spherical observation geometry, rather than applying a uniform feature update across all spatial locations.

\subsubsection{Ablation for SER}
\label{sec:ablation_ser}

\begin{table*}[ht]
\centering
\caption{
Ablation study of spherical evidence re-querying on SphereOcc.
The table reports overall and class-wise semantic occupancy results. 
All metrics are reported in \%, and higher values indicate better performance.
Base. denotes the original SurroundOcc~\cite{surroundocc} baseline.
\ding{172}, \ding{173}, and \ding{174} denote Cartesian XYZ re-querying, spherical evidence re-querying (SER), and Cartesian--Spherical Representation Remodeling (CSRR), respectively.
Pers., Veh., Bldg., Veg., Pill., and Surf. denote Person, Vehicle, Building, Vegetation, Pillar, and Surface, respectively.
}
\label{tab:ser_ablation}

\begingroup

\definecolor{abperson}{RGB}{156,39,176}
\definecolor{abvehicle}{RGB}{255,167,38}
\definecolor{abbike}{RGB}{66,133,244}
\definecolor{abbuilding}{RGB}{126,87,194}
\definecolor{abvegetation}{RGB}{67,160,71}
\definecolor{abpillar}{RGB}{141,110,99}
\definecolor{abroad}{RGB}{84,110,122}
\definecolor{absurface}{RGB}{176,190,197}
\definecolor{abothers}{RGB}{236,64,122}
\definecolor{absectioncolor}{RGB}{247,248,250}

\newcommand{\abClassHead}[2]{%
\mbox{%
\raisebox{-0.2ex}{\textcolor{#1}{{\large$\bullet$}}}\,#2%
}%
}
\newcommand{\abMIoU}{\mbox{mIoU~$\uparrow$}}
\newcommand{\abGeoIoU}{\mbox{GeoIoU~$\uparrow$}}

\setlength{\tabcolsep}{1.5pt}
\renewcommand{\arraystretch}{1.10}
\scriptsize

\resizebox{0.99\textwidth}{!}{%
\begin{tabular}{*{4}{c}|cc|*{9}{c}}
\toprule
\rowcolor{absectioncolor}
\multicolumn{15}{c}{
\textbf{Overall and class-wise semantic occupancy results}
} \\
\midrule

\multicolumn{4}{c|}{\textbf{Configuration}}
& \multicolumn{2}{c|}{\textbf{Overall}}
& \multicolumn{9}{c}{\textbf{Class-wise IoU}} \\

\cmidrule(lr){1-4}
\cmidrule(lr){5-6}
\cmidrule(lr){7-15}

Base.
& \ding{172}
& \ding{173}
& \ding{174}
& \abMIoU
& \abGeoIoU
& \abClassHead{abperson}{Pers.}
& \abClassHead{abvehicle}{Veh.}
& \abClassHead{abbike}{Bike}
& \abClassHead{abbuilding}{Bldg.}
& \abClassHead{abvegetation}{Veg.}
& \abClassHead{abpillar}{Pill.}
& \abClassHead{abroad}{Road}
& \abClassHead{absurface}{Surf.}
& \abClassHead{abothers}{Other} \\
\midrule

\checkmark & -- & -- & --
& 12.05 & 22.55
& 1.65 & 13.97 & 5.16 & 9.40 & 14.08
& 12.57 & 36.58 & 11.99 & 3.07 \\

\checkmark & \checkmark & -- & --
& 12.38 & 23.09
& 1.52 & 14.65 & 6.27 & 9.17 & 14.68
& 12.85 & 37.40 & 11.85 & 3.06 \\

\checkmark & -- & \checkmark & --
& 12.49 & 23.16
& 1.54 & 14.54 & 6.68 & 9.42 & 14.64
& 13.17 & 37.65 & 11.75 & 2.98 \\

\checkmark & -- & \checkmark & \checkmark
& \textbf{13.17} & \textbf{23.85}
& \textbf{2.72} & \textbf{15.50} & \textbf{6.62}
& \textbf{9.99} & \textbf{15.01}
& \textbf{14.55} & \textbf{38.47}
& \textbf{12.39} & \textbf{3.32} \\

\bottomrule
\end{tabular}%
}
\endgroup
\end{table*}

\begin{figure*}[ht!]
\centering
\includegraphics[width=0.99\textwidth]{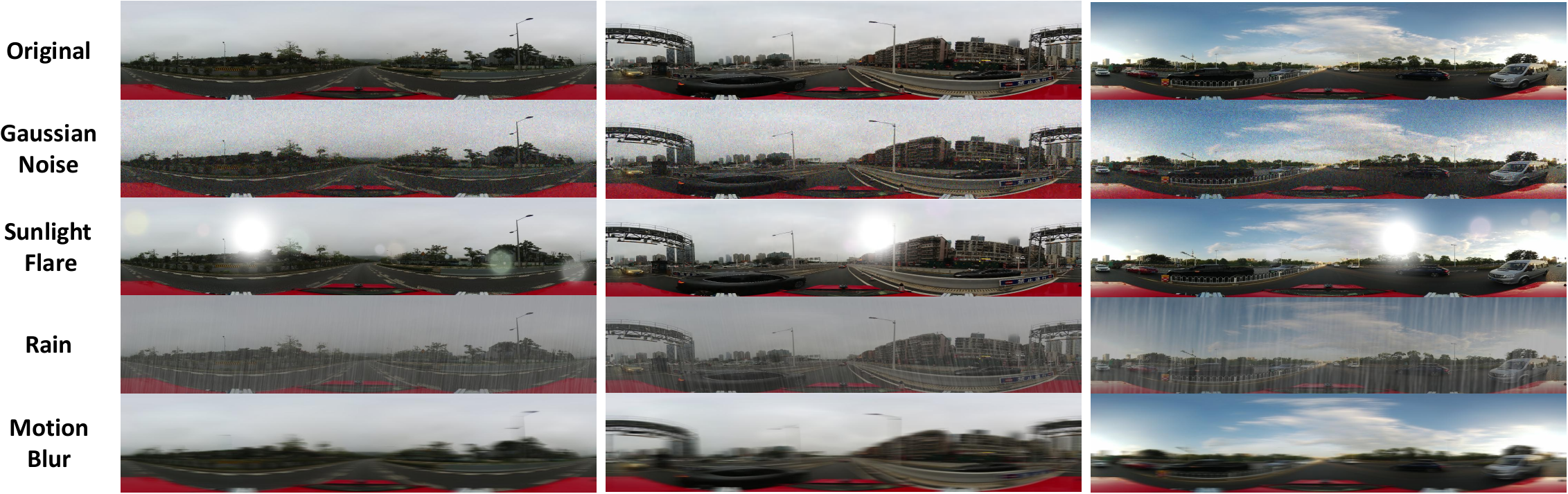}
\caption{
Visual examples of synthetic image degradations on Spheriverse.
Each column shows one representative spherical image, and each row corresponds
to a different input condition: original image, Gaussian noise, sunlight/lens
flare, large raindrops, and motion blur.
}
\label{fig:synthetic_corruptions}
\end{figure*}

To verify the effectiveness of the spherical descriptor
$[\rho_{\ell,n}, z_{\ell,n}, \sin\theta_{\ell,n}, \cos\theta_{\ell,n}]$
in SER, we compare it with Cartesian XYZ re-querying under the same
SurroundOcc~\cite{surroundocc} baseline. The quantitative results are reported in Table~\ref{tab:ser_ablation}. Replacing
Cartesian XYZ with the spherical descriptor improves the overall mIoU from
$12.38\%$ to $12.49\%$ and GeoIoU from $23.09\%$ to $23.16\%$. The results indicate that
spherical range and azimuth information provides a more suitable geometric
prior for spherical feature retrieval than Cartesian coordinates alone. When further combined with CSRR, SER achieves $13.17\%$ mIoU and $23.85\%$
GeoIoU, demonstrating the complementary effect of spherical evidence re-querying
and Cartesian--spherical voxel remodeling. CSRR enriches the lifted voxel
features with explicit spherical range--azimuth geometry, making the Cartesian
voxel representation more aware of the spherical observation. SER
then addresses the remaining evidence-missing problem by re-querying the source
image features for each voxel, allowing the model to retrieve additional
semantic cues that may be weakly captured during the initial lifting stage.
Together, the two modules improve both geometric alignment and semantic
evidence aggregation in the spherical occupancy representation.

\section{Extended Ablation Studies and Analyses}
\label{Extended_Ablation}
\subsection{Robustness under Visual Degradations}
\label{app:Robustness_under_Visual_Degradations}

To assess the robustness of semantic occupancy prediction under degraded
spherical observations, we evaluate different models under four representative
image degradations, including Gaussian noise, large raindrops, sunlight/lens
flare, and motion blur. All degradations are applied to the validation
panoramas of Spheriverse. Specifically, we use Gaussian noise with
$\sigma=0.18$, large raindrops with a density of $0.10$--$0.15$,
sunlight/lens flare with ten secondary highlights, and motion blur with a
kernel size of 74. For a fair comparison, all methods are evaluated with
the same corrupted observations. Fig.~\ref{fig:synthetic_corruptions}
provides representative examples of these degradations. The degradations cover
both local visibility degradation, such as raindrop occlusion and motion blur,
and global appearance shifts, such as Gaussian noise and sunlight/lens flare,
thereby testing whether the model can preserve semantic and geometric reasoning
under different types of visual perturbations.

\begin{table*}[ht]
\centering
\caption{
Overall robustness comparison of semantic occupancy prediction under synthetic
image corruptions on Spheriverse.
Each entry reports mIoU / GeoIoU in \%.
TPVFo., Surro., MonoS., and Quadr. denote TPVFormer~\cite{TPVFormer},
SurroundOcc~\cite{surroundocc}, MonoScene~\cite{monoscene}, and
QuadricFormer~\cite{quadricformer}, respectively.
Gaussian noise, large rain drops, sunlight/lens flare, and motion blur are
applied deterministically, such that all methods receive identical corrupted
observations.
}
\label{tab:noise_overall_robustness}

\begingroup

\newcommand{\robpair}[2]{\mbox{#1\,/\,#2}}
\newcommand{\robbest}[2]{\textbf{\robpair{#1}{#2}}}

\setlength{\tabcolsep}{3.6pt}
\renewcommand{\arraystretch}{1.10}
\scriptsize

\resizebox{0.95\textwidth}{!}{%
\begin{tabular}{lccccc}
\toprule
\textbf{Method}
& \textbf{Clean}
& \textbf{Gaussian}
& \textbf{Large Rain Drops}
& \textbf{Sunlight/Flare}
& \textbf{Motion Blur} \\
\midrule

TPVFo.
& \robpair{12.21}{22.35}
& \robpair{6.53}{15.71}
& \robpair{6.39}{16.73}
& \robpair{11.74}{21.85}
& \robpair{4.96}{12.46} \\

Surro.
& \robpair{12.05}{22.55}
& \robpair{7.74}{17.54}
& \robpair{8.38}{18.37}
& \robpair{11.82}{22.19}
& \robpair{6.48}{15.35} \\

MonoS.
& \robpair{11.44}{21.68}
& \robpair{4.99}{13.10}
& \robpair{5.26}{13.08}
& \robpair{11.04}{21.21}
& \robpair{2.89}{8.89} \\

Quadr.
& \robpair{11.09}{20.76}
& \robpair{6.14}{13.57}
& \robpair{6.40}{15.17}
& \robpair{10.63}{20.08}
& \robpair{3.89}{9.31} \\

\textbf{Ours}
& \robbest{13.91}{24.65}
& \robbest{11.24}{21.26}
& \robbest{11.87}{22.21}
& \robbest{13.68}{24.31}
& \robbest{7.71}{17.14} \\

\bottomrule
\end{tabular}%
}

\endgroup
\end{table*}

As shown in Table~\ref{tab:noise_overall_robustness}, our method consistently
outperforms the compared baselines under both clean and corrupted inputs.
On clean images, our method improves the best baseline from $12.21\%$ to
$13.91\%$ in mIoU and from $22.55\%$ to $24.65\%$ in GeoIoU. Under Gaussian
noise, the best baseline obtains $7.74\%$ mIoU and $17.54\%$ GeoIoU, whereas
our method achieves $11.24\%$ mIoU and $21.26\%$ GeoIoU. Under large
raindrops, our method improves the best baseline from $8.38\%$ to $11.87\%$
in mIoU and from $18.37\%$ to $22.21\%$ in GeoIoU. Under sunlight/lens flare,
the improvement is from $11.82\%$ to $13.68\%$ in mIoU and from $22.19\%$ to
$24.31\%$ in GeoIoU. Motion blur leads to the most severe degradation across
all methods; nevertheless, our method still improves the best baseline from
$6.48\%$ to $7.71\%$ in mIoU and from $15.35\%$ to $17.14\%$ in GeoIoU. 

These results indicate that the proposed spherical occupancy representation
benefits robustness in two complementary aspects. First, after the initial
lifting, CSRR injects explicit spherical range--azimuth geometry into the
Cartesian voxel features, enabling the voxel representation to better preserve
the spatial organization of spherical observations. Second, SER re-queries the
source spherical features according to the remodeled voxel representation,
allowing each voxel to retrieve additional semantic evidence from geometrically
relevant image regions. Consequently, the proposed representation improves the
consistency of voxel features and strengthens semantic evidence aggregation
under noise, weather artifacts, illumination changes, and motion-induced
degradation.

\begin{figure}[ht]
\centering
\includegraphics[width=\columnwidth]{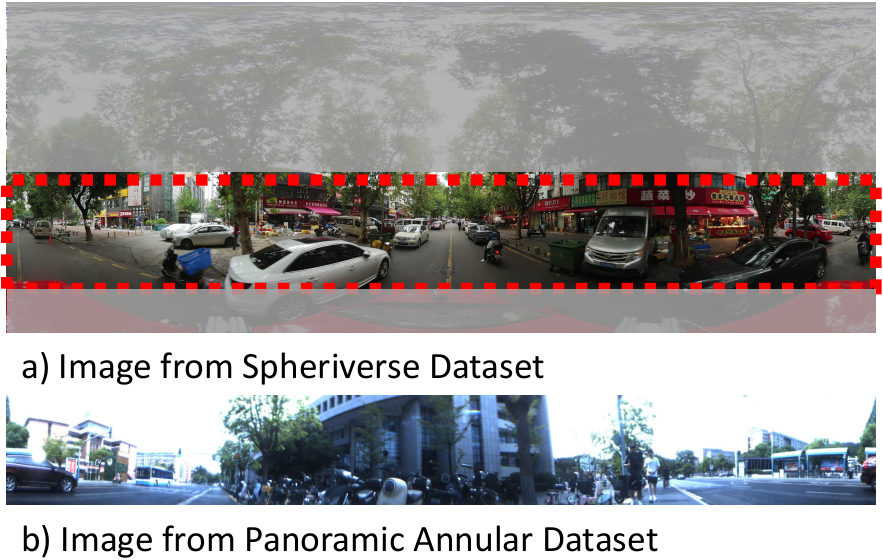}
\caption{
Comparison between Spheriverse spherical imagery and panoramic annular imagery.
(a) A spherical panorama from Spheriverse, which provides broader vertical
field-of-view coverage. (b) A representative panoramic annular image. The red
dashed box indicates the perceptual range of the panoramic annular dataset when
mapped onto the Spheriverse spherical panorama. ($136.7^\circ$ for Spheriverse and $45^\circ$ for previous works~\cite{oneocc,PanoRingCalib,luo2025omnitrack,gao2022review}.)
}
\label{fig:dataset_compare}
\end{figure}

\subsection{Comparison of Panoramic Annular Datasets} \label{sec:ring_pano_comparison}

Annular panoramic datasets have been widely used in recent perception studies~\cite{oneocc,PanoRingCalib,luo2025omnitrack,gao2022review}, as they provide complete $360^\circ$ horizontal observations. 
Compared with existing annular datasets, Spheriverse offers the following advantages: \textbf{Expanded vertical FoV.}
Spheriverse adopts a $136.7^\circ$ vertical FoV, which is substantially wider than the vertical coverage of existing annular panoramic datasets. 
As shown in Fig.~\ref{fig:dataset_compare}, the red dashed box indicates the perceptual range of panoramic annular imaging when mapped onto a Spheriverse spherical panorama. 
The comparison shows that annular imagery mainly covers a limited vertical band, whereas Spheriverse captures broader ground-level, surrounding,
and elevated regions within a single spherical observation. This expanded
coverage better matches the requirement of embodied agents for holistic spatial
understanding. \textbf{High-quality image acquisition.}
Spheriverse uses Sony 8MP starlight-level sensors with high-quality fisheye
lenses, providing clearer spherical images under diverse illumination
conditions. This is particularly important for nighttime, sunrise, sunset, and
other challenging lighting scenarios, where low image quality may directly
degrade downstream 3D perception. \textbf{Diverse scenes and semantic annotations.}
Spheriverse spans $13$ fine-grained scene types and provides a richer semantic
taxonomy with $33$ annotated classes. This diversity enables systematic
evaluation across expressways, rural roads, structural spaces, urban regions,
and functional areas, supporting more realistic assessment of spherical
perception under open-world scene distributions. \textbf{High-precision LiDAR sensing.}
Spheriverse incorporates a 128-beam LiDAR to provide dense and accurate 3D
geometric supervision. In contrast, existing annular datasets often rely on
lower-resolution geometric sensors or are primarily designed for 2D visual
understanding, which limits their ability to support reliable evaluation of 3D
spatial reasoning and semantic occupancy prediction. Table~\ref{tab:dataset_comparison_ring_based} summarizes representative panoramic annular datasets.

\begin{table*}[!t]
\centering
\caption{
{Comparison of recent ring-based panoramic datasets.
The acquisition-type icon denotes data acquired from real-world scenes; check
and cross icons indicate whether an attribute is available.
The four lighting icons denote sunrise, daytime, sunset, and evening
conditions.
`\textit{\#T.Scenes}', `\textit{\#Classes}', and `\textit{\#Frames}'
denote the numbers of scene types, semantic classes, and frames, respectively.
`3D Anno.' indicates the availability of 3D annotations.
`B.' denotes the number of LiDAR beams.
`LiDAR Range' reports the horizontal annotation range in the $x$--$y$ plane,
and `Voxel Grid' denotes the dimensions of the occupancy grid.}
}
\label{tab:dataset_comparison_ring_based}

\begingroup
\newcommand{\bigtableicon}[1]{%
\raisebox{-0.35ex}{\includegraphics[height=1.65em]{#1}}%
}
\newcommand{\bigmapo}{\bigtableicon{figs/icons/mapo.png}}
\newcommand{\bigrighticon}{\bigtableicon{figs/icons/right2.png}}
\newcommand{\bigwrongicon}{\bigtableicon{figs/icons/wrong5.png}}
\newcommand{\bigsr}{\bigtableicon{figs/icons/sr.png}}
\newcommand{\bigda}{\bigtableicon{figs/icons/day.png}}
\newcommand{\bigsd}{\bigtableicon{figs/icons/sd.png}}
\newcommand{\bigev}{\bigtableicon{figs/icons/ev.png}}
\newcommand{\biglidar}{\bigtableicon{figs/icons/lidar.png}}

\renewcommand{\arraystretch}{1.25}
\setlength{\tabcolsep}{4.2pt}
\small

\resizebox{0.98\textwidth}{!}{%
\begin{tabular}{l|ccc|cccc|c|ccc}
\toprule
\multicolumn{1}{c|}{\multirow{2}{*}[-0.7ex]{\textbf{Datasets}}}
& \multicolumn{3}{c|}{\textbf{Acquisition}}
& \multicolumn{4}{c|}{\textbf{Diversity}}
& \multicolumn{1}{c|}{\multirow{2}{*}[-0.7ex]{\textbf{3D Anno.}}}
& \multicolumn{3}{c}{\textbf{Scale}} \\
\cmidrule(lr){2-4}
\cmidrule(lr){5-8}
\cmidrule(lr){10-12}
& \textit{Year}
& \textit{Type}
& \textit{Geo. Sen.}
& \textit{\#T.Scenes}
& \textit{Weather}
& \textit{Lighting}
& \textit{\#Classes}
&
& \textit{\#Frames}
& \textit{LiDAR Range}
& \textit{Voxel Grid} \\
\midrule

QuadOcc~\cite{oneocc}
& 2025
& \bigmapo
& $<16$ B.
& 4
& \bigwrongicon
& \bigda\ \bigsd\ \bigev
& 6
& \bigrighticon
& 24K
& $25.6{\times}25.6$ m
& $64{\times}64{\times}8$ \\

PanoMMOcc~\cite{PanoRingCalib}
& 2026
& \bigmapo
& $<16$ B.
& 6
& \bigwrongicon
& \bigda\ \bigsd\ \bigev
& 15
& \bigrighticon
& 21.6K
& $25.6{\times}25.6$ m
& $64{\times}64{\times}16$ \\

OmniTrack~\cite{luo2025omnitrack}
& 2025
& \bigmapo
& \bigwrongicon
& 3
& \bigwrongicon
& \bigda
& 2
& \bigwrongicon
& 19.2K
& --
& -- \\

\specialrule{0.08em}{0ex}{0ex}
\rowcolor{yellow!8}
\textbf{Ours}
& 2026
& \bigmapo
& \textbf{128 B.}, \biglidar
& \textbf{13}
& \bigrighticon
& \bigsr\ \bigda\ \bigsd\ \bigev
& \textbf{33}
& \bigrighticon
& \textbf{64.4K}
& $\mathbf{100{\times}100}$ m
& $\mathbf{200{\times}200{\times}16}$ \\
\bottomrule
\end{tabular}%
}
\endgroup
\end{table*}

\begin{figure*}[h!]
\centering
\includegraphics[width=0.96\textwidth]{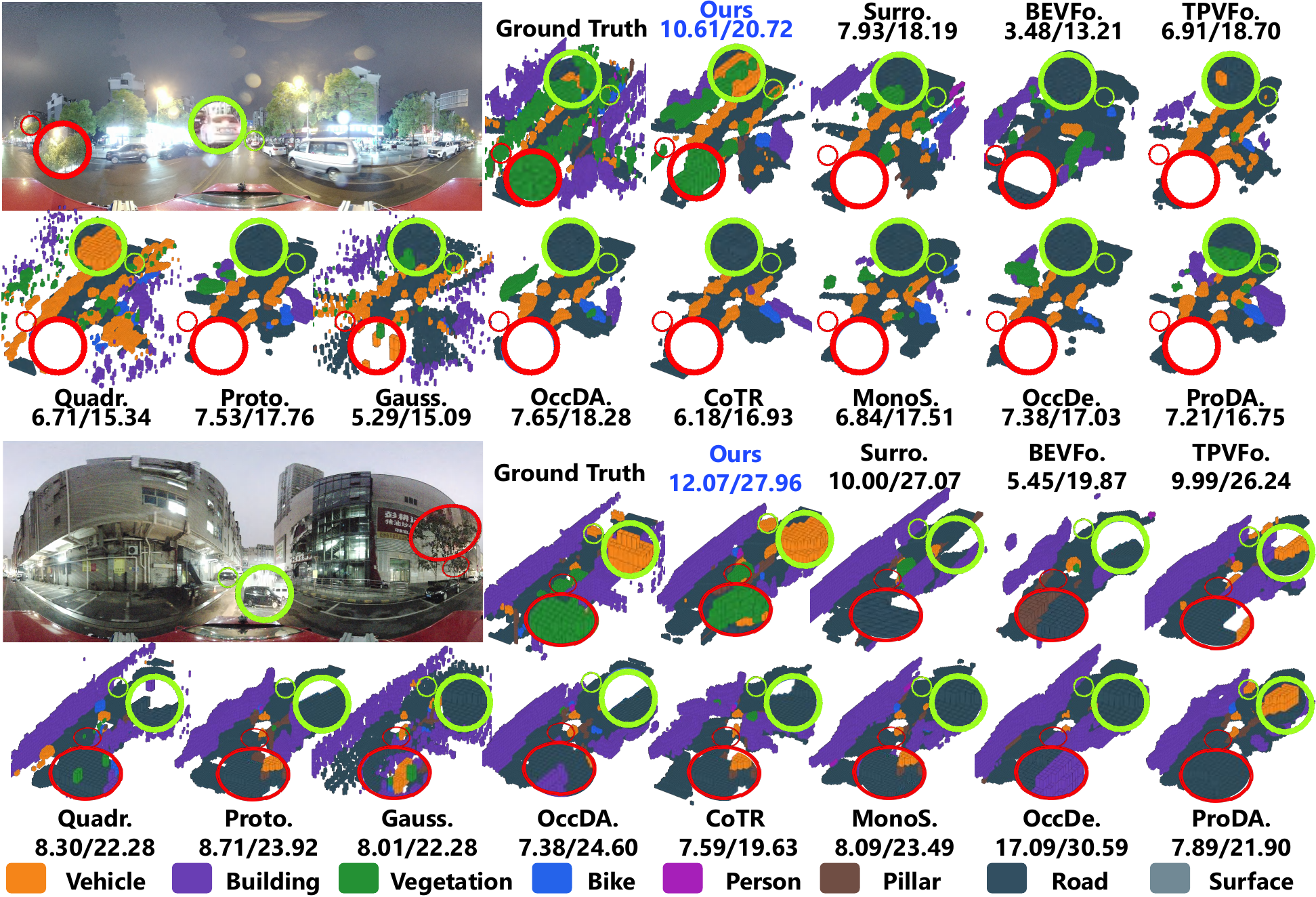}
\caption{
Qualitative results of semantic occupancy prediction on Spheriverse.
Each example shows the spherical input observation and the corresponding
3D semantic occupancy prediction. The mIoU and GeoIoU scores (in \%) are reported above each prediction.
Surro., BEVFo., TPVFo., Quadr., Proto., Gauss., CoTR, MonoS., and OccDe. denote SurroundOcc~\cite{surroundocc}, BEVFormer~\cite{bevformer}, TPVFormer~\cite{TPVFormer}, QuadricFormer~\cite{quadricformer}, ProtoOcc~\cite{protoocc}, GaussianFormer~\cite{gaussianformer}, Compact Occupancy Transformer~\cite{cotr}, MonoScene~\cite{monoscene}, and OccDepth~\cite{occdepth}, respectively.
OccDA. and ProDA. denote OccDepth and ProtoOcc augmented with Depth Anything~\cite{depth}, respectively.
Colors represent the semantic classes shown in the bottom legend: Vehicle, Building, Vegetation, Bike, Person, Pillar, Road, and Surface.
}
\label{fig:supp_occ_visualization}
\vspace{-1em}
\end{figure*}

\begin{figure*}[ht!]
\centering
\includegraphics[width=0.96\textwidth]{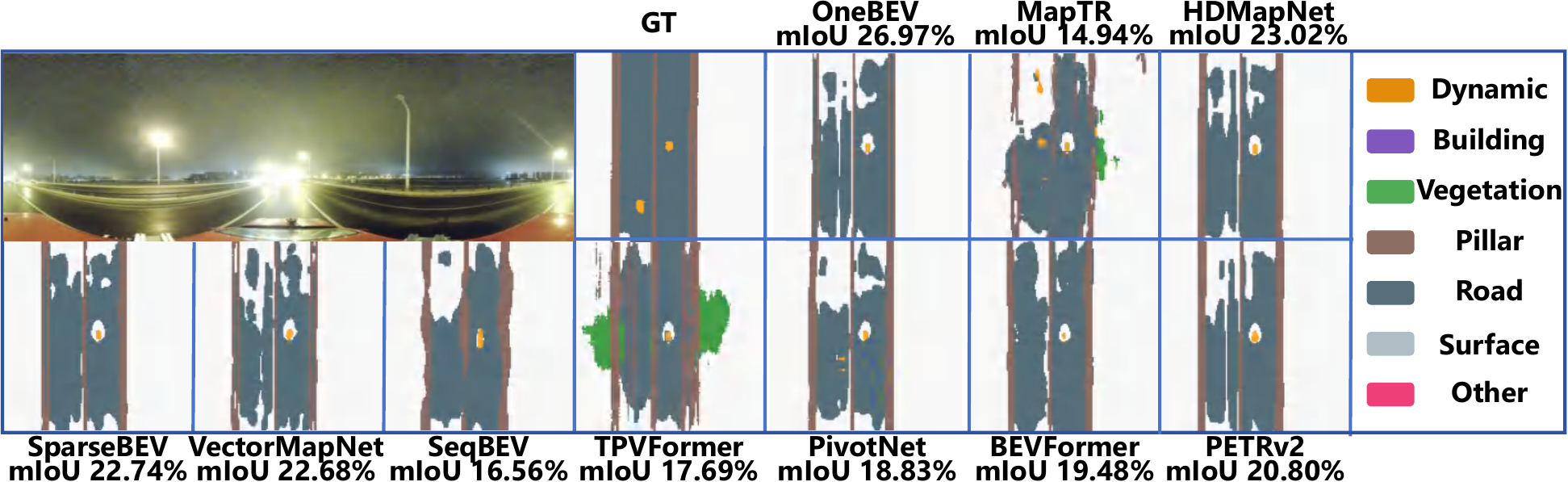}
\caption{
Qualitative results of bird's-eye-view (BEV) semantic mapping from spherical observations on Spheriverse.
Ground-truth (GT) semantic maps are compared with predictions from ten representative methods.
The compared methods include
OneBEV~\cite{onebev},
MapTR~\cite{maptr},
HDMapNet~\cite{hdmapnet},
SparseBEV~\cite{sparsebev},
VectorMapNet~\cite{vectormapnet},
SeqBEV~\cite{bevseq},
TPVFormer~\cite{TPVFormer},
PivotNet~\cite{pivotnet},
BEVFormer~\cite{bevformer}, and
PETRv2~\cite{petr}.
}
\label{fig:supp_semantic_mapping}
\end{figure*}

\begin{figure*}[ht!]
\centering
\includegraphics[width=0.98\textwidth]{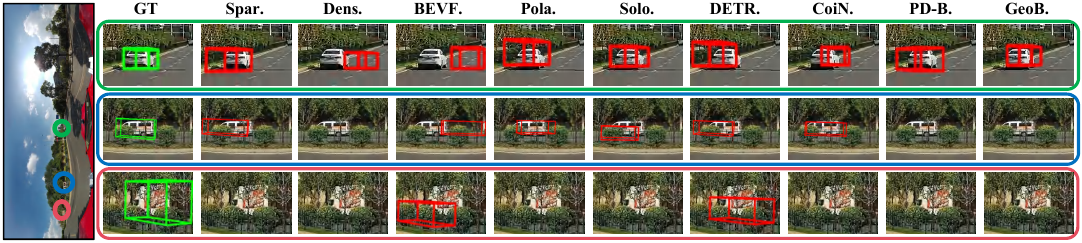}
\caption{
Qualitative results of spherical 3D object detection on Spheriverse.
The examples illustrate object localization from spherical observations under
diverse traffic scenes. The three color-coded regions in the input image are enlarged by row, while the ground truth (GT) and predictions from representative methods are compared by column. Spar., Dens., BEVF., Pola., Solo., DETR., CoiN., PD-B., and GeoB. denote SparseBEV~\cite{sparsebev}, DenseBEV~\cite{densebev}, BEVFormer~\cite{bevformer}, PolarBEVDet~\cite{polarbevdet}, SOLOFusion~\cite{Solofusion}, DETR3D~\cite{detr3d}, CoIn3D~\cite{coin3d}, PD-BEV~\cite{lu2025towards}, and GeoBEV~\cite{zhang2025geobev}, respectively. Green and red boxes denote the projected ground-truth and predicted 3D boxes.
}
\label{fig:supp_object_detection}
\end{figure*}

\subsection{Efficiency Analysis}
\label{app:efficiency}
As shown in Table~\ref{tab:parameter_efficiency}, CSRR and SER introduce only $0.001$M and $0.327$M additional parameters, respectively. Together, they increase the parameter count of the InternImage-T configuration by approximately $0.45\%$, while improving mIoU from $12.14\%$ to $13.91\%$ and GeoIoU from $23.35\%$ to $24.65\%$. These results demonstrate improved semantic occupancy prediction with a small parameter overhead. Combined with the backbone replacement, the complete model uses $18.43\%$ fewer parameters than the original baseline while achieving gains of $1.86$ and $2.10$ percentage points in mIoU and GeoIoU, respectively.

\begin{table}[!t]
\centering
\caption{
Parameter counts and semantic occupancy performance of progressive
model configurations on Spheriverse.
Parameters are reported in millions (M), and mIoU and GeoIoU in \%.
$\Delta$ Params denotes the change relative to the preceding
configuration. The final row compares the complete model with
the original baseline, with metric improvements in percentage points.
}
\label{tab:parameter_efficiency}

\setlength{\tabcolsep}{4.5pt}
\renewcommand{\arraystretch}{1.10}

\resizebox{\columnwidth}{!}{%
\begin{tabular}{lcccc}
\toprule
\textbf{Model}
& \textbf{Params (M)}
& \textbf{$\Delta$ Params (M)}
& \textbf{mIoU $\uparrow$}
& \textbf{GeoIoU $\uparrow$} \\
\midrule

Baseline
& 90.070
& --
& 12.05
& 22.55 \\

InternImage-T~\cite{internimage}
& 73.141
& $-16.929$
& 12.14
& 23.35 \\

\quad + CSRR
& 73.142
& $+0.001$
& 12.71
& 23.50 \\

\quad \textbf{+ SER (Ours)}
& \textbf{73.469}
& $+0.327$
& \textbf{13.91}
& \textbf{24.65} \\

\midrule
\multicolumn{2}{c}{\textbf{Overall Improvement}}
& $16.601 \downarrow$
& $1.86 \uparrow$
& $2.10 \uparrow$ \\

\bottomrule
\end{tabular}%
}
\end{table}

\subsection{More Qualitative Results}
\label{app:detection}
\textbf{More visualization of semantic occupancy prediction.}
Fig.~\ref{fig:supp_occ_visualization} presents additional qualitative results
under challenging spherical observations, including elevated bridge scenes and
camera-contaminated nighttime scenes. In the first example, the scene contains
a relatively rare elevated bridge structure. SphereOcc correctly preserves the
semantic occupancy geometry of the bridge, whereas the compared methods tend to
misclassify this structure as Pillar due to the similarity in vertical shape and
structural appearance. In the second example, the spherical image is affected by
nighttime contamination and contains a complex mixture of trees and vehicles.
SphereOcc still produces a more faithful semantic occupancy layout, while the
compared methods tend to over-predict a dominant category, such as Vegetation
or Vehicle. These qualitative results further support our motivation: visual information is
encoded in an angular domain, whereas the target occupancy representation lies
in a dense Cartesian voxel space. This domain discrepancy makes it difficult for
conventional methods to preserve rare structures and ambiguous semantic
boundaries reliably. By introducing CSRR and SER, SphereOcc injects
spherical-aware geometric priors into Cartesian voxel features and further
re-queries relevant semantic evidence from the source spherical observation.
As a result, the model can better align angular visual cues with Cartesian
occupancy features, leading to more reliable semantic predictions under rare
structures, visual contamination, and complex layouts. \textbf{Additional visualization of semantic mapping.}
As shown in Fig.~\ref{fig:supp_semantic_mapping}, we present BEV semantic mapping results spanning 50\,m in both the forward and rearward directions from the ego vehicle.
Unlike densely structured and cluttered urban environments, this example depicts a relatively simple road corridor with limited structural complexity.
Nevertheless, the nighttime spherical panorama is severely affected by glare from streetlights and distant headlights, resulting in widespread saturation and weakened texture and boundary cues, particularly at long range.
While the ground-truth map preserves continuous road geometry and well-defined roadside boundaries, existing methods frequently produce fragmented, truncated, or locally overextended road predictions in the far field.
These errors are especially pronounced along viewing directions dominated by intense glare, indicating that glare-affected far-field perception remains a key bottleneck for reliable camera-based BEV mapping.
This observation motivates future research into recovering and effectively exploiting long-range visual evidence under illumination saturation, thereby improving the stability of large-scale semantic mapping. \textbf{More visualization of Spherical 3D object detection.} 
Fig.~\ref{fig:supp_object_detection} presents additional qualitative results of
spherical 3D object detection on Spheriverse. Different from conventional
front-view detection, spherical 3D object detection needs to infer the
three-dimensional position and orientation of real-world objects from visual
evidence encoded in the angular image domain. The examples show that current
spherical 3D detection methods still suffer from missed detections, absolute
position errors, and orientation deviations. These failure cases reveal the
remaining challenges of full-surround object localization and provide useful
directions for future research on geometry-aware spherical 3D perception.

\section{Social Impact}
\label{sec:social_impact}

\textbf{Full-surround spatial understanding for embodied agents.}
Panoramic perception can extend embodied agents from a narrow front-view setting
to full $360^\circ$ environmental understanding. Compared with conventional
front-facing cameras, spherical images simultaneously cover the front, side, and
rear regions, allowing an agent to perceive road structures, building layouts,
obstacle distributions, and traversable spaces more comprehensively. Compared
with ring-shaped multi-camera systems, spherical imaging also provides
advantages in vertical field-of-view coverage, cross-directional scene
continuity, and image-level alignment consistency. These properties are
important for autonomous driving, mobile robots, and low-speed unmanned systems,
where critical spatial cues do not always appear in the forward-facing view.

\textbf{Social interaction perception.}
Embodied agents often operate in environments shared by pedestrians, vehicles,
cyclists, and other dynamic participants. Spherical observations provide richer
surrounding behavioral cues, such as pedestrians approaching from the side or
rear, parallel vehicles, merging traffic, and multi-directional interactions at
intersections. Such information can help agents reason about spatial
relationships and potential motion intentions among surrounding entities,
providing a perceptual basis for more natural and safer social interaction.

\textbf{Safety-critical object perception.}
Many safety risks originate outside the conventional front-view field of view,
including lateral vehicles, pedestrians in blind spots, rear-approaching traffic,
and roadside obstacles. Spherical perception can reduce visual blind spots and
enable models to detect, localize, and represent safety-critical objects within
a full-surround range. This capability is important for improving risk awareness
and spatial completeness in safety-critical perception.

\textbf{Cross-view localization and spatial services.}
Because spherical images contain broad environmental context, they are also
useful for cross-view localization, visual relocalization, map construction, and
spatial services. For example, an agent may match spherical observations with
existing maps, satellite views, or street-view imagery to improve localization
robustness. Spherical perception outputs can also be converted into semantic
spatial maps to support navigation, inspection, and city-scale spatial
understanding. Therefore, Spheriverse is not limited to occupancy prediction,
but can also serve as a data resource for future spherical spatial intelligence
tasks.

\textbf{Potential for aerial--ground interaction and low-altitude autonomy.}
With the development of low-altitude autonomy, UAV inspection, aerial--ground
collaborative perception, and broader spatial intelligence applications, visual
systems with wide field-of-view coverage, strong spatial understanding, and
multi-directional perception are becoming increasingly important. Spherical perception can provide low-altitude platforms and ground agents with
more complete environmental context, enabling collaborative perception for
spatial modeling, object discovery, and path understanding in complex scenes.
By studying spherical 3D perception, Spheriverse may provide a
foundation for extending spherical vision to low-altitude perception,
aerial--ground collaboration, and more general embodied spatial intelligence.

\end{document}